%% file: main.tex
\documentclass[final]{clv2025}

\usepackage{paralist}
\usepackage{xcolor}
\definecolor{darkblue}{rgb}{0, 0, 0.5}
\usepackage{hyperref}
\hypersetup{colorlinks=true,citecolor=darkblue, linkcolor=darkblue, urlcolor=darkblue}
\usepackage{xspace}
\usepackage{tcolorbox} 
\usepackage{longtable}
\usepackage{booktabs} 
\usepackage{multirow}
\usepackage{arydshln} 
\usepackage{float} 
\usepackage{array} 
\usepackage{tabularx} 
\usepackage{bigstrut}
\usepackage{amsmath}
\usepackage{amsfonts}
\usepackage{amssymb}
\usepackage{graphicx}
\usepackage{svg}
\usepackage{subcaption}
\usepackage{enumitem}
\usepackage{url}

\usepackage{listings}
\usepackage{placeins} 
\usepackage[export]{adjustbox}
\usepackage{makecell}
\usepackage{pifont}

\usepackage[textwidth=4pc,colorinlistoftodos]{todonotes}
\setuptodonotes{linecolor=black,color=yellow!30,size=\tiny}

\newcommand{\A}[0]{\textbf{A}\xspace}
\newcommand{\B}[0]{\textbf{B}\xspace}
\newcommand{\Target}[0]{\textit{Target}\xspace}
\newcommand{\Built}[0]{\textit{Built}\xspace}

\newcommand{\exampleutt}[1]{\textit{``#1"}}

\DeclareMathOperator*{\argmin}{arg\,min}

\newcommand{\block}{\mathbf{b}}
\newcommand{\cell}{\mathbf{c}}
\newcommand{\Colors}{\textsc{BlockColors}}
\newcommand{\BR}{\textsc{BuildRegion}}
\newcommand{\actionspace}{\textsc{ActionSpace}}

\newcommand{\colorfont}[1]{\textsc{#1}}

\newcommand{\red}{\colorfont{Red}}
\newcommand{\orange}{\colorfont{Orange}}
\newcommand{\yellow}{\colorfont{Yellow}}
\newcommand{\green}{\colorfont{Green}}
\newcommand{\blue}{\colorfont{Blue}}
\newcommand{\purple}{\colorfont{Purple}}
\newcommand{\loc}{\langle x, y, z\rangle}
\newcommand{\locPrime}{\langle x', y', z'\rangle}
\newcommand{\locB}{\langle x_\B, y_\B, z_\B\rangle}

\newcommand{\locMin}{\langle x_{\min}, y_{\min}, z_{\min}\rangle}
\newcommand{\locMax}{\langle x_{\max}, y_{\max}, z_{\max}\rangle}

\newcommand{\dr}[0]{\ensuremath{D_{\mathrm{r}}}\xspace}
\newcommand{\dbs}[0]{$D_{\mathrm{sb}}$\xspace}
\newcommand{\dss}[0]{$D_{\mathrm{ss}}$\xspace}
\newcommand{\dmc}[0]{$D_{\mathrm{mc}}$\xspace}

\newcommand{\syndata}[0]{\dbs, \dss, \dr}
\newcommand{\alldata}[0]{\syndata, \dmc}

\newcommand{\mmc}[0]{$M_{\mathrm{mc}}$\xspace}
\newcommand{\mbs}[0]{$M_{\mathrm{sb}}$\xspace}
\newcommand{\mss}[0]{$M_{\mathrm{ss}}$\xspace}
\newcommand{\mr}[0]{$M_{\mathrm{r}}$\xspace}
\newcommand{\magg}[0]{$M_{\mathrm{agg}}$\xspace}
\newcommand{\maggcl}[0]{$M_{\mathrm{agg+cl}}$\xspace}

\newcommand{\mmcp}[0]{$M_{\mathrm{mc}}^{\prime}$\xspace}
\newcommand{\maggp}[0]{$M_{\mathrm{agg}}^{\prime}$\xspace}

\newcommand{\step}[1]{\textit{\hyperref[list:step_#1]{Step #1}}}

\newcommand{\mmcpbpos}[0]{$M_{\mathrm{mc+posb}}^{\prime}$\xspace}
\newcommand{\maggpbpos}[0]{$M_{\mathrm{agg+posb}}^{\prime}$\xspace}
\newcommand{\maggpbposws}[0]{$M_{\mathrm{agg+posb+s}}^{\prime}$\xspace}

\newcommand{\repV}{\(N\)\xspace}
\newcommand{\repVPosB}{\(N + \textsc{Pos}_{\B}\)\xspace}
\newcommand{\repVPosBS}{\(N + \textsc{Pos}_{\B} + S\)\xspace}

\newcommand{\llmbest}[0]{Llama-CRAFTS\xspace}
\newcommand{\llmbaseline}[0]{Nebula\xspace}

\newcommand{\figposspace}{\vspace{2mm}}
\newcommand{\fignegspace}{\vspace{-8mm}}

\runningtitle{BAP v2}
\runningauthor{Jayannavar et al.}

\begin{document}

\title{BAP v2: An Enhanced Task Framework for Instruction Following in Minecraft Dialogues}

\author{Prashant Jayannavar\thanks{Corresponding author}$^{,1}$, Liliang Ren\thanks{Work done while author was a graduate student or undergraduate intern at Illinois}$^{,2}$, Marisa Hudspeth$^{**,3}$, Risham Sidhu$^{1}$, Charlotte Lambert\thanks{Equal contribution}$^{,**,1}$, Ariel Cordes$^{\dagger,**}$, Elizabeth Kaplan$^{\dagger,**,4}$, Anjali Narayan-Chen$^{\dagger,**,5}$, Julia Hockenmaier$^{*,1}$}

\affilblock{
    \affil{University of Illinois Urbana-Champaign\\\quad \email{\{paj3, juliahmr\}@illinois.edu}}
    \affil{Microsoft}
    \affil{University of Massachusetts Amherst}
    \affil{Amazon}
    \affil{Amazon AGI}
}

\maketitle

\begin{abstract}
Developing interactive agents that can understand language, perceive their surroundings, and act within the physical world is a long-standing goal of AI research. The \textbf{Minecraft Collaborative Building Task (MCBT)} \citep{narayan-chen-etal-2019-collaborative}, 
a two-player game in which an Architect (\A) instructs a Builder (\B) to construct a target structure in a simulated 3D Blocks World environment, offers a rich platform to work towards this goal. 

In this work, we focus on the \textbf{Builder Action Prediction (BAP)} subtask: predicting \B's actions in a multimodal game context  \citep{jayannavar-etal-2020-learning} -- a challenging testbed for grounded instruction following, with limited training data. 
We holistically re-examine this task and introduce \textbf{BAP v2} to  address key challenges in evaluation, training data, and modeling. Specifically, we define an \textbf{enhanced evaluation benchmark}, featuring a cleaner test set and fairer, more insightful metrics that also reveal spatial reasoning as the primary performance bottleneck. To address data scarcity and to teach models basic spatial skills, we generate different types of \textbf{synthetic MCBT data}. We observe that current, LLM-based SOTA models trained on the human BAP dialogues fail on these simpler, synthetic BAP ones, but show that training models on this synthetic data improves their performance across the board. We also introduce  \textbf{a new SOTA model, \llmbest}, which leverages richer input representations, and achieves an F1 score of 53.0 on the BAP v2 task and strong performance on the synthetic data. While this result marks a notable 6  points improvement over previous work, it also underscores the task's remaining difficulty, establishing BAP v2 as a fertile ground for future research, and providing a useful measure of the spatial capabilities of current text-only LLMs in such embodied tasks.
\end{abstract}

\input{intro}

\input{background}

\input{models}

\input{eval}

\input{data}


\input{incorporate_syn_data}

\input{analysis}

\input{discussion}

\input{related_work}

\input{conclusion}

\input{appendix_pdoc/appendix_full}

\begin{acknowledgments}
We would like to thank Hetvi Patel, Sana Madhavan, Marc Canby, Siddarth Madala, Rajarshi Haldar, Katya Yegorova, Adam Davies, Ansel Blume, and Sara Aghajanzadeh for their valuable help with the  annotation, and Dan Roth, Martha Palmer, Janna Doppa and Sriraam Natarajan and all other members of our CwC team for many helpful discussions earlier in this project. 
This work was supported by Contract W911NF-15-1-0461 with the US Defense Advanced Research Projects Agency (DARPA) Communicating with Computers Program and the Army Research Office (ARO). Approved for Public Release, Distribution Unlimited. The views expressed are those of the authors and do not reflect the official policy or position of the Department of Defense or the U.S. Government.
The work was also supported by the Distributed Research Experiences for Undergraduates (DREU) program, a joint project of the CRA Committee on the Status of Women in Computing Research (CRA-W) and the Coalition to Diversify Computing (CDC), which is funded in part by the NSF Broadening Participation in Computing program (NSF BPC-A \#1246649).
The work utilizes resources \cite{10.1145/3311790.3396649} supported by the National Science Foundation’s Major Research Instrumentation program, grant \#1725729, as well as the University of Illinois at Urbana-Champaign, as well as the Delta advanced computing and data resource, which is supported by the National Science Foundation (award OAC 2005572) and the State of Illinois. Delta is a joint effort of the University of Illinois Urbana-Champaign and its National Center for Supercomputing Applications.
\end{acknowledgments}

\bibliographystyle{compling}
\bibliography{main}

\end{document}

%% file: intro.tex
\section{Introduction}
\label{sec:introduction}
Developing interactive agents that can understand language, perceive their surroundings, and act within the physical world is a long-standing goal of AI research (e.g.,~\citet{winograd1971procedures}). We work toward this goal in Minecraft ({\footnotesize \url{https://minecraft.net/}}), a popular 3D game that has become a prominent platform for AI experimentation. Our previous work introduced the \textbf{Minecraft Collaborative Building Task (MCBT)} and the corresponding \textbf{Minecraft Dialogue Corpus (MDC)}~\cite{narayan-chen-etal-2019-collaborative}, one of the first efforts to use this environment to study grounded, task-oriented dialogue.

\begin{figure*}[htbp]
    \centering
    \includegraphics[width=\textwidth]{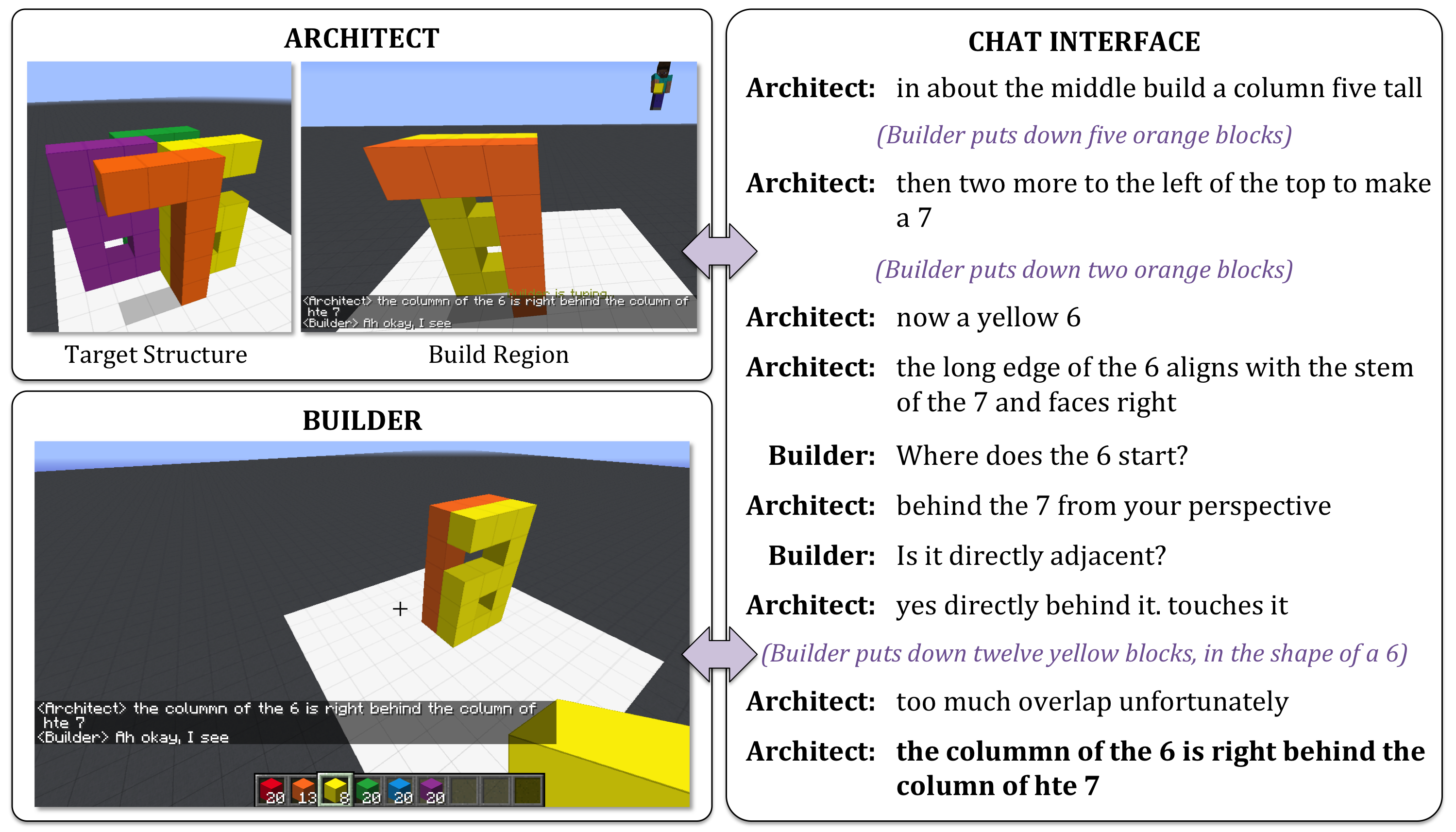}
    \caption{
    In the Minecraft Collaborative Building Task, the Architect (\A) has to instruct a Builder (\B) to build a target structure. \A can observe \B, but remains invisible to \B. Both players communicate via a chat interface. (NB: We show \B's actions in the dialogue as a visual aid to the reader.)
    }
    \label{fig:dialoguescreenshots}
\end{figure*}

In the MCBT, two players collaborate to construct a target structure from multi-colored blocks (Figure~\ref{fig:dialoguescreenshots}): an Architect (\A) provides instructions to a Builder (\B). The task is situated in a dynamic 3D world that participants must refer to and modify from constantly shifting perspectives. The dialogue is asynchronous, asymmetric, and has minimal linguistic constraints. The MDC contains 509 human-human game logs from this task. While \citet{narayan-chen-etal-2019-collaborative} focused on generating Architect utterances, \citet{jayannavar-etal-2020-learning} introduced end-to-end models for an automated Builder. The latter work defined the challenging \textbf{Builder Action Prediction (BAP)} task: predicting the sequence of actions (block placements and/or removals) that a human Builder performed at a given point in a human-human game. We review prior work on the Minecraft Builder and related recent works in Section~\ref{sec:background}.


The BAP task remains \textbf{a relevant and challenging testbed for grounded instruction following}, presenting a unique combination of difficulties. Unlike navigation or object-centric benchmarks such as VLN~\citep{anderson2018vision}, CVDN~\citep{thomason:arxiv19}, ALFRED~\citep{Shridhar_2020_CVPR}, DialFRED~\citep{Gao2022DialFREDDA}, and TEACh~\citep{Padmakumar_Thomason_Shrivastava_Lange_Narayan-Chen_Gella_Piramuthu_Tur_Hakkani-Tur_2022}, where agents primarily interact with existing objects, the MCBT requires grounding instructions that refer to objects and structures which do not yet exist, demanding greater abstraction and planning. The dialogue is also more complex than in un-embodied benchmarks like MultiWoz~\citep{budzianowski2018multiwoz} or other Minecraft datasets like IGLU~\citep{mohanty2024idat, pmlr-v220-kiseleva23a, pmlr-v176-kiseleva22a}; it features free-form, asynchronous communication within a world that is dynamically modified. Finally, the task poses a significant planning problem due to its  large action space---over 7,600 possible actions compared to just six in typical VLN work---and its reliance on complex, perspective-dependent spatial relations (e.g., \exampleutt{to your left}), which are more difficult to resolve than the absolute cardinal directions used in IGLU. This combination of challenges makes BAP a compelling setting for research.

However, to enable future research to make more efficient and meaningful progress, this paper holistically re-examines the BAP task framework. In particular, we identify and address key challenges in \textbf{evaluation, training data, and modeling}, and propose \textbf{BAP v2}, an upgraded version of the task:\footnote{All code and data developed for this research will be released at  \url{https://github.com/prashant-jayan21/bap-v2}}


\paragraph{Evaluation (Sections~\ref{sec:eval} and~\ref{sec:eval_metrics})}
The complexity of the BAP task makes evaluation nuanced. We systematically address challenges that hinder fair and insightful assessment by introducing a cleaner \emph{v2 test set} (Section~\ref{sec:eval}),  propose a \emph{fairer F1 metric} and introduce new \emph{auxiliary metrics} to measure specific model capabilities (Section~\ref{sec:eval_metrics}). Together, these contributions establish the \textbf{BAP v2 evaluation benchmark} and help us move beyond the original, opaque F1 score, providing deeper insights. Adopting this new benchmark also reveals that \textbf{spatial reasoning is the key performance bottleneck} in the BAP task.


\paragraph{Training Data (Section~\ref{sec:syn_data_gen})}
To overcome the dual challenges of limited training data for the BAP task—stemming from the small size of the MDC (509 game logs)—and the key performance bottleneck of spatial reasoning, we create a diverse set of \textbf{synthetic training data}. This data is generated using novel Minecraft dialogue and target structure simulators that were designed emulate the MCBT. While simpler than the human MDC data, the synthetic dialogues are rich in the spatial relations and referring expressions crucial for this task. The combination of this new data with the original corpus forms the \textbf{BAP v2 training set}.
Our approach builds on established work showing the effectiveness of synthetic data for various tasks, from vision-and-language navigation to dialogue systems \citep{Kamath_2023_CVPR, ku-etal-2020-room, Wang_2023_ICCV, Kang_2023_CVPR, visdial, kim-etal-2022-generating, bao-etal-2023-synthetic, zhan-etal-2023-turning, zhan-etal-2024-going, sparkel}, although synthesizing complete, task-oriented embodied dialogues remains a relatively underexplored area \citep{padmakumar-etal-2023-multimodal}, especially since the MCBT is significantly more complex than other Blocks World settings \citep{bisk-etal-2016-natural, dan-etal-2021-compositional-data}.

\paragraph{Modeling and Analysis (Sections~\ref{sec:syn_data_training} and~\ref{sec:analysis})}
To demonstrate the utility of our synthetic data in addressing the challenge of limited original BAP data, we show that it produces more performant models, for both LLM and non-LLM architectures, even with straightforward training methods (Section~\ref{sec:syn_data_training}). 
The data also reveals that \textbf{both LLM and non-LLM models trained solely on the complex BAP human dialogues lack robustness}, counterintuitively underperforming on our simpler synthetic data.
As the final component of the BAP v2 framework, we extend prior work on LLM-based models for the BAP task \citep{chaturvedi-etal-2024-nebula} by introducing richer input text representations that capture essential game context, and we find that these enhanced representations and the synthetic data are both necessary and mutually reinforcing. Our best model, \textbf{\llmbest} (\textbf{C}ontext \textbf{R}ich \textbf{A}nd \textbf{F}ine-\textbf{T}uned On \textbf{S}ynthetic Data), establishes a new state of the art on the BAP task as well as achieves strong performance on the synthetic data. But while its F1 score of 53.0 on BAP represents a 6 point improvement over the previous SOTA \citep{chaturvedi-etal-2024-nebula} (under the new v2 evaluation), it is nevertheless clear that there is much room for further improvement. 
A detailed analysis of model output (Section~\ref{sec:analysis}) illustrates how \llmbest improves over the previous SOTA (especially in spatial reasoning), where it falls short, what this implies for current text-only LLMs' spatial capabilities in such embodied tasks, and indicates both strengths and limitations of our new evaluation metrics.

While spatial reasoning in language models has been explored in text and across other modalities \citep{mirzaee-kordjamshidi-2022-transfer, shi2022stepgamenewbenchmarkrobust, rizvi-etal-2024-sparc, gopinathan-etal-2024-spatially, li-etal-2024-topviewrs, 10.5555/3504035.3504651, du-etal-2024-embspatial, kamath2023whatsupvisionlanguagemodels, wang2024pictureworththousandwords}, our work addresses a specific gap in the literature: exploring the capabilities of text-only LLMs in a 3D, embodied, collaborative dialogue setting. Prior work has shown LLMs struggle with even simple 2D grid navigation and path-planning in text \citep{yamada2024evaluating, aghzal2024can}, and has explored spatial reasoning in games like chess \citep{feng2023chessgptbridgingpolicylearning, zhang-etal-2025-complete} or puzzles \citep{chollet2025arcprize2024technical}, but no existing research combines this with the challenges of 3D grids in text, and embodied, collaborative dialogue, thus motivating our focus.

\paragraph{Part 4: Looking Ahead (Sections~\ref{sec:discussion} and~\ref{sec:concl_fw})}
Finally, we discuss future research directions (Section~\ref{sec:discussion}) We outline the key next challenges for BAP that can be tackled by building on the BAP v2 framework, and explore the broader implications of our work beyond BAP, including way in which ideas from our synthetic data and simulators  hold the potential for furthering other related works and tasks \citep{narayan-chen-etal-2019-collaborative, mohanty2024idat, bonn-etal-2020-spatial, thompson-etal-2024-discourse, bonial-etal-2021-builder, madge2025mdcrminecraftdialoguecorpus}. We conclude in Section~\ref{sec:concl_fw}.

%% file: background.tex
\section{Prior Work on the Minecraft Builder}
\label{sec:background}

Sections \ref{sec:minecraft_task},\ref{sec:BAP},\ref{sec:dataaug},\ref{ssec:background_eval}  and \ref{sec:baseline_model_desc} provide a high-level overview of the Minecraft Collaborative Building Task and Dialogue corpus, the BAP task and dataset,  evaluation metrics, and baseline model, summarizing our prior work \citep{narayan-chen-etal-2019-collaborative, jayannavar-etal-2020-learning}. Section \ref{sec:Formalizing-BAP} formalizes the BAP task. Section~\ref{sec:rw_llms_bap} discusses recent related work around more contemporary LLM-based models for BAP \citep{chaturvedi-etal-2024-nebula, ch-etal-2024-retrieval}, and Section~\ref{sec:rw_task_and_data} discusses related work around the task and dataset itself, explaining why the BAP task continues to be a relevant and challenging testbed for grounded instruction following.

\begin{figure*}[htbp]
\centering
\setlength{\fboxsep}{0pt}
\newcommand{\panelwidth}{0.22\textwidth}

\begin{tabular}{cccc}
\toprule
\multicolumn{4}{l}{\textbf{A snippet from the MDC corpus}}\\
\midrule
  \includegraphics[width=\panelwidth,valign=t]{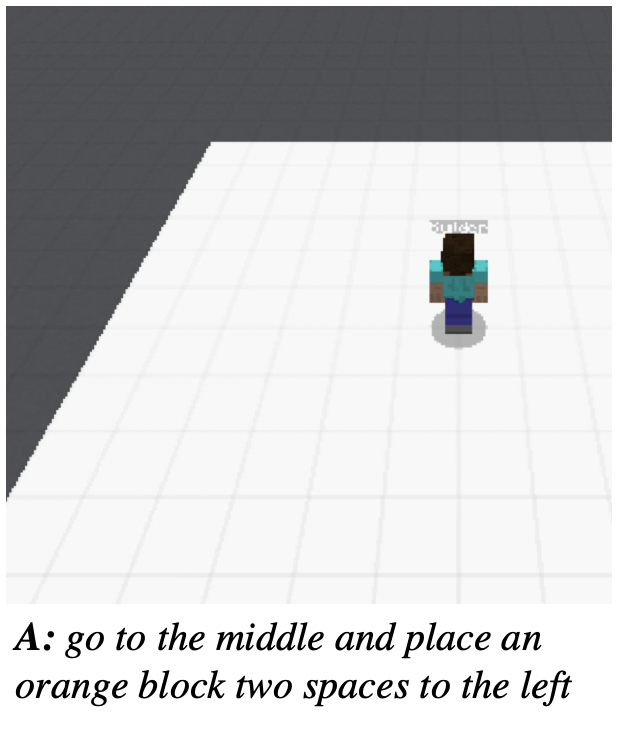}
 &  \includegraphics[width=\panelwidth,valign=t]{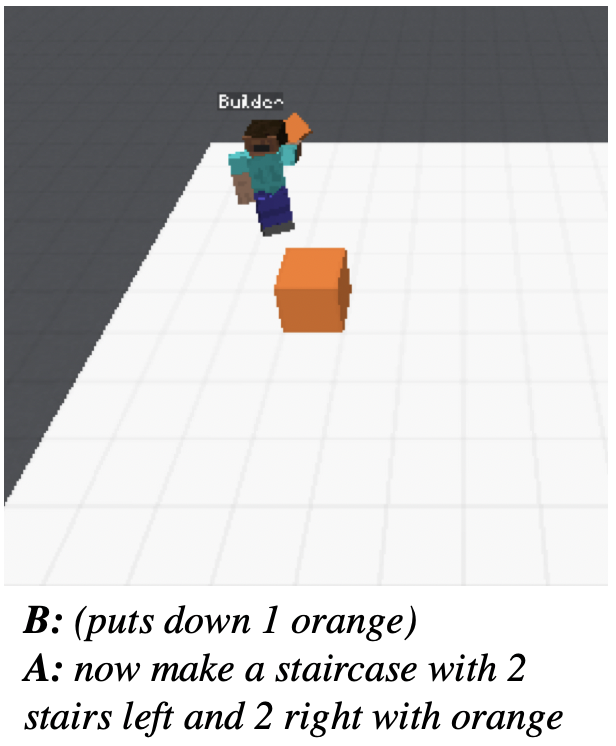}
 &
  \includegraphics[width=\panelwidth,valign=t]{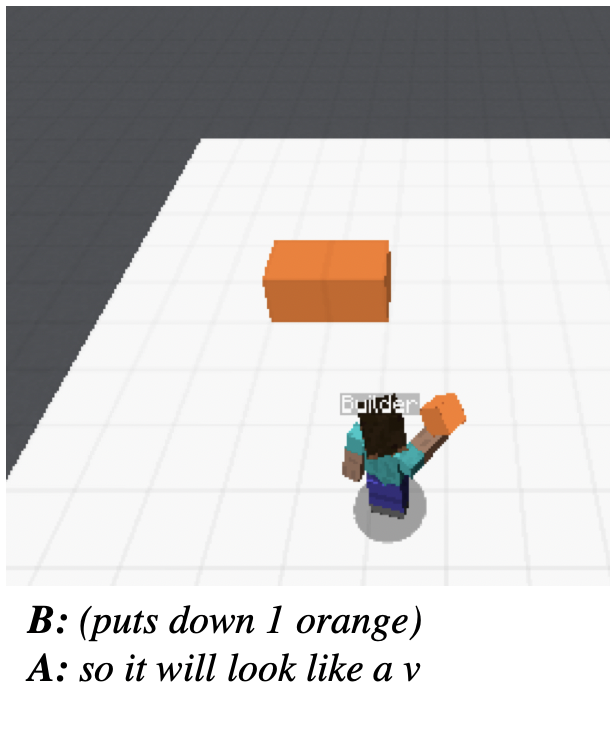}

&
  \includegraphics[width=\panelwidth,valign=t]{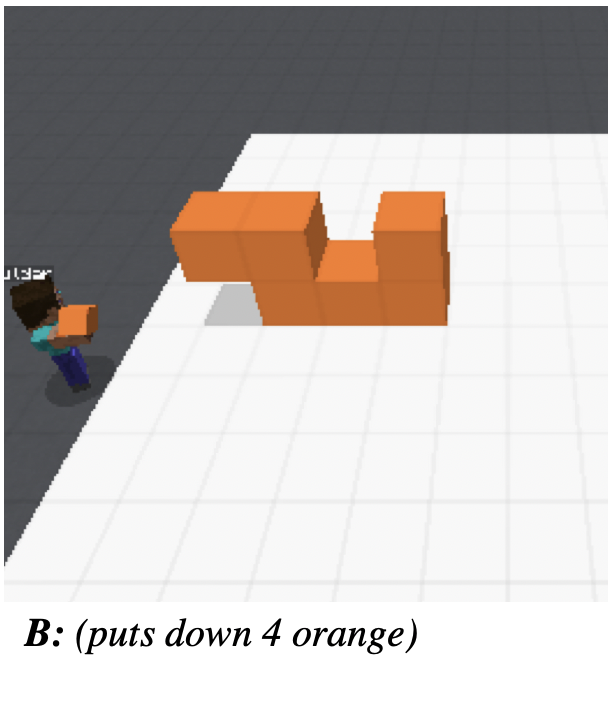}
 \\
 (a) & (b) & (c) & (d)\\
 
& & &  \\
  \includegraphics[width=\panelwidth,valign=t]{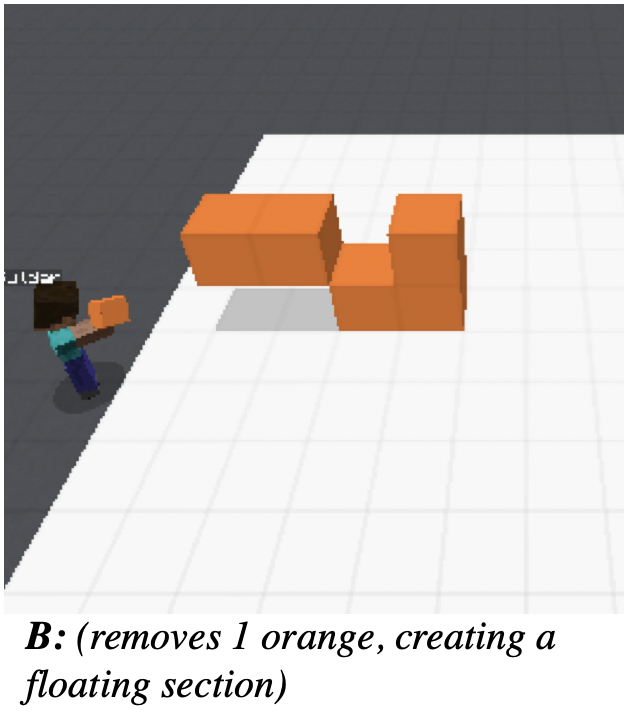}
 &
  \includegraphics[width=\panelwidth,valign=t]{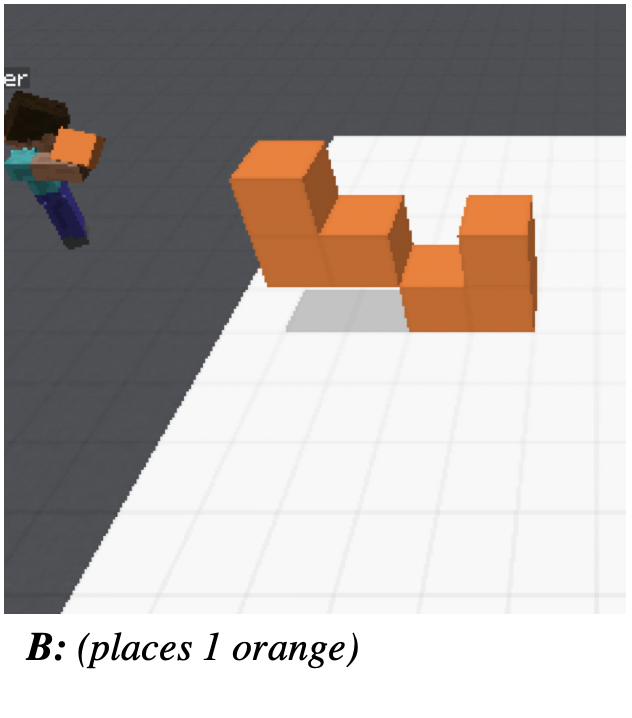}
 &
  \includegraphics[width=\panelwidth,valign=t]{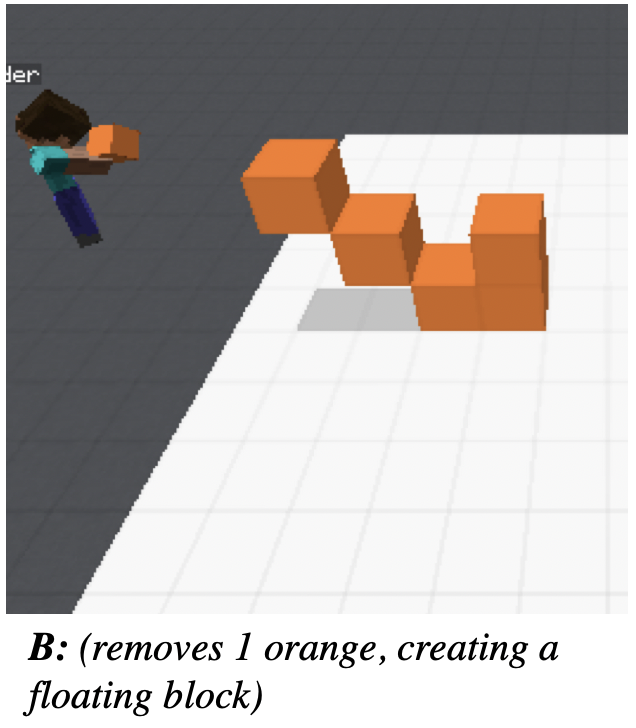}
 &
  \includegraphics[width=\panelwidth,valign=t]{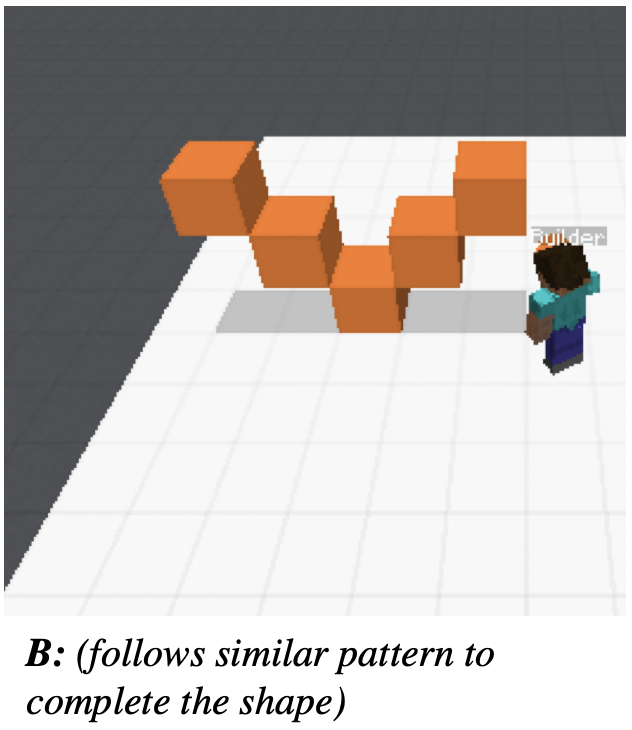}
\\
 (e) & (f) & (g) & (h)\\
 \bottomrule\\
\end{tabular}
\caption{A sample sequence of human-human game states from the MDC corpus. The game starts with an empty board and an initial \A instruction (a), which \B executes in the first action sequence (b) by placing a single block. In (c), \B  begins to execute the next \A instruction given in (b). But \A interrupts \B in (c), leading to a new \B action sequence in  (c)--(h) (multiple placements and removals).}
\label{fig:sequence}
\end{figure*}

\subsection{The Minecraft Collaborative Building Task and Dialogue Corpus}
\label{sec:minecraft_task}
The \textbf{Minecraft Collaborative Building Task (MCBT)} is a two-player game in which player \A (the \textit{Architect}) has to instruct player \B (the \textit{Builder}) to create a copy of a \Target structure that is only shown to \A. \B controls a Minecraft agent that is given a fixed inventory of blocks.  \A has access to two Minecraft windows, one which contains the \Target, and one in which it can observe \B's actions.  \A remains invisible to \B and cannot place blocks itself. \A and \B can only communicate via a text-based chat interface that both can use freely throughout the game.   The \textbf{Minecraft Dialogue Corpus (MDC)} consists of 509 human-human dialogues and game logs for this task, collected via a modification of the Malmo Minecraft client \citep{johnson2016malmo}. In the MCBT, each block \(\block\)  has one of six colors (red, orange, yellow, green, blue, purple), and  \B  starts with an inventory of 120 blocks (twenty of each color) that it can place into or remove from a predefined \(\BR\). Like in standard Minecraft, a discrete grid that is superimposed on the environment defines where blocks can be placed.
In the Minecraft API, grid cells are indexed by integer-valued coordinates \(\cell = \loc\), where  
\(y\)  indicates vertical height, 
but  these coordinates are not exposed in the graphical interface, and players are unable to use them to identify specific cells in their conversations. 
Blocks can only be placed in empty grid cells that are either on the ground (\(y=1\)) or adjacent to an existing block.  If the Builder picks up (removes) a block, its grid cell becomes empty again, and the block returns to the inventory. 

The games in the MDC consist mainly of \A providing instructions, often involving multiple actions to be taken, and grounded in the Builder’s perspective, while \B executes those instructions and resolves any confusion through clarification questions and further dialogue. They are based on 150 distinct target structures, split into disjoint test, training, and development sets such that training targets do not appear during test or development. Figure~\ref{fig:dialoguescreenshots} (left) shows the perspectives seen by each player and a snippet of their conversation (right) taken from this corpus. 
The environment in which \B operates contains the predefined \(\BR\) (shown as the white square on the ground). The task is complete when the \Built structure inside the Build Region matches the \Target, allowing for translations and rotations in the horizontal plane.

\subsubsection{Features and Challenges of the MDC}
\label{sec:features_mcbt}
Since target structures can be fairly complex, Architects typically give step-by-step instructions (\exampleutt{now a yellow 6}) for different parts of the target. Builders should execute these instructions, but  may also ask questions (\exampleutt{Where does the 6 start}) which the Architects has to answer (\exampleutt{behind the 7 from your perspective}). Architects may also need to identify and correct mistakes the Builder may have made (\exampleutt{too much overlap unfortunately}). 
\A can move around freely, but remains invisible to \B and views the structure from behind \B when giving instructions. As a result, \A instructions frequently include spatial relations, both between pairs of blocks or substructures (\exampleutt{put ... on top of...}), and relative to \B's current position and perspective (\exampleutt{left}, \exampleutt{right}). Humans also often use higher-level descriptions to refer to complex (sub)shapes (e.g. \exampleutt{staircase}, \exampleutt{v}). Due to the asynchronous nature of the dialogue, the Architect often talks while the Builder is placing blocks (\exampleutt{so it will look like a v}), leading to an apparent interruption of the Builder's action sequences. Finally, Minecraft, blocks do not need to be placed on the ground if their cell is adjacent to a supporting block.   If that supporting block is later removed, the remaining block (and any structure supported by it) will ``float'' in place. Thus, placing floating blocks needs the placement and subsequent removal of such placeholder supporting blocks.
Instructions for floating structures vary greatly, ranging from step-by-step instructions involving temporary supporting blocks to single-shot descriptions such as, simply, \exampleutt{place a floating yellow block}. 
Figure~\ref{fig:sequence} shows an example from the MDC that highlights some of these features and challenges.

\subsubsection{MCBT subtasks: Architect Utterance Generation and Builder Action Prediction}
\label{sec:mcbt_subtasks}
Although the MDC should ultimately contribute to the creation of fully interactive agents that can complete an entire game, we have found it helpful, and even necessary, to start with simpler tasks. Specifically, we have defined the \textbf{Architect Utterance Generation (AUG) Task} as the task of generating a suitable Architect utterance at any point in a  human-human game at which the human Architect produced an utterance \citep{narayan-chen-etal-2019-collaborative}, and the \textbf{Builder Action Prediction (BAP) Task}  as the corresponding task of generating a suitable Builder action sequence (consisting of block placements/removals) at any point in a human-human game at which the human Builder placed or removed blocks \citep{jayannavar-etal-2020-learning}.
Although both of these tasks consider only a single turn, and assume the context of a game and dialogue between two human players, they are important first steps towards the creation of agents that can complete an entire game in the MCBT with a human counterpart.

\begin{figure}[!htbp]
\begin{tabular}{l}
\toprule
\textbf{Three BAP task items} corresponding to the snippet in Figure~\ref{fig:sequence}\\
\midrule
\textbf{Item 1} corresponding to (a)–(b) in Fig.~\ref{fig:sequence}\figposspace\\
\includegraphics[width=\textwidth]{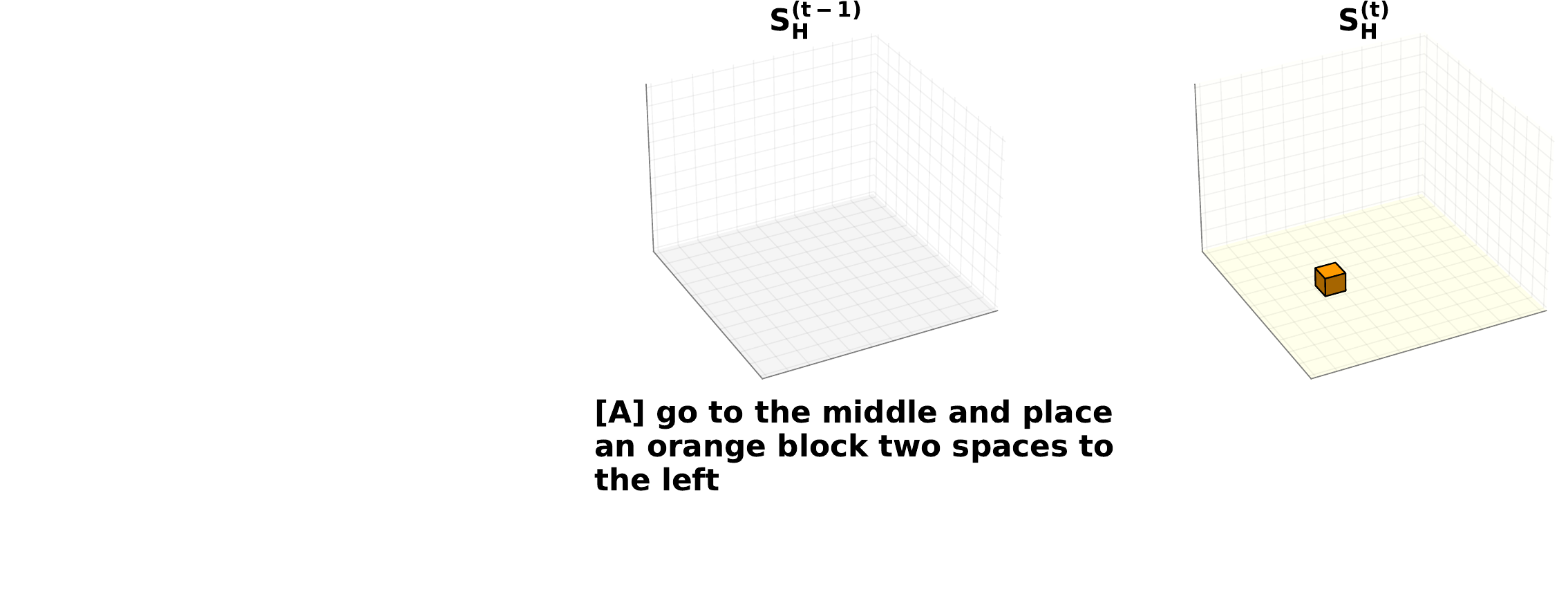}
\fignegspace\\

\midrule
\textbf{Item 2} corresponding to (b)–(c) in Fig.~\ref{fig:sequence}\figposspace\\
\includegraphics[width=\textwidth]{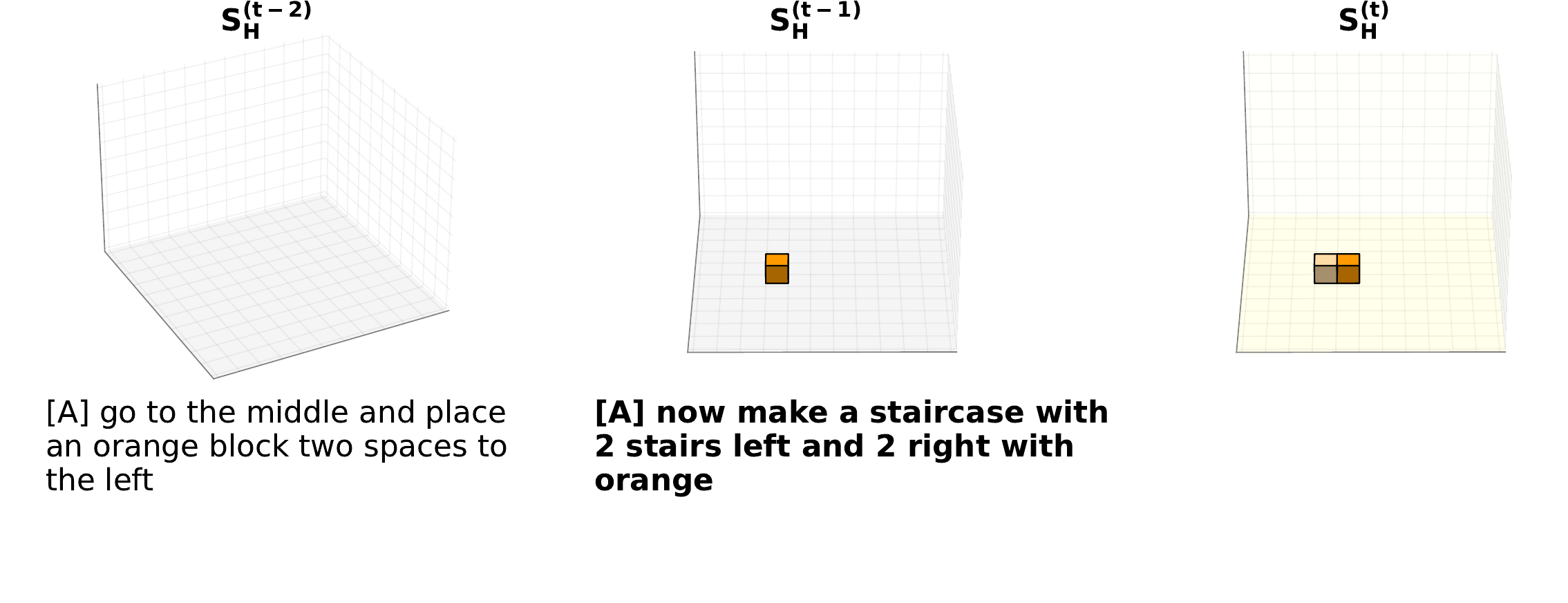}\fignegspace\\
\midrule
\textbf{Item 3} corresponding to (c)–(h) in Fig.~\ref{fig:sequence}\figposspace\\
\includegraphics[width=\textwidth]{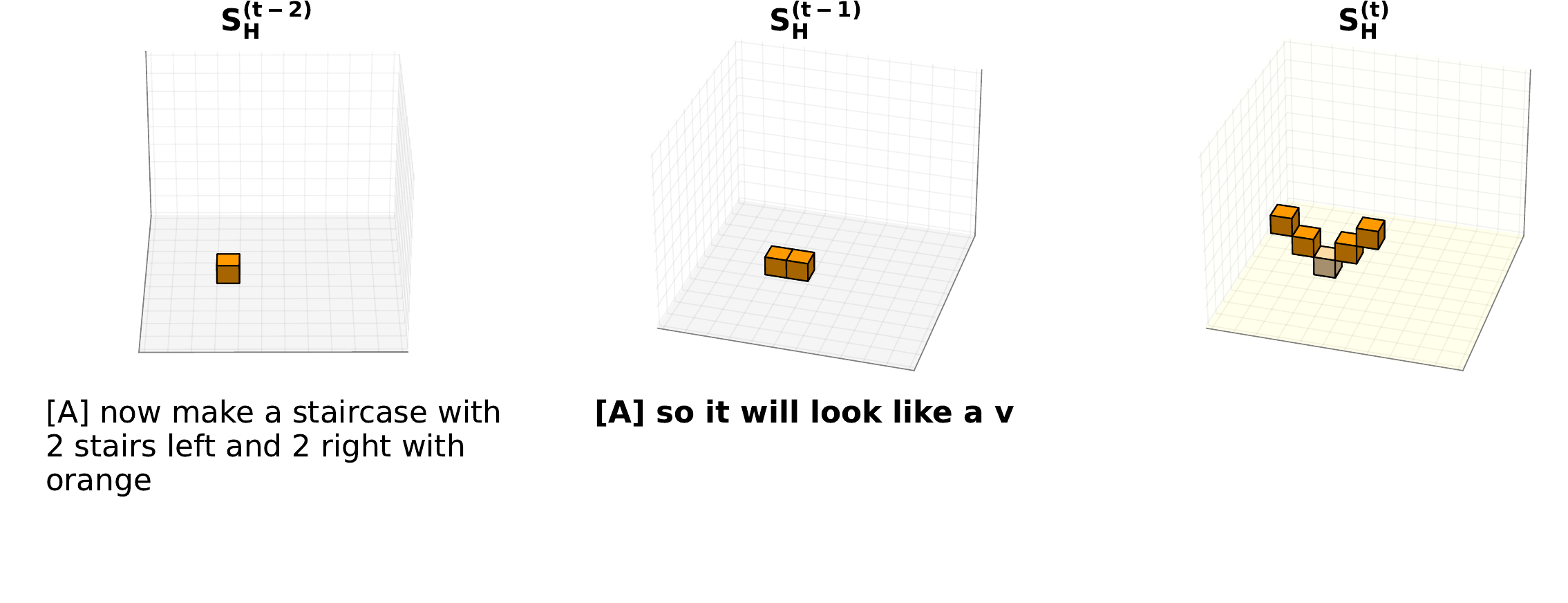}\fignegspace\\
\bottomrule 
\end{tabular}
\caption{The snippet in Figure~\ref{fig:sequence} corresponds to three BAP task items shown here.}
\label{fig:bap_examples}
\end{figure}

\subsection{The Builder Action Prediction (BAP) Task}
\label{sec:BAP}
The Builder Action Prediction (BAP) task is defined as the task of predicting the actions (block placements and/or removals) that a human Builder performed at a particular point in a human-human game. Games start with an empty board, and consist of an alternating sequence of dialogues \(D^{t}\)  (which may each consist of a single utterance, or a back-and-forth exchange between the two players) and Builder actions  \(A^{t}\) that result in an updated built structure \(S^{t}\). 
Each action sequence \(A^{t}\) yields a \textbf{BAP item} \((\mathcal{H}^{t-1}, S^{t-1}, B^{t}, D^{t}, A^{t}, S^{t})\) that consists of the \textbf{game history} \(\mathcal{H}^{t-1}\) that culminated in the \textbf{previous built structure} \(S^{t-1}\), the Builder's position and orientation, 
\(B^{t}\), the \textbf{new dialogue} \(D^{t}\), and the \textbf{new action sequence} \(A^{t}\) that resulted in the \textbf{new structure} \(S^{t}\). Given \(\mathcal{H}^{t-1}\), \(S^{t-1},\) and \(D^{t}\), BAP models \(M\) should predict an action sequence \(A^{t}_{M}\)  that yields \(S^{t}\). 

Figure~\ref{fig:bap_examples} illustrates how the game snippet shown in Figure~\ref{fig:sequence} corresponds to three different items for the BAP task.
The game starts with an empty board \(S^{0}\), and an initial dialogue between the two players \(D^{1}\), and the first BAP item  consists of predicting that first block placement \(A^{1}\) (and consequent structure \(S^{1}\)) (Figure~\ref{fig:bap_examples}, Item 1). Then, another instruction follows, which the Builder executes by placing another block, resulting in the next structure to be predicted (Figure~\ref{fig:bap_examples}, Item 2), and the third instruction yields a complex action sequence that results in the desired v-shape structure (Figure~\ref{fig:bap_examples}, Item 3). 

\subsubsection*{How to interpret BAP item figures in this paper}
The leftmost columns in each BAP item figure (e.g. Figure~\ref{fig:bap_examples}) display the (truncated) game history (\(S^{t-2}\), \(D^{t-1}\), \(S^{t-1}\), and \(D^{t}\)), while the rightmost column shows the resulting built structure \(S^{t}\). Columns are read top to bottom and left to right. We use \(S_{H}\) instead of \(S\) to indicate that the structure was built by a human (\(H\)), not a model). To highlight the effect of the final action sequence, previously placed blocks are shown in lighter colors.
Structures like $S^{t-1}$ are rotated to align with the \B's POV $B^t$ (corresponding to the next turn and dialogue segment $D^t$ below it). 
For \B's POV, we only use the yaw angle, and \B's exact position and pitch angle are omitted for clarity. Any temporary supporting blocks are not shown as well.


\subsection{The original BAP dataset}\label{sec:dataaug}
The training, test and development splits of the MDC contain 3709, 1616, and 1331 BAP items respectively, and the average sequence length of an action sequence (in the development set) is 4.3 (with a standard deviation of 4.5). Target structures in the test data do not appear in the training or development data.
Since the small size of the training set is a major limiting factor for training complex models, in \citet{jayannavar-etal-2020-learning}, we generated synthetic variants of the original game logs in the training data by combining three simple data augmentation techniques: utterance paraphrasing (synonym-based substitutions), color substitution (random color permutations across logs), and spatial transformations (rotating structures and Builder \B). Empirically, we found that increasing the training data to 14,836 (4x) items gave the best performance for the GRU-based baseline model (Section~\ref{sec:baseline_model_desc}). Henceforth, we will refer to this combination of the augmented training data and the original test and development data as the original BAP data.

\subsection{Formalizing the BAP task}
\label{sec:Formalizing-BAP}

\paragraph{The built structure}
In Minecraft, blocks can be placed into the \textbf{cells}  \(\cell = \loc\) of a discrete 3D grid if \(\cell\) is empty and either on the ground or adjacent to any cell \(\cell'\) that contains a block. A cell \(\cell = \loc\) is on the ground if its height \(y=1\), and is adjacent to any cell  \(\cell' = \locPrime\)  if they touch on one side.\footnote{Cells \(\loc\) and \(\locPrime\) are adjacent if either \(x=x'\pm1, y=y'\), and \(z=z'\), or \(x=x', y=y'\pm1\), and \( z=z'\), or  \(x=x', y=y'\), and \( z=z'\pm1\)} 
In the MCBT, there are six different \textsc{BlockColors}, and Builders place blocks inside a predefined \textbf{build region} \(\BR\), an \(11 \times 9 \times 11\)  box that contains all grid cells \(\cell = \loc\)  from \(\locMin\) to \(\locMax\).\footnote{In the  Minecraft Dialogue Corpus,  the \(\BR\)'s horizontal coordinates \(x\) and \(z\) range from \(x_{\min}=z_{\min} = -5\) to \(x_{\max} = z_{\max} = +5\) and vertical coordinates \(y\) range from \(y_{\min}= 1\) to \(y_{\max} = 9\).} 

The current \textbf{Built structure} \(S\)  can be represented as a list of blocks and their locations:

\begin{align*}
  \BR & =_{\textit{def}} \{\cell = \langle x,y,z\rangle \mid x_{\min} \leq x \leq x_{\max},  y_{\min} \leq y \leq y_{\max}, z_{\min} \leq z \leq z_{\max} \} \\
    \Colors &=_{\textit{def}}  \{\red, \orange,\yellow,\green,\blue,\purple\}\\
    S&  =_{\textit{def}}  \{(\cell, \mathsf{c}) \mid \cell \in \BR, \mathsf{c}\in\Colors \}
  \end{align*}

\paragraph{The Builder}
\label{sec:bpos}
The Builder can move freely around the environment. Since spatial relations in the instructions (\exampleutt{the block in front of you}, \exampleutt{to your left}) often depend  on the Builder's current location and viewpoint, we record the \textbf{position and pose/orientation of the Builder agent \(\textsc{Pos}_\B = (\locB, \phi_\B, \gamma_\B)\)}, where \(\locB\) is a (real-valued) location  (that may or may not be inside the build region), \(\phi_\B \in [-90, ..., +90] \) indicates a pitch (vertical rotation)  and \(\gamma_\B \in  [-180, ..., +180]\) a yaw (horizontal orientation) angle.
For brevity, we use \(B^{t}\) to refer to \(\textsc{Pos}_\B\) at time \(t\) in the formal definition of BAP items.

\paragraph{The Builder's actions}
We specify Builder \textbf{actions}  \(a = (t, c, \cell)\) as triplets consisting of an action type \(t \in \mathcal{T} = \{\textsc{Place}, \textsc{Remove}\}\), a block color \(c \in \Colors\), and a grid cell \(\cell = \loc \in \BR\) whose state is changed by \(a\).\footnote{Since the Malmo API can be used to move agents into positions in which a particular block is accessible to them, we  ignore the Builder's movement, and only consider its block placement and removal actions.} 
If a feasible action \(a\) is executed, it changes the built structure  \(S\) to \(S'\neq S\):  \(S~\xrightarrow{a}~S'\). A placement is feasible if \(\cell\) is empty  and either on the ground  or adjacent to a non-empty cell, and results the addition of a  \(c\)-colored block located in \(\cell\) to \(S'\). A removal action is feasible if \(S\) contains a contains a block of color \(c\) in \(\cell\), and results the exclusion of that block from \(S'\).
When Builders follow an instruction, they execute an \textbf{action \textit{sequence}} \(A = \langle a^{1},..., a^{k}\rangle\) for \(k\geq 1\) and \(a^{1},...,a^{k} \in \actionspace\) that changes  \(S\) to \(S'\): 
\(
S \xrightarrow{A = \langle a^{1},...a^{k}\rangle} S'
\).
\(\actionspace\) is the set of all 7623 possible actions (7 actions available per cell in the  \(11 \times 9 \times 11\) build region—placing a block in one of 6 colors or removing a block).

\subsection{Evaluating BAP models}
\label{ssec:background_eval}
To compare two action sequences, or to measure the change from \(S\) to \(S'\), it is helpful to note that  any action \(a = (t,c,\cell)\) is undone by the \textbf{inverse action} \(a^{-} = (t^{-},c,\cell)\) where \(t^{-} = \textsc{Remove} \) if \(t = \textsc{Place} \), and \(t^{-} = \textsc{Place} \) if \(t = \textsc{Remove} \). 
Most commonly, this occurs when blocks are placed to serve as a necessary supporting block for a floating structure and removed in the same action sequence. These supporting blocks can be ignored since they do not occur in \(S'\). Moreover, builders are free to use any color they wish, and can often choose among a variety of possible locations where to place these blocks. Human Builders are also prone to accidentally placing or removing blocks, but typically recognize and correct such mistakes immediately within the same action sequence.  Finally, many structures do not require their blocks to be placed in a particular order. 
We can therefore transform any (feasible) action \textit{sequence} \(A\) into a \textbf{\textit{set} of net actions} \(A^{\mathrm{net}}\) by first removing from \(A\) any actions \(a\) and their inverse \(a^{-}\)  and turning the resulting (infeasible) sequence \(A'\) into a set \(A^{\mathrm{net}}\). 
The \textbf{distance} \(\Delta(S,S')\) between two structures \(S\) and \(S'\) can then be defined as the number of net actions of any action sequence \(A: S \xrightarrow{A} S'\) that changes \(S\) to \(S'\): 
\(\Delta(S,S') = \vert A^{\mathrm{net}}\vert\), and any two action sequences  \(A_1\) and \(A_2\) are \textbf{equivalent} (\(A_1 \equiv A_2\)), i.e. lead to the same subsequent structure, if they have the same set of net actions (\(A_1^{\mathrm{net}} = A_2^{\mathrm{net}}\)). 
We therefore evaluate BAP models against human Builders by comparing their respective net actions, \(A_{M}^{\textrm{net}}\) and \(A_{H}^{\textrm{net}}\) to compute a (strict) F1 score (we report a \textbf{micro-averaged} F1 score over all action sequences in the test/dev data):

\begin{definition}
Given a BAP item consisting of a (human) reference action sequence \(A_{H} = \langle a_{H}^{1},...,a_{H}^{k}\rangle\) that leads from \(S\) to \(S_{H}\), corresponding to a reference net action set \(A_{H}^{\textrm{net}}\),  and a (model) predicted action sequence \(A_{M} = \langle a_{M}^{1},...,a_{M}^{l}\rangle\) that leads from \(S\) to \(S_{M}\), corresponding to a predicted net action set  \(A_{M}^{\textrm{net}}\), \textbf{\textit{strict} Precision, Recall and F1} scores assume that a Builder action \(a_{M}^{m} = (t,c,\loc) \in A_{M}^{\textrm{net}}\) is correct if and only if there is an equal reference action \(a_{H}^{h} = (t,c,\loc) \in A_{H}^{\textrm{net}}\):
\begin{align*}
\textbf{Strict Precision}~P_{\mathrm{strict}}(A_{M}^{\textrm{net}}, A_{H}^{\textrm{net}}) & = \frac{\vert(A_{M}^{\textrm{net}} \cap A_{H}^{\textrm{net}})\vert}{\vert A_{M}^{\textrm{net}} \vert}\\
\textbf{Strict Recall}~R_{\mathrm{strict}}(A_{M}^{\textrm{net}}, A_{H}^{\textrm{net}}) &  = \frac{\vert(A_{M}^{\textrm{net}} \cap A_{H}^{\textrm{net}})\vert}{\vert A_{H}^{\textrm{net}}\vert}\\
\textbf{Strict F1}~F1_{\mathrm{strict}}(A_{M}^{\textrm{net}}, A_{H}^{\textrm{net}}) &  = \frac{2\cdot P_{\mathrm{strict}}\cdot R_{\mathrm{strict}}}{P_{\mathrm{strict}}+R_{\mathrm{strict}}}
\end{align*}
\end{definition}

%% file: models.tex
\subsection{A CNN- and GRU-based baseline model}
\label{sec:baseline_model_desc}

In \citet{jayannavar-etal-2020-learning}, we proposed a neural model for the BAP task. In this work, we build upon it and refer to it as \textbf{the GRU-based baseline model}.
The model is based on a recurrent encoder-decoder architecture~\citep{sutskever2014sequence,cho-etal-2014-learning} in which a GRU-based encoder captures the game context (dialogue and action history) via GloVe embeddings \citep{pennington2014glove}, and a CNN-based encoder captures the world state at each time step. The world state encompasses the current built structure, action history and the Builder's position and orientation.
We trained this model with teacher forcing and cross entropy loss on the original BAP data, and used greedy decoding (max. sequence length of 10 actions) at test time. This yields an F1 score of 21.2\% on the BAP test set.

\subsection{Related Work: LLMs for BAP}
\label{sec:rw_llms_bap}

Related to our work are the recent works of \citet{chaturvedi-etal-2024-nebula} and \citet{ch-etal-2024-retrieval}, both exploring more modern LLMs for the BAP task.
The former finetunes the Llama-3-8B model \citep{dubey2024llama} using QLoRA \citep{dettmers2023qlora} on the original BAP data, resulting in the Nebula model, which predicts the next action sequence in a text-to-text fashion using all preceding conversation and action sequences as context. The latter employs in-context learning (ICL) with GPT-4.
Both achieve state-of-the-art results over the GRU-based model.
Nebula achieves an F1 score of 39.2\% and GPT-4 scores 39.0\% (under the now legacy BAP evaluation), demonstrating that both approaches yield very similar top-line performance (Nebula being marginally better).
A slightly more detailed comparison to \citet{chaturvedi-etal-2024-nebula} is provided later in Section~\ref{sec:rw_bap_eval} given some other relevant aspects of their work.

\input{rw_task_dataset}

%% file: rw_task_dataset.tex
\subsection{Related Work: Tasks and Datasets}
\label{sec:rw_task_and_data}

In this section, we explain why the BAP task continues to be a relevant and challenging testbed for grounded instruction following.
The MDC dataset is a collaborative embodied task, while BAP focuses on grounded language instruction following. 
General collaborative task-oriented dialogue tends to focus on un-embodied domains. Typical of the domain is the MultiWoz dataset \citep{budzianowski2018multiwoz} which explores conversations around tasks such as trip planning, which require both dialogue and information retrieval. Even tasks that include a visual modality are not explicitly embodied: take \citet{Udagawa_Aizawa_2019}'s OneCommon task, which requires two agents to work together to identify the overlap in their fields of view, yet does not include turning to adjust a perspective, etc.

The MDC, on the other hand, is closer to some vision-and-language works. Vision-and-Language Navigation (VLN) \citep{anderson2018vision}, and its dialogue counterpart, Cooperative Vision-and-Dialog Navigation (CVDN) \citep{thomason:arxiv19}, which explore instruction-following in photorealistic navigation settings, are good examples. Within this domain, others have combined vision and dialogue navigation with task completion, including CEREALBAR \citep{suhr2019executing}, a game task in which an instruction-follower and an instruction-giver cooperate to traverse a map and collect items. Others have combined vision and language-based task completion in ways that go beyond navigation. \citet{roy2021dialogueobjectsearch} consider a human and a robot collaborating to search for an object, while ALFRED \citep{Shridhar_2020_CVPR} is a benchmark for mapping natural language instructions and egocentric vision to action sequences for household tasks in simulated environments. \citet{Gao2022DialFREDDA} extend this by incorporating dialogue instead of just instructions.
Many of the challenging aspects of the MDC are shared across these tasks, e.g. asking for clarification or help is a central component in VLN works \cite{Nguyen2019HelpAV, chi2019justaskaninteractivelearning}. Central to all of these tasks is that agents and humans must communicate with natural language about a world that exists outside of text. 
Although more recent tasks require real vision to represent this world, their underlying world state space (defined by fixed viewpoints and underlying navigation graphs) is just as highly discretized as a Minecraft grid. And, while many works do opt for visual inputs, \citet{jia-etal-2024-langsuit} explore the potential of leaving vision out of task-oriented dialogue, and works in less grounded settings, such as \citet{xu-etal-2024-rethinking}, often utilize fully textual agents. Despite not requiring vision as most of the counterparts discussed above do, our task poses a possibly more challenging planning problem: its action space is much larger (7623 possible actions vs. six actions in the vision-language navigation work). By excluding vision, MDC can focus solely on instruction following and the decision-making inherent in that. 
Furthermore, Minecraft dialogues are more complex than the other discussed datasets due to the MDC's asynchronous communication, longer turn sequences, and dynamic environment changes during construction. Unlike navigation or simple cooperative games, where referring expressions target existing objects, construction often involves instructions for objects yet to be built.

The  TEACh benchmark \citep{Padmakumar_Thomason_Shrivastava_Lange_Narayan-Chen_Gella_Piramuthu_Tur_Hakkani-Tur_2022} addresses the former of these limitations with free-form dialogue, similar to the MDC, but not the latter as the task does not require references to unbuilt objects. While it does feature object state changes in its 3000+ human-human dialogues, the instruction giver also receives a predefined plan of steps, limiting the diversity since there is only one way to proceed (compared to the many unique ways to build a final MDC structure). Additionally, the MDC allows for greater abstraction in instructions, such as referring to substructures, individual blocks, or shapes, whereas TEACh focuses on well-defined objects in the environment.  This combination of dialogue structure, references to unbuilt objects, plan diversity, and the instruction abstraction make the MDC and BAP tasks both difficult and compelling, as well as providing areas of exploration not addressed by other similar tasks.

\subsubsection*{Minecraft Environment} Minecraft has garnered a lot of interest as an AI experimentation platform in recent years. Our previous work \citep{narayan-chen-etal-2019-collaborative} was among the first to use it to study grounded task-oriented dialogue, and introduced the MCBT and MDC.
Facebook’s CraftAssist \citep{DBLP:journals/corr/abs-1907-08584,DBLP:journals/corr/abs-1905-01978,DBLP:journals/corr/abs-1907-09273} is another such example, enabling two-way human-bot interactions where a human architect instructs an automated builder to build complex structures. Their data includes synthetically generated and crowdsourced instructions paired with logical tree structures consisting of action primitives, unlike the MDC, which features human-human dialogues with greater ambiguity, variety, and noisier Builder actions.
MineDojo \citep{NEURIPS2022_74a67268} focuses on creating versatile agents for diverse tasks via an internet-scale knowledge base, contrasting with tasks like IGLU \citep{mohanty2024idat, pmlr-v220-kiseleva23a, pmlr-v176-kiseleva22a} and MCBT, which prioritize grounded natural language dialogue and clarification for interactive agents. 

Among Minecraft-related efforts, IGLU is most closely related to the MDC/MCBT and was directly inspired by it. Unlike the full dialogue structure of the MDC, the IGLU dataset contains mostly single turns (i.e. instruction-action pairs without a broader game context or history). A small multi-turn subset is available (127 dialogues), but has strict turn taking in place of MDC's free-form asynchronous dialogue. It further simplifies the task by using cardinal directions (e.g \exampleutt{North}, \exampleutt{South}) which are absolute and more consistent that relative references (e.g. \exampleutt{left}, \exampleutt{right}).

Some other works also build on or are inspired by the MCBT/MDC: \citet{bonn-etal-2020-spatial}, \citet{thompson-etal-2024-discourse}, \citet{bonial-etal-2021-builder}, and \citet{madge2025mdcrminecraftdialoguecorpus} provide additional linguistic annotations for the MDC, while \citet{shi-etal-2022-learning} explore when the Builder should ask for clarification or execute an instruction, although their building task consists of predicting single actions instead of action sequences, and is therefore not directly comparable to ours.

%% file: eval.tex
\section{Revisiting BAP Evaluation: A Cleaner Test Set}
\label{sec:eval}

\begin{figure*}[!htbp]
 \centering
 \begin{tabular}{l}
\toprule
 \textbf{Suitable BAP test items} (with correct structures)\\
 \midrule
 \textbf{Clear context with unique interpretation (non-empty board)}\\
 \includegraphics[width=.9\textwidth]{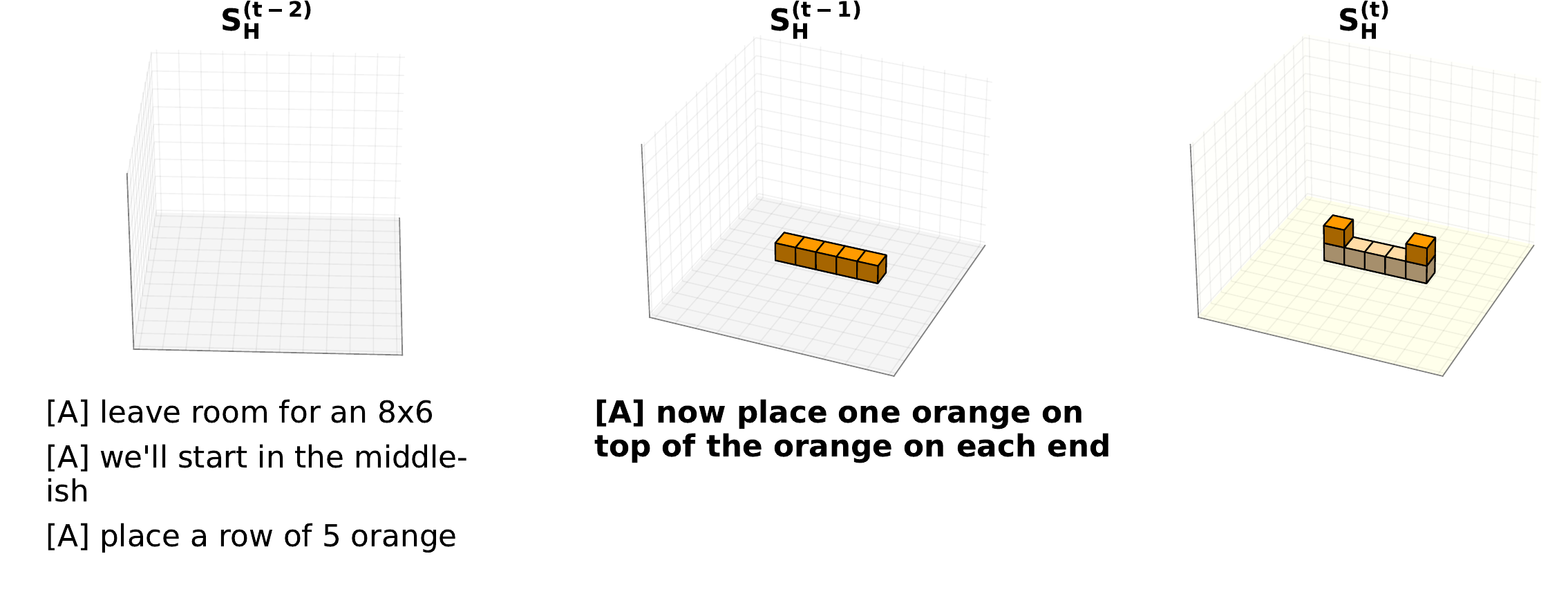}\\
 \midrule
 \textbf{Clear context with unique interpretation (empty board)}\\
 \includegraphics[width=.9\textwidth]{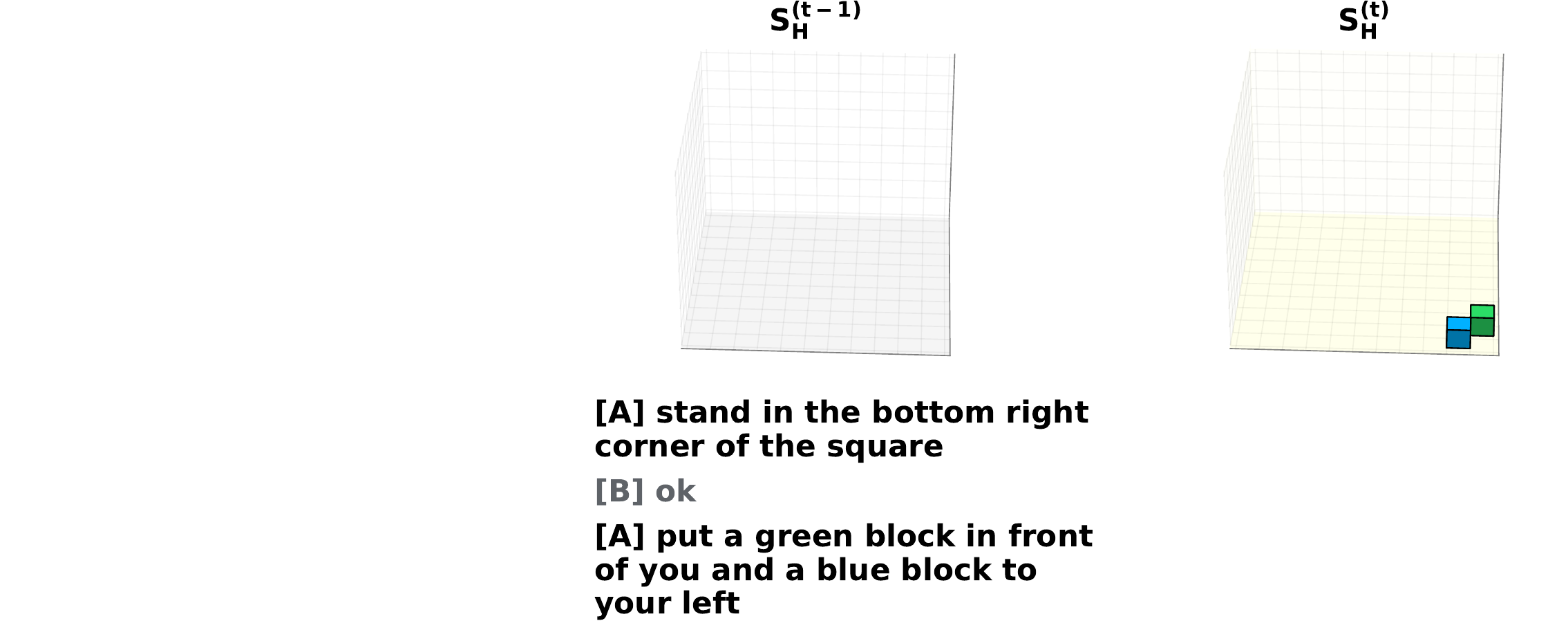}\\
 \midrule
 \textbf{Clear context with multiple interpretations (empty board)}\\
 \includegraphics[width=.9\textwidth]{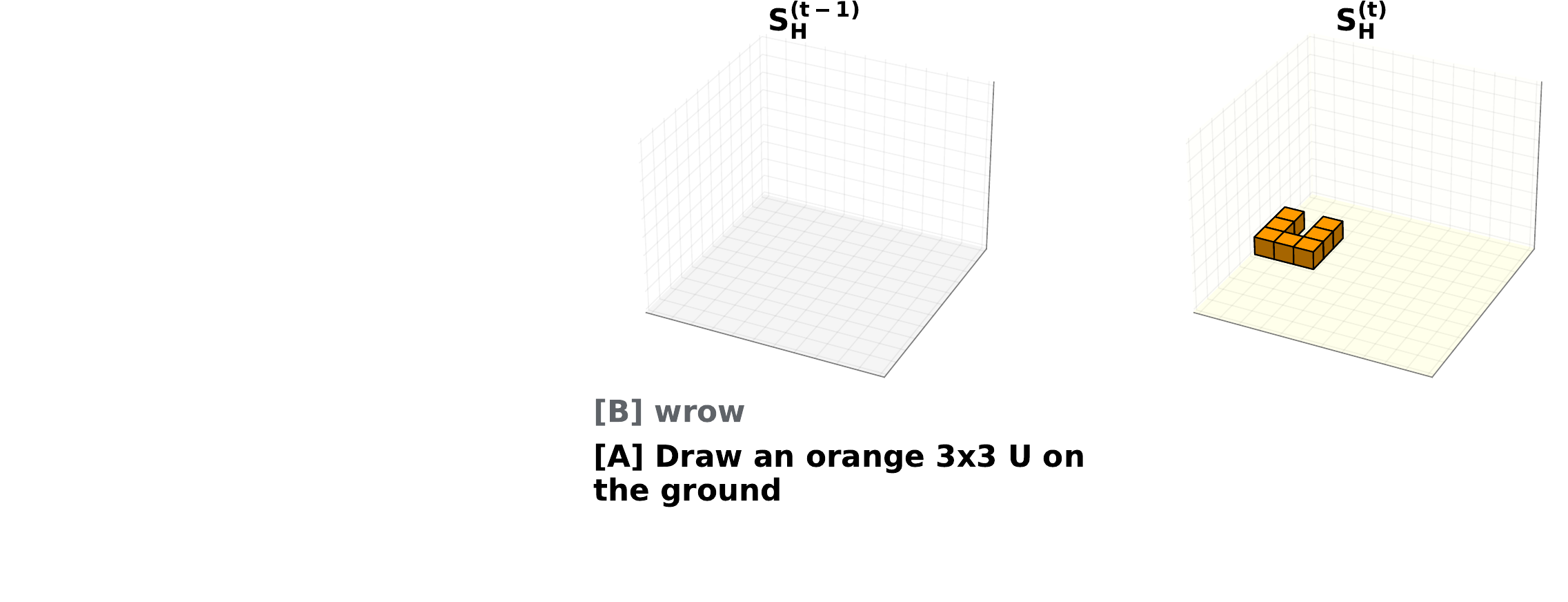}\fignegspace\\
 \bottomrule
 \end{tabular}
     \caption{BAP items with clear contexts and correct structures are suitable for evaluation purposes. If the board is empty, there may be multiple valid interpretations.}
     \label{fig:UsableBAPitems}
 \end{figure*}

\begin{figure}[!htbp]
\centering
\begin{tabular}{l}
\toprule
\textbf{Unsuitable BAP test items: Unclear contexts}\\
\midrule
\includegraphics[width=.9\textwidth]{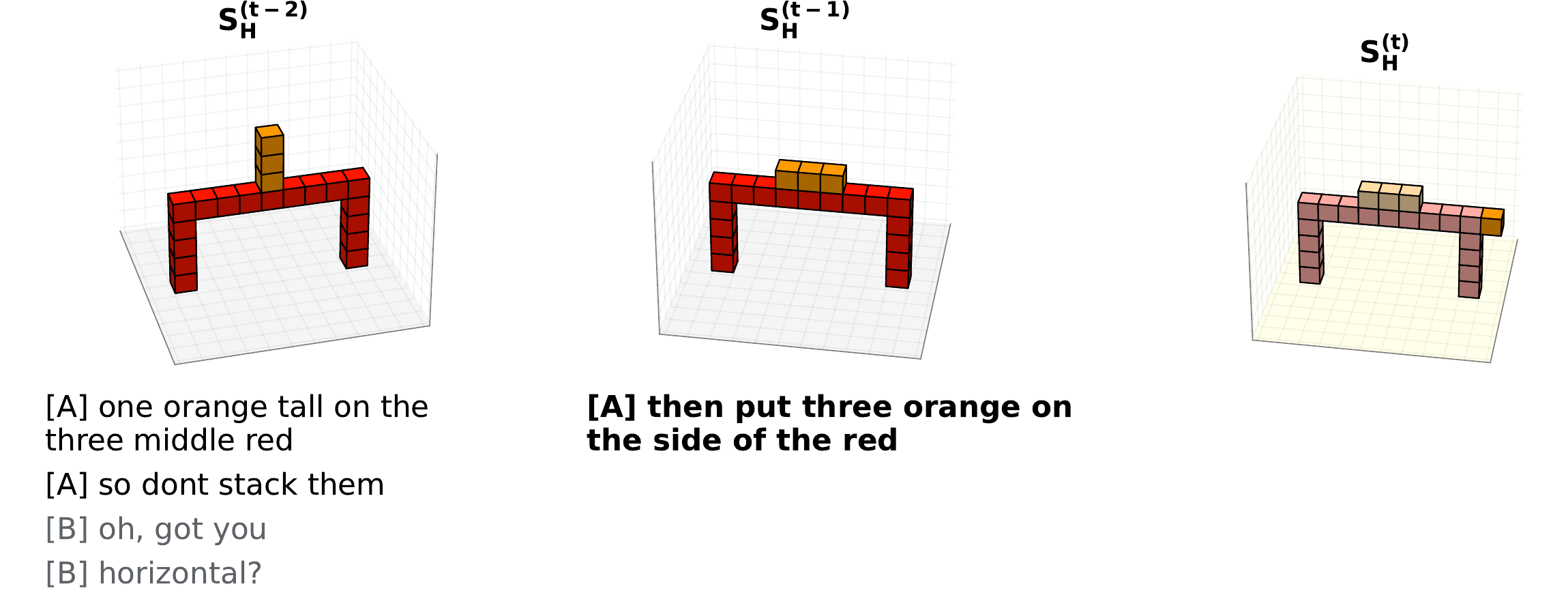}
\\ 

\midrule
\includegraphics[width=.9\textwidth]{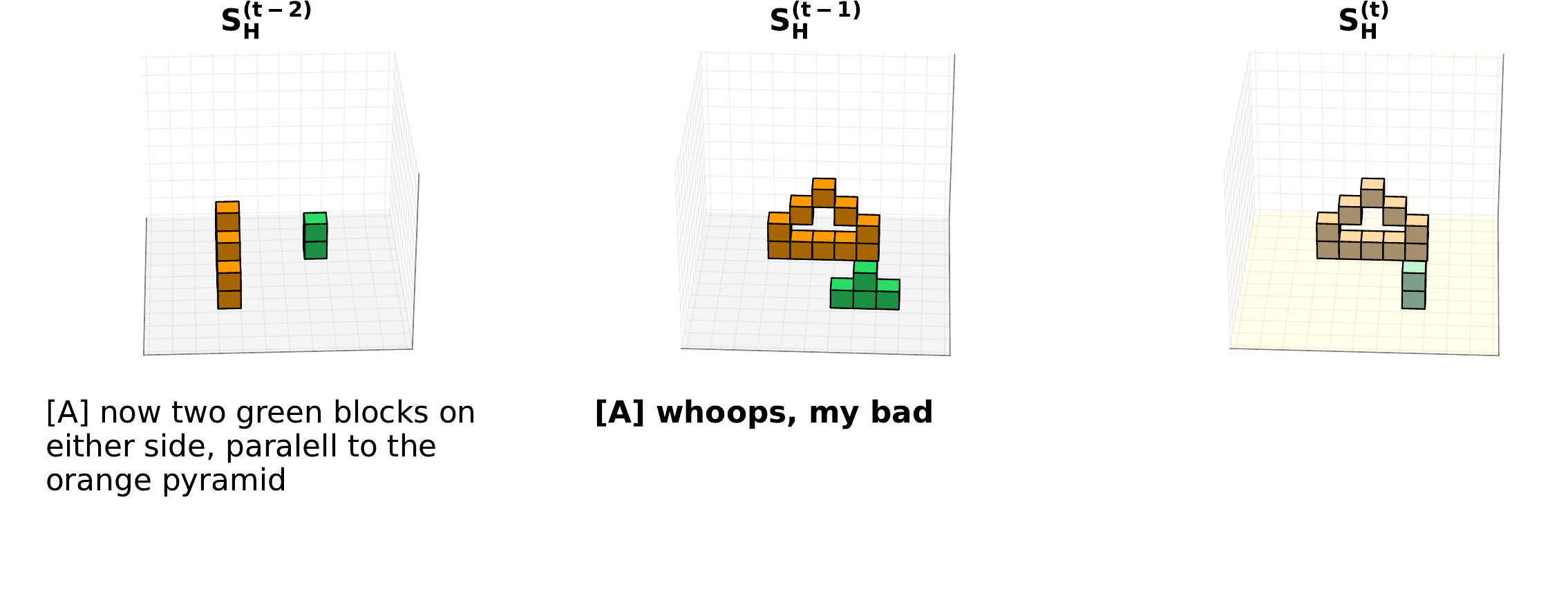}\fignegspace
\\
\bottomrule
\end{tabular}
\caption{BAP items with unclear contexts are unsuitable for evaluation purposes}
\label{fig:UnusableBAPitemsUnclearContexts}
\end{figure}
\begin{figure}[!htbp]
\centering
\begin{tabular}{l}
\toprule
\textbf{Unsuitable BAP test items: Incorrect structures}\\
\midrule
\includegraphics[width=.9\textwidth]{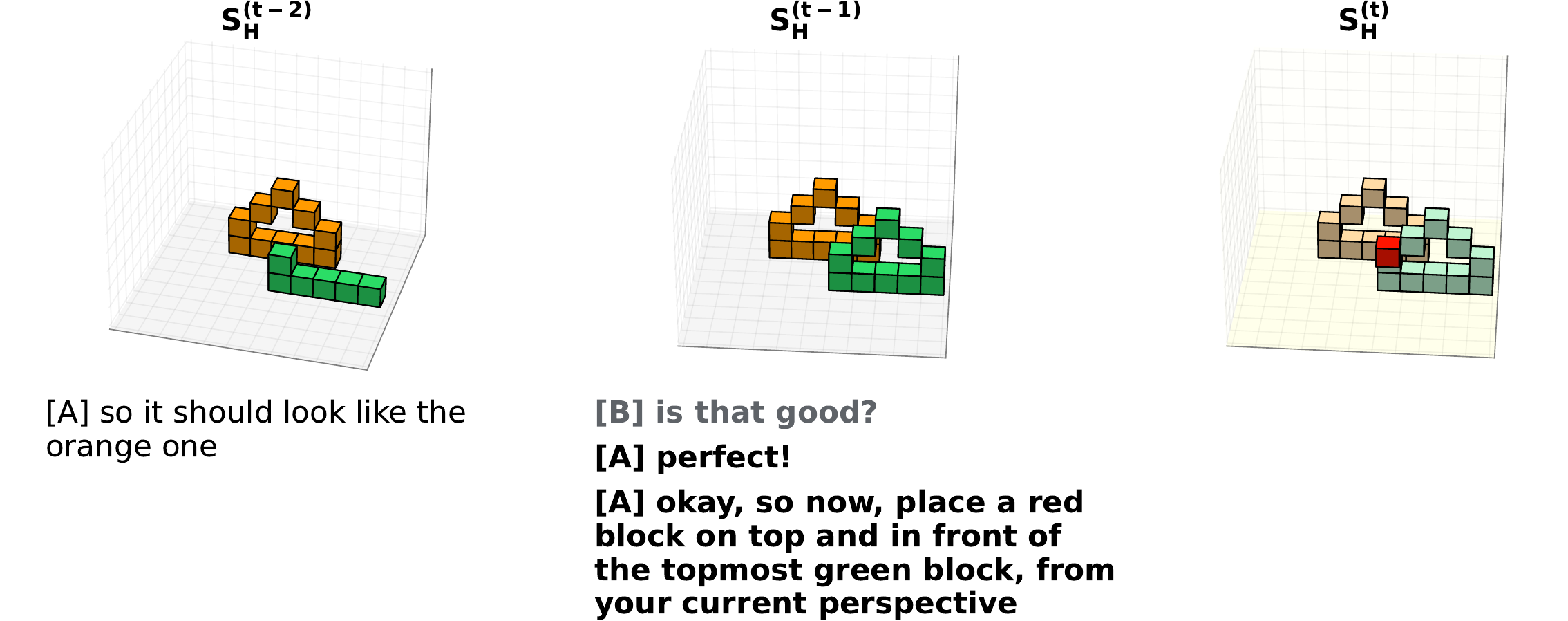}
\\
\midrule
\includegraphics[width=.9\textwidth]{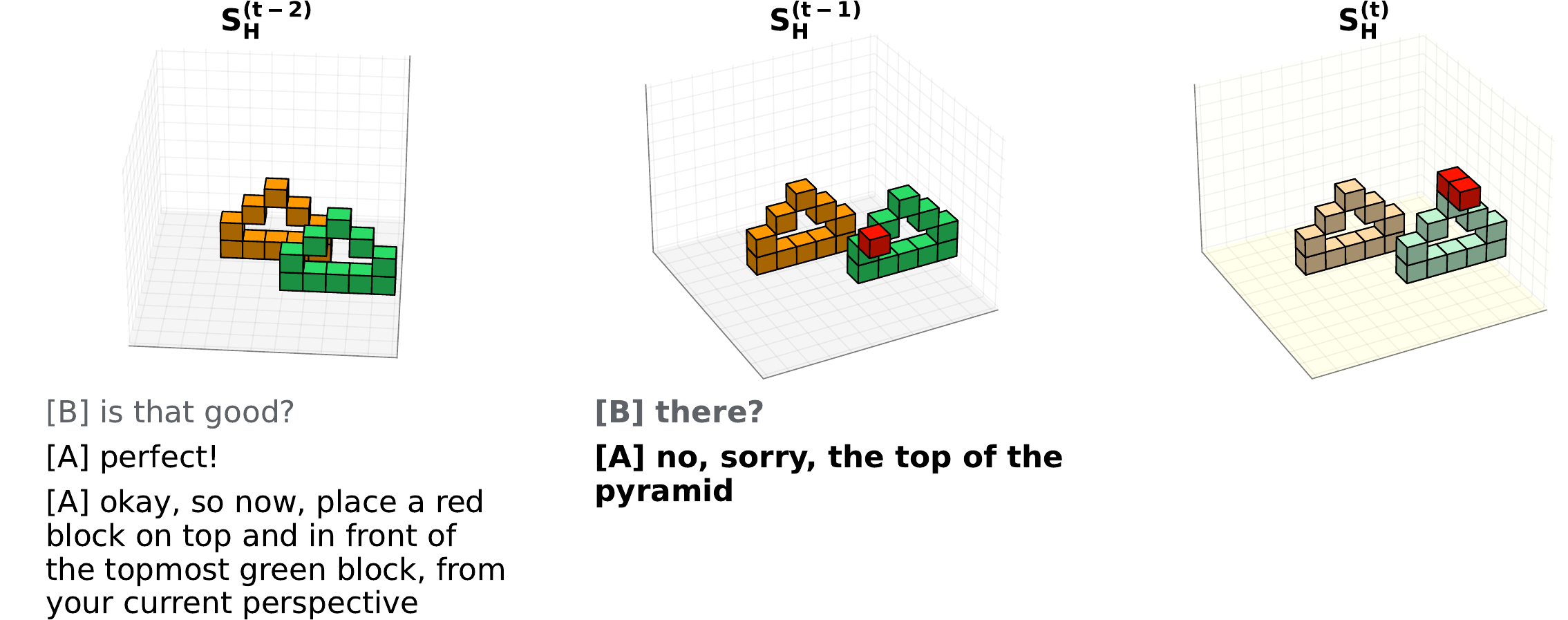}\\
\bottomrule
\end{tabular}
\caption{BAP items with incorrect structures are unsuitable for evaluation purposes.}
\label{fig:UnusableBAPitemsIncorrectStructs}
\end{figure}
\begin{figure}[!htbp]
    \centering   
      \begin{tabular}{ll}
        \toprule
        \multicolumn{2}{l}{  \textbf{Fixable BAP items with incorrect structures }}\\
        \midrule
        \multicolumn{2}{l}{ \textbf{The structure of item $i$ can be replaced by the structure of item $i+1$}}\figposspace\\
         \textbf{Item $i$} \vspace{-1.1pc}\\
         & \includegraphics[width=.6\textwidth]{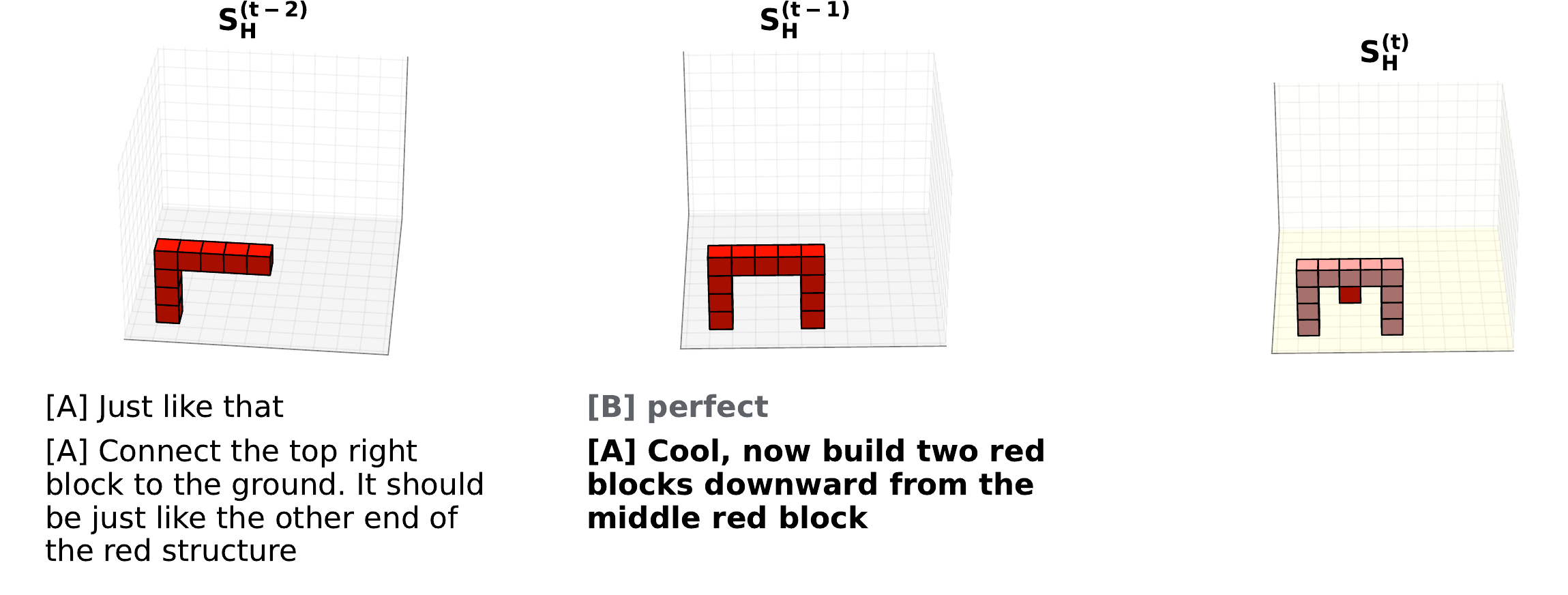}\\
         \textbf{Item $i+1$}  \vspace{-1.1pc}\\     
         & \includegraphics[width=.6\textwidth]{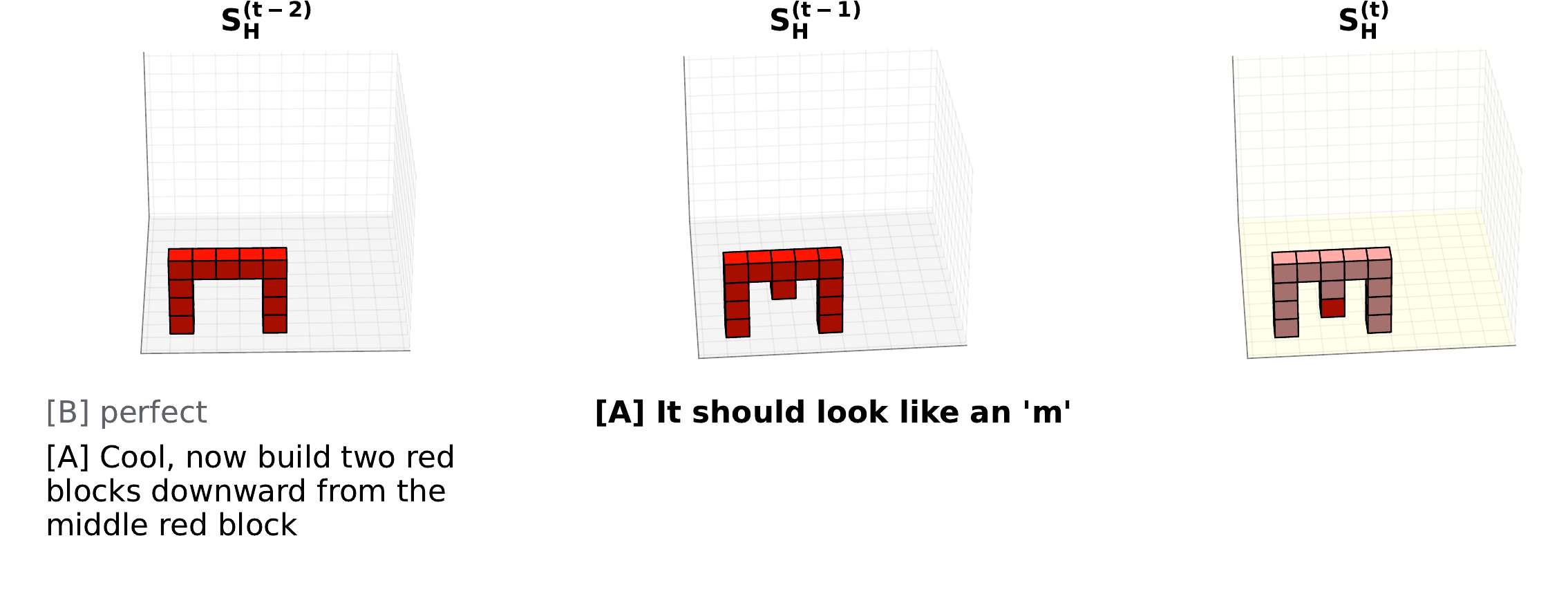}
\\
\multicolumn{2}{l}{     \textbf{The structures of item $i$ and $i+1$ can be replaced by the structure of item $i+2$}}\figposspace\\
   \textbf{Item $i$} \vspace{-1.1pc}\\  &\includegraphics[width=.6\textwidth]{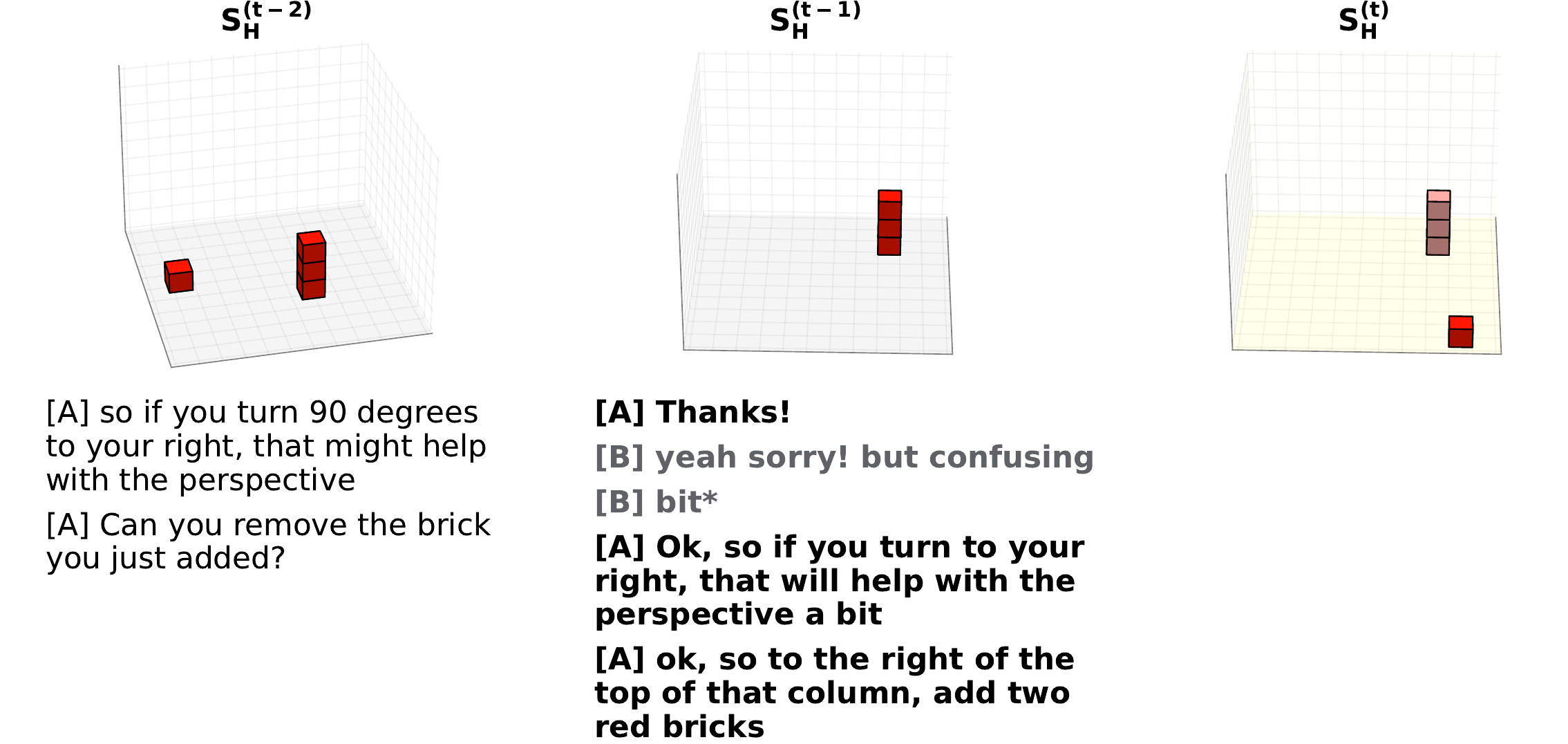}\\
   \textbf{Item $i+1$}  \vspace{-1.1pc}\\  &\includegraphics[width=.6\textwidth]{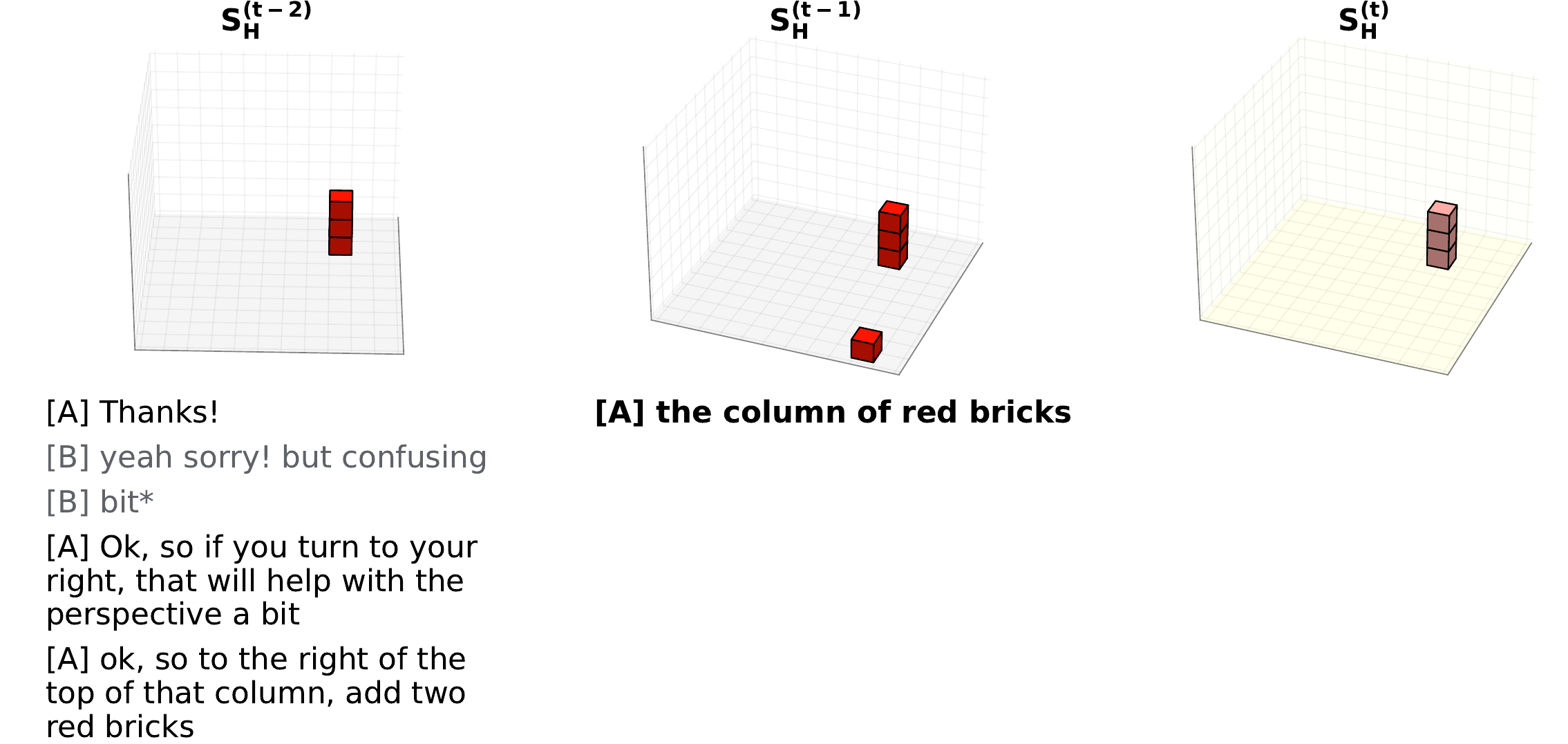}\\ 
   \textbf{Item $i+2$}  \vspace{-1.1pc}\\ & \includegraphics[width=.6\textwidth]{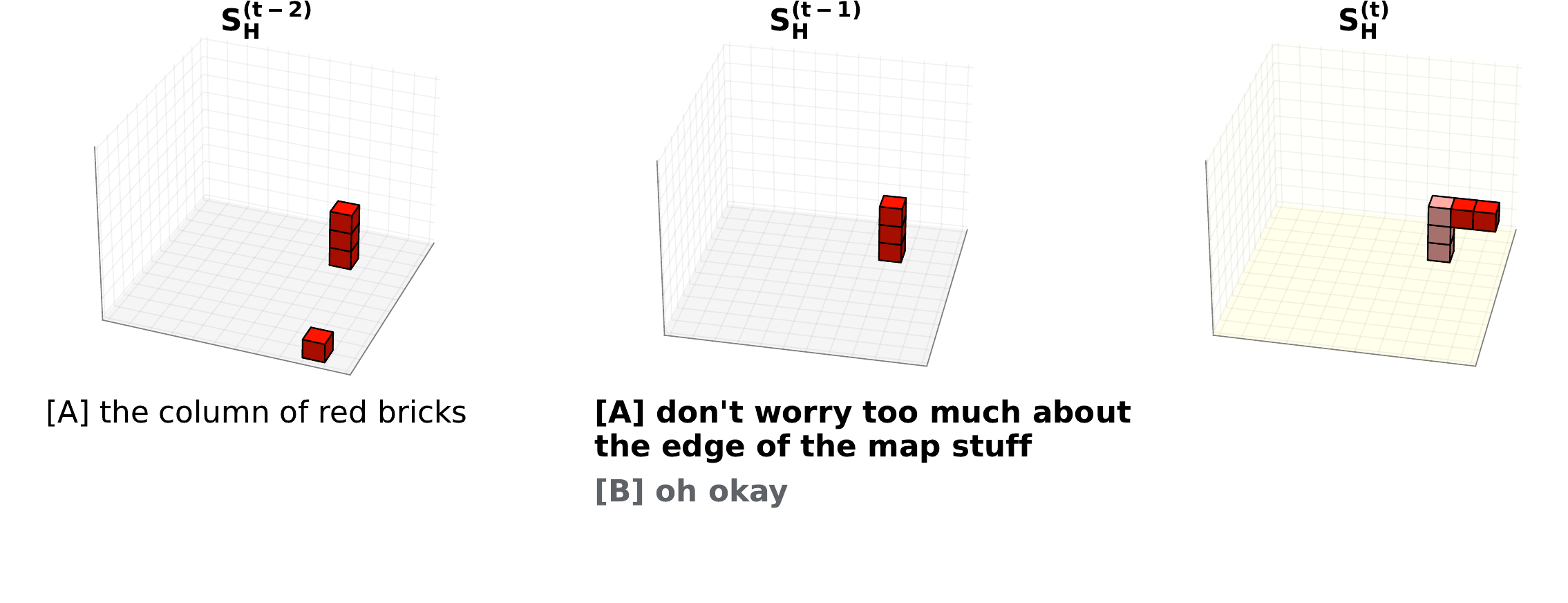}
 
   \\
    \bottomrule
    \end{tabular}         
    \caption{A correct structure following one or more BAP items with incorrect built structures can be used to fix the incorrect BAP items.}
    \label{fig:fixable}
\end{figure}

Since the MDC consists of real human-human game logs collected in a low-stakes, free-flowing setting with minimal constraints on the participants’ language and actions, rather than standalone instructions, not all action sequences correspond to BAP items that are suitable for evaluation purposes. First, Builders might move blocks even when no new instructions were given, or before it is completely clear what they should be building next (e.g. when the Architect only instructs the Builder to \exampleutt{build a rectangle to the right of that}, without specifying the size or color of the rectangle, let alone its orientation and distance to \exampleutt{that}). In such cases, the corresponding BAP items have an \textbf{unclear context}. Second, Builders occasionally make mistakes, and are frequently interrupted by the Architect, who may want to provide more information or simply affirm that the Builder's actions are correct. In those cases, the BAP item's action sequence may lead to an \textbf{incorrect structure}. Finally, when the board is empty, we observe that Architects do not always identify a specific location or orientation for the blocks to be placed, even when they clearly describe the structure to be built (e.g., the instruction \exampleutt{place a red block on the ground}). This likely occurs because there are no blocks on the board or landmarks in the environment that could serve as spatial anchor to be mentioned, and because Architects are aware that the game only requires the Builder's structure to match the Target as long as it appears anywhere inside the Build Region  and in any orientation. In those situations, two different Builders might start the game by placing the same structure in two different locations and/or orientations and both might be correct, since the context provides \textbf{multiple (valid) interpretations}. This phenomenon arises almost exclusively with an \textbf{empty board}, since the blocks on a \textbf{non-empty board} and the current position of the Builder typically provide enough constraints for a clear instruction to only have a unique (valid interpretation) in that case. In this section, we perform an in-depth analysis of the items in the BAP test set that identifies these phenomena, and  introduce a new, cleaner BAP test set that should be used to evaluate BAP models instead of the original, legacy test set. In the next section, we revisit the strict F1 metric, and define a fairer variant that accounts for the fact that some empty-board test items have multiple valid interpretations.

\subsection{Identifying suitable BAP test items}
\label{sec:annotations}
To distinguish the cases mentioned above, we manually annotate all BAP items in the test set (for details, see Appendix Section~\ref{sec:app_ann_details}). For each \textbf{non-empty board (NEB)} item, annotators were  asked whether the \textbf{context is clear and has a unique interpretation} (i.e. it is clear what structure to be built next, and where to place it on the board), or whether the \textbf{context is unclear} (i.e. it is unclear what to build  next, e.g. because more information is required, or no new instruction was given). For \textbf{empty board (EB)} items, annotators could also indicate that the \textbf{context is clear and has multiple interpretations}, (i.e. it is clear what structure should be built, even though there are multiple possible placements or orientations of that structure on the empty board.) If the context is clear, a second question asked  whether the built structure is \textbf{correct}  or \textbf{incorrect}.

\paragraph{\textbf{Suitable BAP test items}}
Items whose contexts are clear and whose structures are correct can clearly be used for evaluation purposes (Figure~\ref{fig:UsableBAPitems}). If the board is non-empty, this requires contexts to have a unique interpretation, but if the board is empty, Builders can chose any of typically many  correct interpretations.

\paragraph{\textbf{Unsuitable BAP test items}}
Items are unsuitable for evaluation if the context is unclear (e.g. because the Architect gave unclear instructions, Figure~\ref{fig:UnusableBAPitemsUnclearContexts}) or if the structure is incorrect (e.g. the  Builder made mistakes, Figure~\ref{fig:UnusableBAPitemsIncorrectStructs}). 
\paragraph{\textbf{Fixable BAP test items}}
Manual analysis shows that sequences of one or more incorrect built structures that are immediately followed by a correct structure can be fixed  by replacing the incorrect structures with the built structure of that subsequent correct item (Figure~\ref{fig:fixable}).
Such cases occur typically because the Builder's (correct) action sequence was interrupted by an utterance, or because the Architect issued a quick correction that the Builder immediately executed.

\subsection{Statistics of the clean BAP v2 test set}
 To obtain our final evaluation set, we start with the original BAP test set of 1,616 BAP items, but correct the structures of 149 items that we were able to identify and fix automatically, and remove 461 items that are unsuitable for evaluation purposes (418 with unclear contexts and 43 with incorrect structures that could not be fixed). This results in a new, \textbf{BAP v2 test set} that consists of \textbf{1,155 BAP items}, all of which have a \textbf{clear context and a correct built structure}. 1,071 of these items have a unique correct interpretation, and 84 of the empty board (EB) items have multiple correct interpretations.

\section{Revisiting BAP Evaluation: Metrics}
\label{sec:eval_metrics}
We now revisit the strict F1 metric and propose a fairer variant that accounts for the fact that some EB items have multiple valid interpretations. Additionally, the current evaluation, relying solely on an aggregated F1 score, lacks detailed insights into model behavior. To address this, we introduce additional metrics. These contributions collectively define an updated \textbf{BAP v2 evaluation benchmark}.

\subsection{Fairer F1}
\label{sec:fairer_f1}
The clean v2 test set 
consists only of BAP items with a clear context and correct structures. 84 of the EB items (7.3\% of the v2 test set) do not have a unique interpretation, since their structure  can be placed and oriented in many different ways (Section~\ref{sec:annotations}). The v2 test set can therefore be split up into items with a unique correct interpretation (\( \mathcal{U} \)) and items with multiple correct interpretations (\( \mathcal{M} \)). But while the strict F1 score defined in Section~\ref{ssec:background_eval} is appropriate for items in \(\mathcal{U} \), it unfairly penalizes Builders on items in \( \mathcal{M} \) when their structure is correct, but placed in a different location and/or orientation than the reference structure. A fair evaluation metric needs to account for these allowable changes in location and orientation for items in \(\mathcal{M}\), i.e. structures placed on an empty board when the context does not uniquely identify their placement. 

Recall from Section~\ref{sec:Formalizing-BAP} that a structure \(S = \{(\loc, \textsf{c})\}\) is a set of blocks that each have a color \textsf{c} and a location  \(\loc\), and that the distance \(\Delta(S,S')\) between two structures \(S\) and \(S'\) is defined as the number of net actions of any action sequence that changes \(S\) to \(S'\) (or vice versa). To fairly compare two structures \(S, S'\) on the empty board, we first align them by searching for a translation \(\mathcal{A}_T^{*}\) of blocks in the horizontal plane and a rotation  \(\mathcal{A}_R^{*}\) about the vertical axis in 90-degree intervals that transform \(S\) into a structure \(S^{*} = \mathcal{A}_R^{*}(\mathcal{A}_T^{*}(S))\) in the \BR that is as close to \(S'\) as possible (i.e. where \(\Delta(S^{*},S')\) is minimized). We refer to the composed transform \(\mathcal{A}_R\mathcal{A}_T\) as alignment \(\mathcal{A}\), and call  \(\mathcal{A}^* = \argmin_\mathcal{A}(\Delta(\mathcal{A}(S),S'))\) the optimal alignment of \(S\) to \(S'\).

To evaluate a model's net action set \(A_{M}^{\mathrm{net}}\) against a human reference net action set  \(A_{H}^{\mathrm{net}}\) for an item in \(\mathcal{M}\), where \(A_{M}\) leads from the (empty) board to a predicted structure \(S_{M}\), and  \(A_{H}\) leads from the (empty) board to a reference structure \(S_{H}\), we therefore first identify an optimal alignment of the predicted structure to the reference, \(\mathcal{A}^* = \argmin_\mathcal{A}(\Delta(\mathcal{A}(S_{M}),S_{H}))\), 
and apply the same transformation \(\mathcal{A}^*\) to all actions in the model's net action set \(A_{M}^{\mathrm{net}}\), yielding an aligned action sequence \(\widetilde{A}_{M}^{\mathrm{net}}\) that can now be fairly compared against \(A_{H}^{\mathrm{net}}\) by our original strict F1 metric. That is, a \textbf{fairer F1 metric} computes a strict F1 score for the predicted action sequences \(A_{M}^{\mathrm{net}}\) for any item in \(\mathcal{U}\), because the reference sequences for these items yield the only correct structures for these items, but for any item in \(\mathcal{M}\), it computes a strict F1 score for the  optimally aligned predicted action sequences \(\widetilde{A}_{M}^{\mathrm{net}}\) because this item has multiple correct interpretations that only differ from each other in location and orientation. 
Thus, in the example in Figure~\ref{fig:UsableBAPitems} (bottom), any predicted orange 3x3 U-shaped structure placed on the ground will receive a fairer F1 score of 1.0.

\subsection{Auxiliary metrics}
\label{sec:aux_metrics}
The current BAP evaluation relies solely on an aggregate F1 metric. While it reasonably reflects overall model performance (especially with fairer F1), it remains opaque and lacks detailed insight into specific model capabilities, such as spatial reasoning. This is a key missing aspect in the evaluation framework.
To address this, we introduce finer-grained auxiliary metrics to complement the aggregate fairer F1 metric. These metrics aim to enhance evaluation robustness, provide deeper insight into model behavior, enable precise error analysis, better differentiate models, identify reasons behind a model's performance changes, and ultimately guide the development of better models.

\paragraph{Type, Color, and Location F1}
In analogy to the strict F1 score, we define auxiliary metrics that only evaluate certain aspects of the predicted actions. Recall that the strict evaluation assumes that a Builder action \(a_{M}^{m} = (t,c,\loc) \in A_{M}^{\textrm{net}}\) is correct if and only if there is an \textit{equal} reference action \(a_{H}^{h} = (t,c,\loc) \in A_{H}^{\textrm{net}}\), and that our strict precision, recall and F1 scores are therefore based on the sizes of Builder and Human net actions and their intersection. To evaluate whether a Builder model correctly predicts action types \(t\), colors \(c\), or locations \(\loc\), we define corresponding evaluation metrics that consider Builder actions correct if and only if there is an \textit{equivalent} reference action in \(A_{H}^{\textrm{net}}\). We consider three different equivalence relations: \textbf{Type}, \textbf{Color} (and type), and \textbf{Location}:
\begin{align*}
 \textbf{Type equivalence:}~ (t,c,\loc) \equiv_{\textrm{Type}} (t',c',\locPrime) &\iff t = t' \\ 
 \textbf{Color equivalence:}~  (t,c,\loc) \equiv_{\textrm{Color}}  (t',c',\locPrime) &\iff t = t' ~\textrm{and}~ c = c' \\
 \textbf{Location equivalence:}~  (t,c,\loc) \equiv_{\textrm{Loc}}  (t',c',\locPrime) &\iff \loc = \locPrime 
\end{align*}
(Note that the condition for Location equivalence is effectively the same as \(t = t' ~\textrm{and}~ \loc = \locPrime\)). Accordingly, we can transform both net action sets   \(A_{M}^{\textrm{net}}\) and   \(A_{H}^{\textrm{net}}\) into Type, Color or Location \textit{multi-sets}: 
\begin{align*}
M^{\textrm{Type}} & = \{  t \mid (t,c,\loc) \in A^{\textrm{net}}\}\\
M^{\textrm{Color}} & = \{  (t,c) \mid (t,c,\loc) \in A^{\textrm{net}}\}\\
M^{\textrm{Loc}}  & = \{  \loc \mid (t,c,\loc) \in A^{\textrm{net}}\}
\end{align*}
which yields  (strict) Type, Color and Location precision, recall and F1 scores by comparing the sizes of the corresponding multisets and their intersections (we only show Type F1 below, as Color and Location F1 are similarly defined): 
\begin{align*}
\textbf{Strict Type Precision}~P_{\textrm{Type}}(A_{M}^{\textrm{net}}, A_{H}^{\textrm{net}}) & = \frac{\vert(M_{M}^{\textrm{Type}} \cap M_{H}^{\textrm{Type}})\vert}{\vert M_{M}^{\textrm{Type}} \vert}\\
\textbf{Strict Type Recall}~R_{\textrm{Type}}(A_{M}^{\textrm{net}}, A_{H}^{\textrm{net}}) &  = \frac{\vert(M_{M}^{\textrm{Type}} \cap M_{H}^{\textrm{Type}})\vert}{\vert M_{H}^{\textrm{Type}}\vert}\\
\textbf{Strict Type F1}~F1_{\textrm{Type}}(A_{M}^{\textrm{net}}, A_{H}^{\textrm{net}}) &  = \frac{2
\cdot P_{\textrm{Type}}\cdot R_{\textrm{Type}}}{P_{\textrm{Type}}+R_{\textrm{Type}}}
\end{align*}
Fairer counterparts to these strict metrics are defined analogously to the overall metric, and we will use the fair variants in the remainder of this paper. 
Intuitively, the Type metric captures how good a model is at understanding what type of actions to perform -- placements, or removals, or both, and how many of each type.
The Color metric is stricter than he Type metric, in that it quantifies how good a model is at understanding the colors of blocks that need to be placed or removed.
The Location  metric quantifies how good a model is at understanding the specific locations of blocks that need to be placed or removed (without regard to color), i.e., it helps to evaluate \textbf{spatial reasoning}.

\paragraph{Shape F1}
To evaluate how good a Builder model is at understanding the structure in which the required blocks need to be placed or removed, we define a metric, Shape F1, that is agnostic to the absolute \(\loc\) locations of the required blocks, but only cares that they be placed/removed in the right locations relative to each other. \textbf{Essentially, this metric compares net actions modulo their exact placement and orientation.}
Similar to the methodology used for fairer F1 (Section~\ref{sec:fairer_f1}), where we align the resulting built structures, we optimally align the model's net actions \(A_{M}^{\mathrm{net}}\) against the human reference net actions \(A_{H}^{\mathrm{net}}\), yielding aligned net actions \(\hat{A}_{M}^{\mathrm{net}}\) that can now be compared against \(A_{H}^{\mathrm{net}}\) by our original strict F1 metric. This is done for all items in the v2 test set (and not just the ones with multiple interpretations like was done for fairer F1).
We will see examples of Shape F1 and the other auxiliary metrics  in Sections~\ref{sec:qual_eval_analysis}.

\subsection{BAP v2 evaluation benchmark}
\label{sec:v2_benchmark}

The above metrics enable a fairer, robust, and more insightful evaluation.
We validate this on the GRU-based baseline model for the BAP task (Section~\ref{sec:baseline_model_desc}).
Table~\ref{tab:micro_eval} reports the full battery of scores included in the \textbf{BAP v2 evaluation benchmark}, i.e. overall and auxiliary F1 scores on the overall test set as well as for the empty and non-empty board (EB and NEB) subsets. 
We note that the overall F1 score has risen from 21.1 reported in \citet{jayannavar-etal-2020-learning} to 27.3 under our newer, fairer evaluation, or 53.2 for EB and 24.2 on the (more frequent) NEB items. The distinction between EB vs. NEB is important since EB examples are generally easier, making EB performance a basic competency or sanity test, and a reflection of a model's maximum potential quality. EB performance is also an indication of a model's ability of a model to begin a game accurately -- an elementary yet important trait to measure, especially for interactive/online settings where errors might cascade throughout a game. The auxiliary metrics provide further insight: in particular, we see that  the model achieves reasonably good F1 scores of 65.2\% for Type and 62.4\% for Color, but only 27.7\% for Location F1, which is also very close to the overall F1 of 27.3\% (similar results hold both for the EB and NEB setting). This suggests that spatial reasoning is the key bottleneck for high performance on this task.

\begin{table}[htbp]
\centering
\begin{small}
\begin{tabular}{llcccc}
\toprule
\textbf{Dataset} & \textbf{Type} & \textbf{Color} & \textbf{Location} & \textbf{Shape} & \textbf{Overall} \\
\midrule
EB & 76.4 & 75.2 & 53.5 & 56.0 & 53.2 \\
\midrule
NEB & 63.8 & 60.8 & 24.6 & 33.8 & 24.2 \\
\midrule
Overall & 65.2 & 62.4 & 27.7 & 36.3 & 27.3 \\
\bottomrule
\end{tabular}
\end{small}
\caption{F1 scores for the GRU baseline model on the v2 benchmark}
\label{tab:micro_eval}
\end{table}

%% file: data.tex
\section{Revisiting BAP Training Data: Generating Synthetic Data}
\label{sec:syn_data_gen}

\begin{figure}[!htbp]
\centering
\begin{tabular}{c@{\hskip 10mm}c@{\hskip -1mm}c}
\toprule
\multicolumn{3}{l}{\textbf{Synthetic Target Structures}}\\
\midrule
\textbf{Random}
&
\multicolumn{2}{c}{\textbf{Shape-Based}}\\
\includegraphics[width=0.2\textwidth]{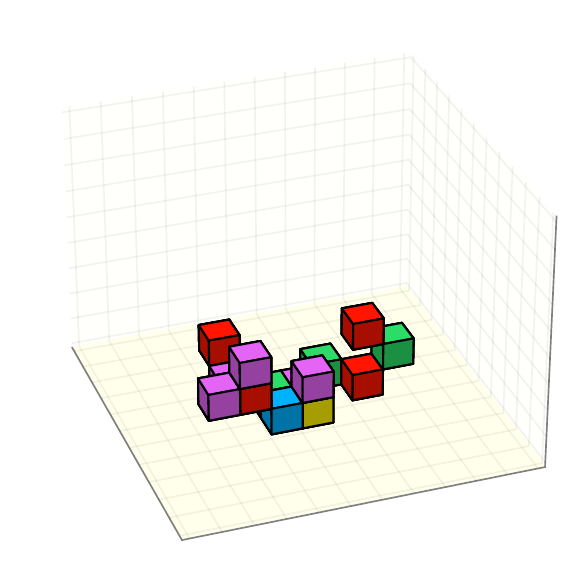}
& \includegraphics[width=.2\textwidth]{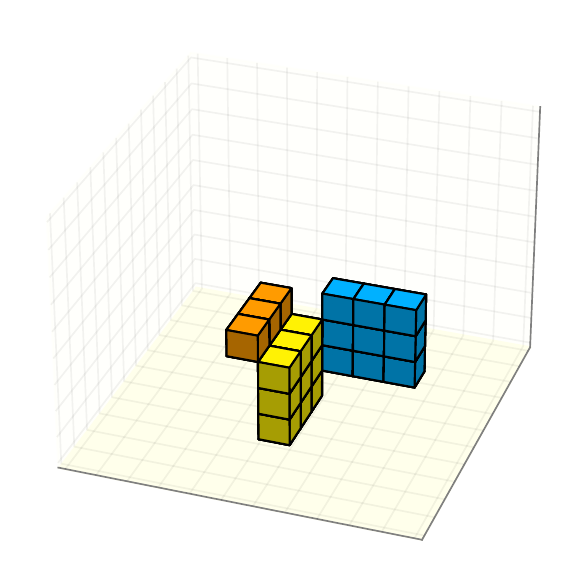} 
& \includegraphics[width=.2\textwidth]{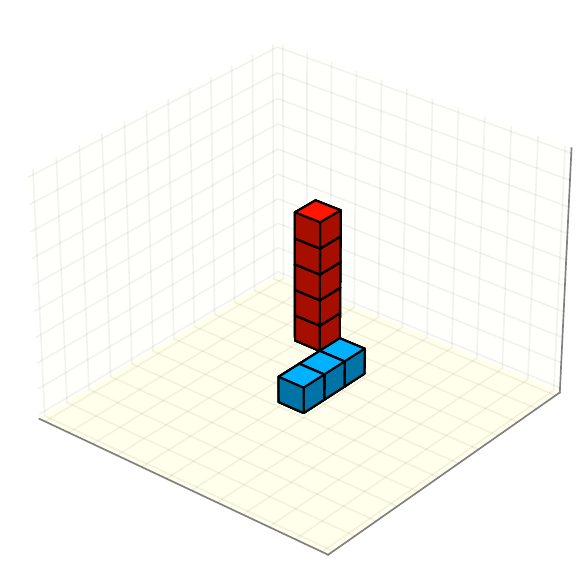}\\
\bottomrule
\end{tabular}
\caption{Synthetic target structures}
\label{fig:synthetic_target_structures}
\end{figure}

\begin{figure*}[!htbp]
\centering
\begin{tabular}{l}
\toprule
\textbf{Synthetic BAP items}\\
\midrule
\textbf{(A) Block-based dialogues for random targets (\dr)}\\
\midrule
 \includegraphics[width=.8\textwidth]{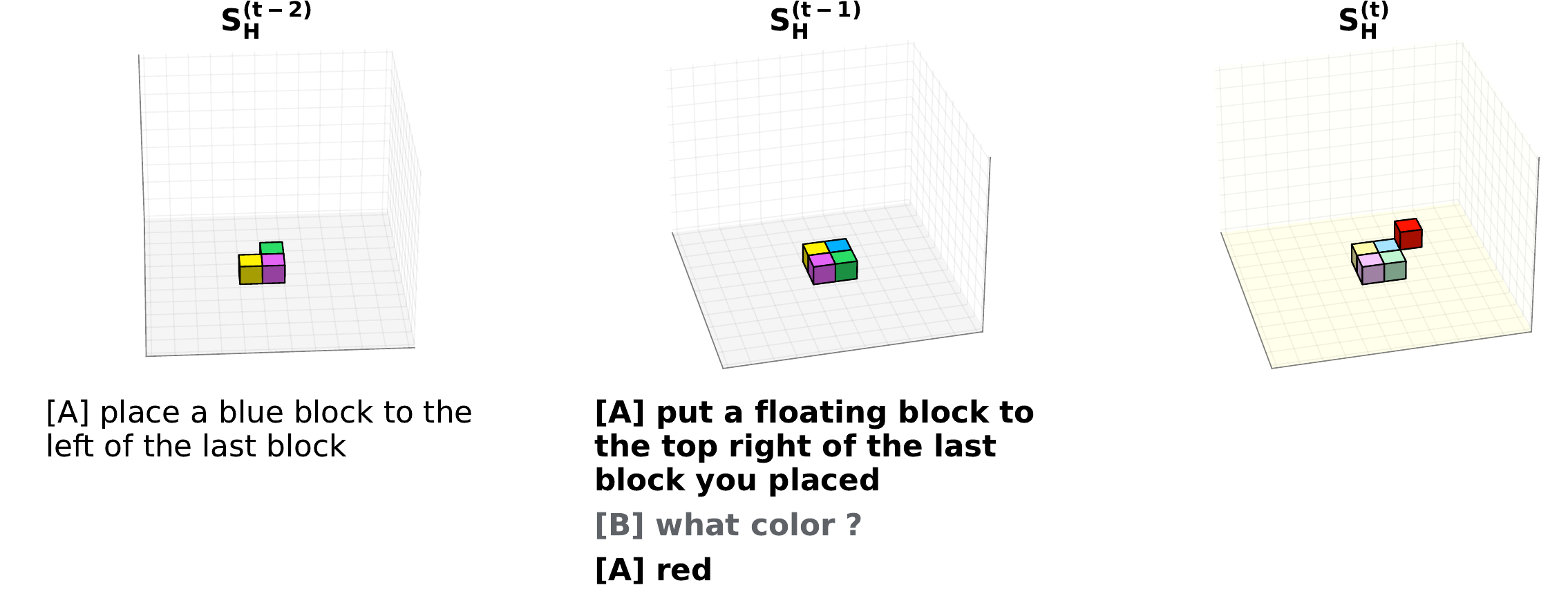}\\
\includegraphics[width=.8\textwidth]{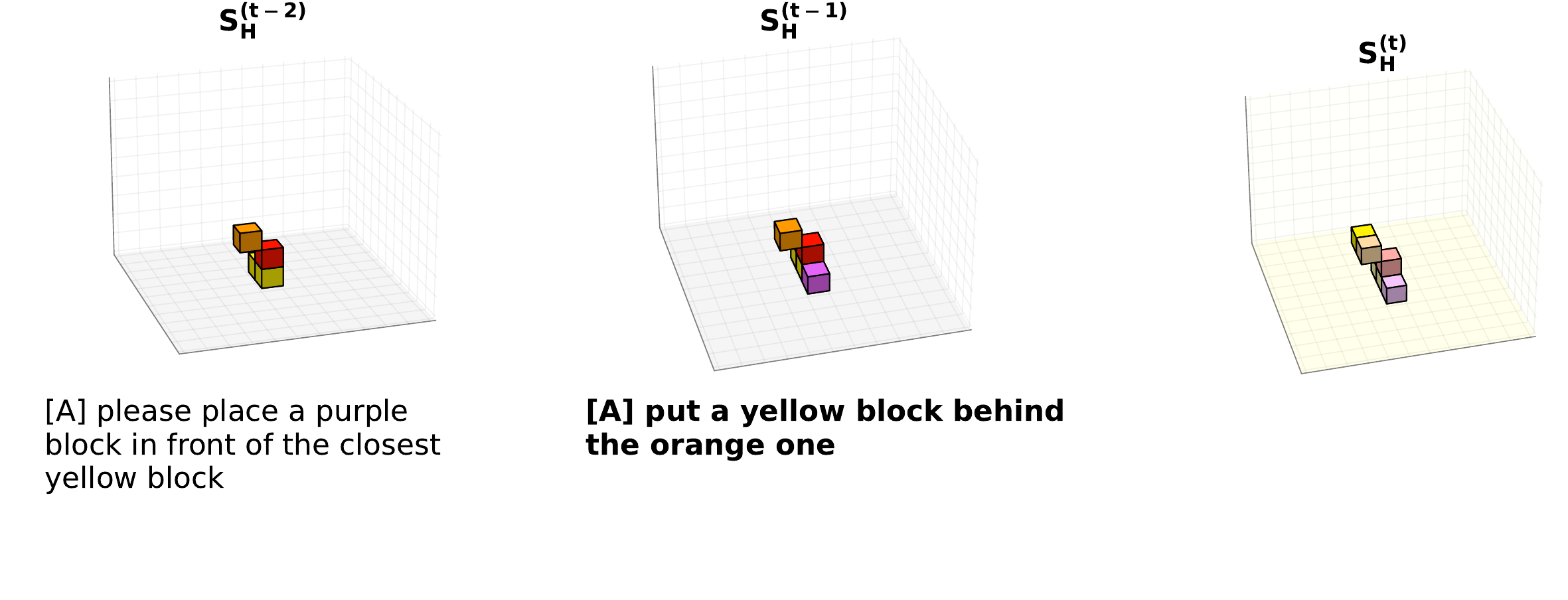}\fignegspace\\
\midrule 
\textbf{(B) Block-based dialogues for shape-based targets (\dbs)}
\\\midrule
  \includegraphics[width=.8\textwidth]{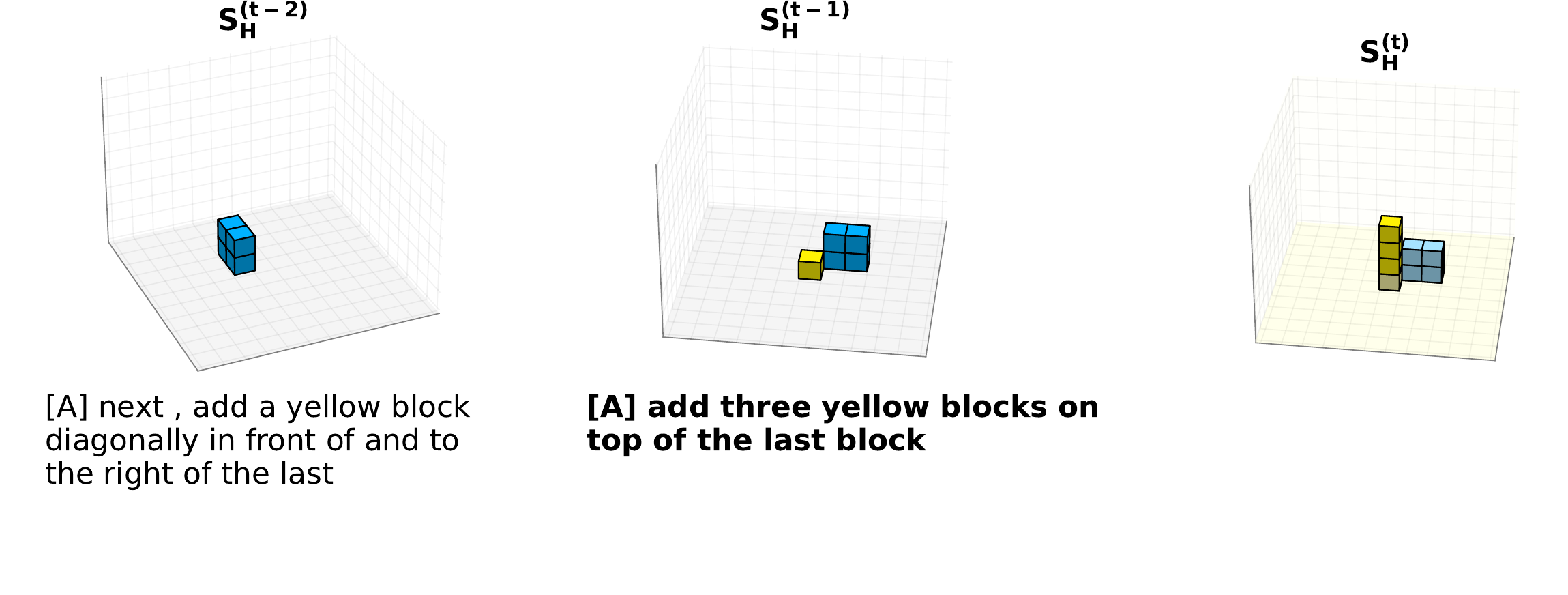}\fignegspace
  \\
\midrule
\textbf{(C) Shape-based dialogues for shape-based targets (\dss)}
\\\midrule
\includegraphics[width=.8\textwidth]{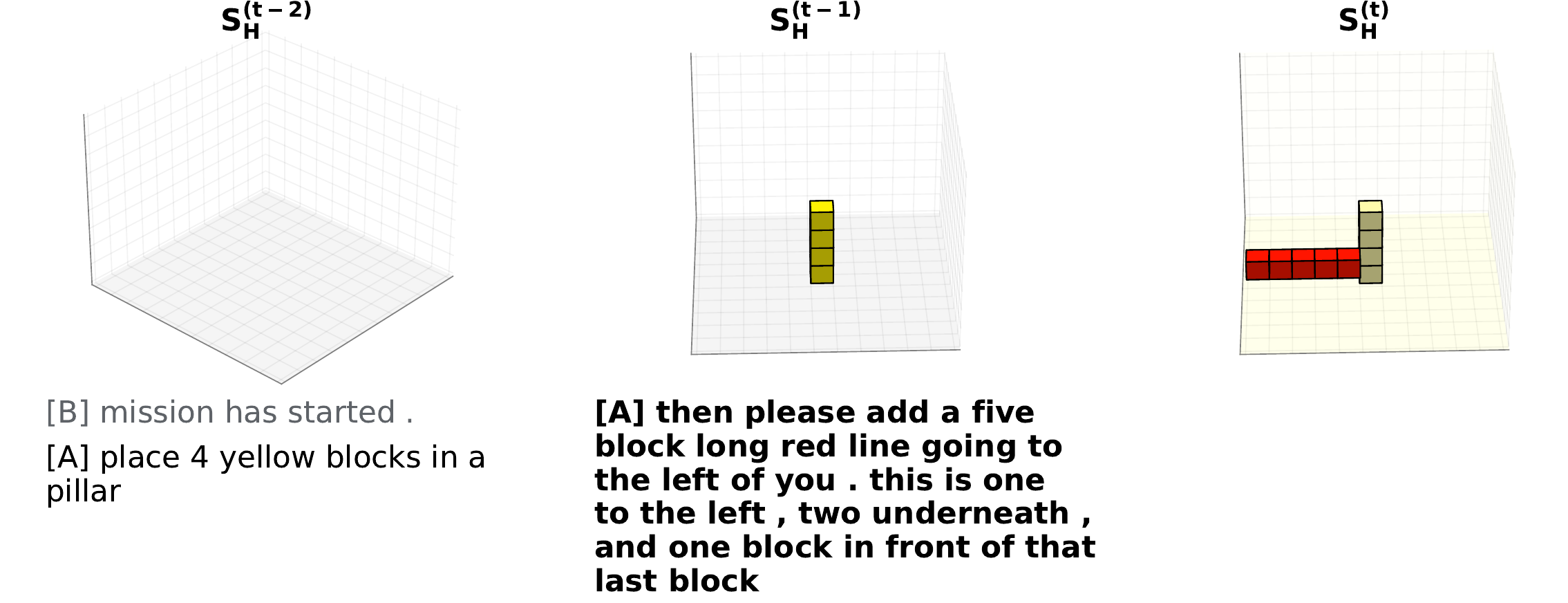}\\
\bottomrule
\end{tabular}
\caption{We generate three different types of synthetic BAP items, resulting in datasets \dr, \dbs and \dss}
\label{fig:synthBAPs}
\end{figure*}

\begin{figure*}[!htbp]
\centering
\begin{tabular}{l}
\toprule
\textbf{Our process to generate synthetic BAP items}
\\
\midrule 
\includegraphics[width=\textwidth]{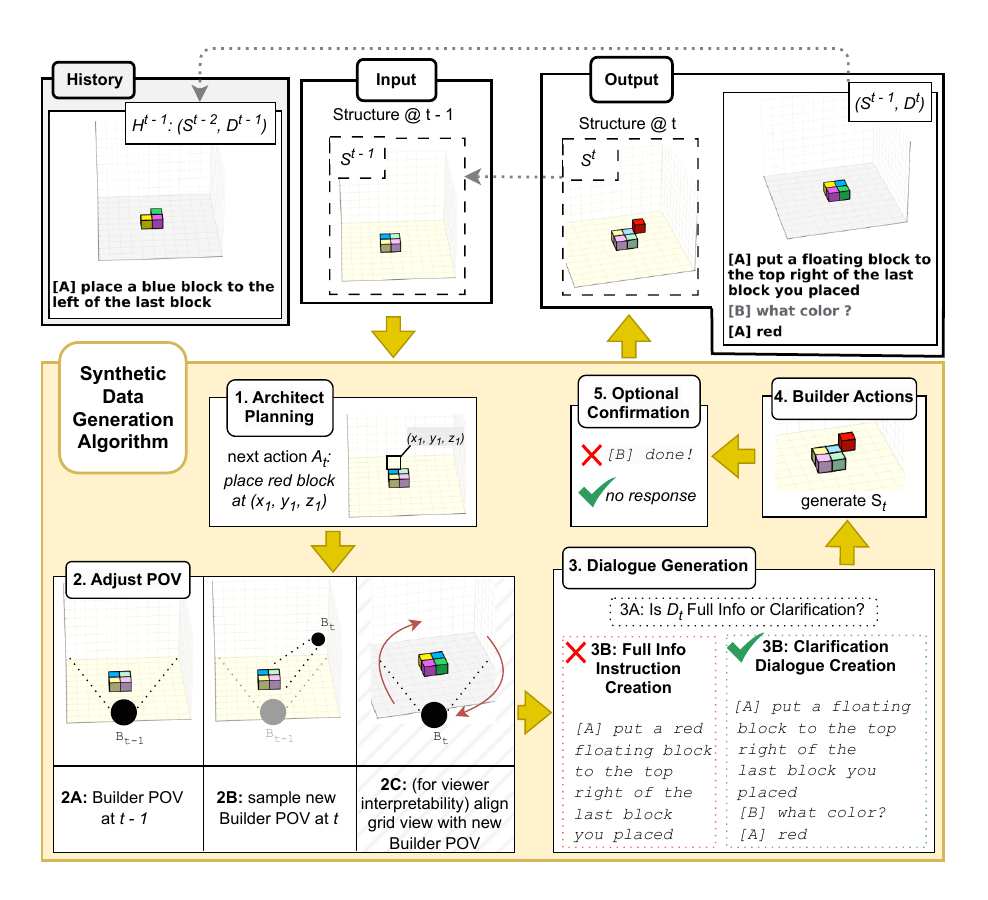}\\
\midrule
\textbf{The synthetic BAP item generated by the process above}\\
\midrule 
\includegraphics[width=0.9\linewidth]{Figures/Tex/dialog_28_example_4_level2.pdf}\\
\bottomrule 
\end{tabular}
 
\caption{Our process to generate synthetic BAP items (top) and a BAP item generated by that process (bottom)}
\label{fig:genSynItems}
\end{figure*}

\begin{table*}[!htbp]
\centering
\resizebox{\linewidth
}{!}{%
\begin{tabular}{@{}>{\raggedright}p{2.5cm} p{3.8cm} p{3.8cm} p{3.8cm} p{3.7cm}@{}}
\toprule
\textbf{Feature} & \textbf{MDC (\dmc)} & \textbf{\dr} & \textbf{\dbs} & \textbf{\dss} \\
\midrule

\textbf{Target Structure} & Human-designed, complex and meaningful (e.g., furniture, animals). & Dynamically and randomly generated; typically disordered. & Predefined composites of 6 elementary shapes. & Predefined composites of 3 simpler shapes. \\
\addlinespace

\textbf{Architect Planning} & Complex, strategic \& adaptive human planning with hierarchical decomposition of target structures based on goals of efficiency, naturalness in instruction giving, etc. & Minimalistic; plans single random block placement/removal per turn. & Mid-level; plans one shape at a time, decomposed into block-level instructions. & High-level; plans construction of one full shape instance per turn. \\
\addlinespace

\textbf{Instruction Abstraction} & Multi-level: from high-level concepts (\exampleutt{flower}) to low-level commands. & Low-level; single block references only. & Low-level; but more abstract than \dr; decomposes shapes into block-level actions. & Mid-level; refers to entire shapes (\exampleutt{row}, \exampleutt{plane}). \\ 
\addlinespace

\textbf{Semantic Naming} & Frequent use of names for commonplace objects (\exampleutt{chicken}), shapes \& user-defined concepts (\exampleutt{r-windows}). & N/A (block-level instructions only). & N/A (block-level instructions only). & Limited; template-based names for shapes (\exampleutt{pillar}, \exampleutt{line}). \\
\addlinespace

\textbf{Spatial Relations} & Highly diverse, complex, and nuanced (\exampleutt{parallel}, \exampleutt{symmetrical}, \exampleutt{hanging}). & Simple 1D/2D relations anchored to arbitrary reference blocks. & Simple 1D/2D relations anchored to the \exampleutt{last placed block}. & All 1D/2D/3D relations to specify shape's starting point and Builder-anchored relations (\exampleutt{going to the left of you}) to specify orientation. \\
\addlinespace

\textbf{Floating Blocks} & Present; instructions range from single-shot commands to step-by-step guidance. & Present; single-shot commands only. & Present; single-shot commands only. & Present; single-shot commands only. \\
\addlinespace

\textbf{Context Clarity} & Variable; includes unclear/ambiguous contexts (Section~\ref{sec:eval}). & Always clear \& unambiguous. & Always clear \& unambiguous. & Always clear \& unambiguous. \\
\addlinespace

\textbf{Dialogue Synchronicity} & Asynchronous; mutual interruptions by \A and \B are possible. & Synchronous; discrete, structured turns. & Synchronous; discrete, structured turns. & Synchronous; discrete, structured turns. \\
\addlinespace

\textbf{Structural Correctness} & Contains correct \& incorrect structures (due to mistakes/interruptions; Section~\ref{sec:eval}). & Always correct. & Always correct.  & Always correct.  \\
\addlinespace

\textbf{Dialogue Act Diversity} & High; wide mix (instructions, confirmations, corrections, social chat, etc.). & Low; dominated by instructions \& short clarification exchanges. & Low; dominated by instructions \& short clarification exchanges. & Low; dominated by instructions and short clarification exchanges. \\
\addlinespace

\textbf{Builder Initiative} & High; \B actively verifies, suggests, gives updates and extrapolates. & Low; limited to simple clarification questions. & Low; limited to simple clarification questions. & Low; limited to simple clarification questions. \\
\addlinespace

\textbf{Clarification Exchanges} & Varied and natural questions about ambiguities (
\exampleutt{is it flat?}). & Simulated; on missing block color or location. & Simulated; on missing block color or location. & Simulated; on missing color, location, size, or orientation. \\
\addlinespace

\textbf{Referring Expressions} & Highly varied; includes implicit arguments and ellipsis (\exampleutt{two more in the same direction}). & Simple; disambiguate blocks by color and relative position (\exampleutt{leftmost red block}). & Simple; refers to the last placed block as spatial anchor. & Simple; refers to last placed block of previous shape as spatial anchor. \\
\addlinespace

\textbf{Builder's Low-Level Planning} & Required \& involved; scales with instruction abstraction. & Minimal; executes single-block instructions. & Minimal; executes block-level instructions. & Required but simple; uses heuristics (e.g., column-by-column, row-by-row, or zigzag patterns) for shape instructions. \\ 

\bottomrule
\end{tabular}%
}
\caption{Detailed qualitative comparison of MDC and synthetic datasets. For all rows, MDC features are a superset of the ones for the synthetic datasets.}
\label{tab:dataset_comparison}
\end{table*}

Despite the data augmentation (Section~\ref{sec:dataaug}), the BAP task suffers from relatively limited training data, but collecting a significantly larger amount of human data  would have been prohibitively expensive  and impractical. More importantly, while the MDC dialogues are far more linguistically rich and varied than our synthetic data, the baseline models struggle even with basic spatial understanding and reasoning (as reflected in the Shape and Location F1 scores). In order to achieve comprehensive coverage of various spatial relations, we therefore opted to create synthetic training data that would allow us to train our models in these basic skills. Specifically, we distinguish between \textbf{two different types of target structures} ––random and shape-based (Figure~\ref{fig:synthetic_target_structures}) –, and 
\textbf{three different types of dialogues} ––block-based for random or shape-based targets, or shape-based for shape-based targets–– (Figure~\ref{fig:synthBAPs}), described in detail below.

We design novel \textbf{Minecraft target structure and dialogue simulators} that emulate the MCBT, and use them to generate a set of three synthetic dialogue datasets with corresponding target structures. 
Each dataset is modeled after the MDC and, while understandably simpler, preserves essential elements of realistic human behavior observed in the MDC as much as possible (including, e.g. clarification exchanges, Architect \A’s and Builder \B’s planning perspectives, dialogue flow, etc.).
While we specifically emphasize spatial relations (as it is a core motivation) and representing examples of different complexities, we also vary the level of instruction abstraction, the complexity of referring expressions, the complexity of target structures, etc. across these datasets. Each is thus unique, contributing to the diversity and richness of the overall set.  

Looking ahead, such data could also be crucial for training more sophisticated, data-hungry deep transformer models and training/fine-tuning increasingly large LLMs. And although our primary focus is on BAP, the simulators and data introduced here are applicable to other MCBT subtasks as well (e.g. the AUG subtask; Section~\ref{sec:mcbt_subtasks}), supporting advancements across the broader MCBT ecosystem. We discuss this in more detail in Section~\ref{sec:within_mcbt}.

\subsection{Our dialogue simulation framework}
As described in Section~\ref{sec:minecraft_task}, the MCBT is characterized by asynchronous, loosely structured dialogue between human players with minimal constraints on their language and actions. To make the simulation of this complex interaction tractable and to facilitate a simplified data generation process, we introduce a simpler and more structured framework. Like the original MDC data, for each game, a target structure and a complete log that interleaves the dialogue between \A and \B (consisting of instructions, potential clarification exchanges), \B's actions (which reflect the evolving structure being built), and \B's changing position and orientation (so that spatial references in the utterances can be correctly interpreted). 
Our simulators generate grounded synthetic dialogues that meet these desiderata by iteratively simulating the MCBT until the target structure is built. Each iteration of the algorithm produces a BAP item, \((\mathcal{H}^{t-1}, S^{t-1}, B^{t}, D^{t}, A^{t}, S^{t})\) (as defined in Section~\ref{sec:BAP}), by simulating the dialogue $D^{t}$ and actions $A^{t}$ following $S^{t-1}$ to produce the next structure, $S^{t}$. Each item  consists of the specification of \B's current position $B^{t}$, a non-empty dialogue segment $D^{t}$—\A's instruction, optional clarification exchanges with \B, and confirmations– followed by \B's actions $A^t$ that alter the built structure from $S^{t-1}$ to $S^{t}$, thus constructing part of the target structure.
\(\mathcal{H}^{t-1}\) and \(S^{t-1}\) represent the current game context that is input to this iteration of the algorithm.

The target structure for the dialogue is handled in one of two ways: it is either provided as a predefined input to the algorithm (generated by a separate target structure simulator) or generated dynamically. If predefined, each iteration involves selecting the next blocks to place from this target. If generated dynamically, each iteration involves choosing blocks randomly to place or remove, and the final constructed configuration is designated as the target structure for the dialogue post-hoc. 

Each iteration proceeds according to the steps detailed below, which is also visualized in the running example in Figure~\ref{fig:genSynItems} (bottom).

\begin{enumerate}[itemsep=12pt]
    \item \textbf{Architect Planning (Step 1):} \label{list:step_1} 
    The simulation begins by sampling the Architect's plan for the next construction phase. This includes determining which blocks to place or remove and in what order. A reference block is chosen, when necessary, enabling \A to describe the location of the new block(s) relative to the reference block in the subsequent dialogue $D^t$. In our running example, \A decides to place a red floating block, using the previously placed blue block as a reference (diagonally from it at \((x_1, y_1, z_1)\).
    
    \item \textbf{Adjusting the POV to a new Builder Position and Orientation (Step 2):} \label{list:step_2} 
    As noted in Section~\ref{sec:features_mcbt}, \A and \B can move around freely, but \A views the structure from behind \B when giving instructions. As a result, \A instructions (including spatial relations, etc.) are grounded in \B's perspective. To emulate this, this step simulates \B's position and orientation $B^t$ (see Section~\ref{sec:bpos} for a formal definition). A valid $B^t$ is sampled for \B, establishing the frame of reference for the upcoming dialogue, $D^t$. $B^t$ is considered valid if \B is at a suitable distance from the action area and oriented to face it, while respecting certain world constraints. E.g., Steps 2A and 2B show $B^t$ being chosen. In Step 2C, the grid is also rotated to align with \B's point-of-view and to view $S^{t-1}$ according to that POV -- just for the reader's visual clarity. Note that technically, \step{1} and \step{2} are independent and can be performed in any order.

    \item \textbf{Generating the Main Dialogue (Step 3):} \label{list:step_3}
    \begin{enumerate}
        \item \textbf{Dialogue Style (Step 3A):} \label{list:step_3A} 
        The simulation first determines the dialogue style for the turn. It randomly chooses between a straightforward, complete instruction and, with much lower probability, an instruction that intentionally omits information to elicit a clarification question from the Builder. 
        
        \item \textbf{Dialogue Generation (Step 3B):} \label{list:step_3B}
        Next, the dialogue $D^t$ is synthesized using templates that incorporate lexical and syntactic variations. If a direct instruction was chosen, the utterance includes all necessary details (e.g., color, spatial relations (\ref{sec:spatial_rels}), referential expressions). If a clarification was chosen, a key detail is omitted, and a corresponding question-answer exchange between \B and \A is generated. We show  both possibilities for $D^t$ (omitting the block's color for the clarification case).
    \end{enumerate}
    
    \item \textbf{Generating the Builder Actions (Step 4):} \label{list:step_4} 
    The Builder's actions ($A^t$) to execute the instruction are then simulated. This includes low-level planning to decide the precise order of actions and to account for temporary supporting blocks required for construction. These actions update the built structure, producing the new state, $S^t$. In our example, \B places a red floating block to produce structure $S^{t}$, temporarily adding and removing a support block to do so.\footnote{As noted in Section~\ref{sec:BAP}, temporary supporting blocks are not shown in our figures.}
    
    \item \textbf{Generating an Optional Confirmation from Builder (Step 5):} \label{list:step_5} 
    Finally, the turn may (with a low probability) conclude with a confirmatory utterance from \B, such as \exampleutt{done!}. 
\end{enumerate}

The complete BAP item produced by this five-step process is shown in Figure~\ref{fig:genSynItems} (bottom), which follows the standard visualization format used throughout this paper, 
so that  structures like $S^{t-1}$ are rotated to align with the Builder's POV $B^t$ (corresponding to the next turn and dialogue segment $D^t$ below it), and not with $B^{t-1}$, whereas Steps 2A-2C in the top of Figure~\ref{fig:genSynItems} explicitly show the transition from the previous perspective ($B^{t-1}$) to the new one ($B^t$).

We developed three distinct simulators based on this general framework. While \step{2}, \step{4}, and \step{5} are implemented largely consistently across all three, each simulator features a unique implementation of Architect Planning (\step{1}) and Generating the Main Dialogue (\step{3}), as well as its own target structure generation. For brevity, we provide only a summary of each simulator in the following sections and provide more details for each in the appendix (Section~\ref{sec:appendix_syn_data_gen}).

\subsection{Generating Grounded Aspects of the Synthetic Dialogs}
\label{sec:grounding_gen}
We now briefly elucidate some key details of \step{2} and the spatial relations part of \step{3B}, common across all three simulators.

\subsubsection{Sampling the Builder's Position and Orientation \step{2}}

For each turn in the simulation, we sample a new position and orientation for the Builder $\text{Pos}_\B = (\text{loc}_\B, \phi_\B, \gamma_\B)$, where $\text{loc}_\B = \langle x_B, y_B, z_B \rangle$ is the Builder's location, $\phi_\B$ is its pitch (vertical rotation), and $\gamma_\B$ is its yaw (horizontal rotation) (Section~\ref{sec:bpos}). 
This sampling is grounded by the current state of the world, specifically the location of the \textit{reference block}, $r_p$, which will be used as a spatial anchor in the Architect's instruction.

The process begins by identifying a set of all possible locations from which the Builder has an unobstructed view of the reference block. We then sample the Builder's position, $\text{loc}_\B$, uniformly from this feasible set. With the position established, we calculate the optimal yaw and pitch angles, $(\gamma_\B^*, \phi_\B^*)$, that would orient the Builder to look directly at the reference block from its new position:
\begin{align*}
    \gamma_\B^* &= \arctan\left(- \frac{r_x - x_B}{r_z - z_B}\right) \\
    \phi_\B^* &= \arctan\left( \frac{y_B - r_y}{\sqrt{(x_B - r_x)^2 + (z_B - r_z)^2}}\right)
\end{align*}
To introduce natural variation and avoid perfectly robotic behavior, the Builder's final orientation is sampled from a normal distribution centered on these optimal angles:
\begin{align*}
    \gamma_\B &\sim \mathcal{N}(\gamma_\B^*, h/2) \\
    \phi_\B &\sim \mathcal{N}(\phi_\B^*, v/2)
\end{align*}
where $h$ and $v$ are the Builder's horizontal and vertical Field of View (FOV), respectively. We assume a ``normal'' game setting and 16:9 ratio of FOV, \emph{i.e.}, $h=102.4^\circ$ and $v=70^\circ$.\footnote{\url{https://minecraft.fandom.com/wiki/Options}} This ensures the Builder is generally facing the area of interest without being perfectly fixated on it.

\subsubsection{Grounding Spatial Relations in \B's Perspective (\step{3B})}
\subsubsection*{Computing perspective coordinates}
Once the Builder's POV for the turn is established, we can generate spatial descriptions that are naturally grounded in this perspective. This is achieved by converting absolute world coordinates into \textbf{`perspective' coordinates} \citep{jayannavar-etal-2020-learning}, which remap the 3D space relative to the Builder's position and orientation. Any absolute location $\text{loc} = \langle x_c, y_c, z_c \rangle$ is transformed into perspective coordinates $\text{loc}' = \langle x_c', y_c', z_c' \rangle$ by first translating the origin to the Builder's location $\text{loc}_\B$, and then applying rotations to account for the Builder's yaw ($\gamma_\B$) and pitch ($\phi_\B$):\footnote{The rotation matrices for pitch ($P$) and yaw ($Y$) are defined as:
 \(P\!=\!\left( \begin{smallmatrix}1 & 0 & 0\\
            0 & \cos \phi_\B & \sin \phi_\B\\
            0 & - \sin \phi_\B & \cos \phi_\B\end{smallmatrix}\right)\) and  
            $Y\!=\!\left(\begin{smallmatrix}
            \cos \gamma_\B & 0 & - \sin \gamma_\B\\
            0 & 1 & 0\\
            \sin \gamma_\B & 0 & \cos \gamma_\B
        \end{smallmatrix}\right)$.
        }
\begin{equation*}
    \langle x_c',y_c',z_c'\rangle  = P \cdot Y \cdot \langle x_c-x_B,y_c-y_B,z_c-z_B\rangle
\end{equation*}

\subsubsection*{Computing spatial relations}
We leverage these perspective coordinates to generate phrases describing the position of the next block to be placed, $p$, relative to the reference block, $r$. We first convert both their absolute positions, $p_p$ and $r_p$, into perspective coordinates, yielding $p_p'$ and $r_p'$. The choice of directional words (e.g., "left/right" vs. "front/behind") is determined by comparing the difference between these new coordinates along the axes.

For example, to determine the primary horizontal relation, we compare the magnitude of the difference along the Builder's perspective x-axis (left/right) with the difference along the z-axis (front/behind). If $|p_x' - r_x'| > |p_z' - r_z'|$, the most salient direction is left or right. The specific term is chosen based on the sign of the difference: if $p_x' > r_x'$, the instruction would be "to the left of" the reference block. Conversely, if the difference along the z-axis is more significant, the instruction would use "in front of" or "behind". This method provides a deterministic way to generate simple and intuitive spatial language that is directly tied to what the Builder would be seeing from its simulated vantage point.

\subsubsection{Types of Spatial Relations}
\label{sec:spatial_rels}
We generate the following types of spatial relations: 
\begin{enumerate}
    \item \textbf{Relative to a Reference Block:} We can describe the location of a target block in relation to another block, referred to as the reference block. Around the reference block, there are 6 rows, 12 quadrants, and 8 octants where the target block may be positioned. This leads to the following spatial relations (all of them are considered with respect to \B's POV/frame of reference):
    \begin{enumerate}
        \item \textbf{1D:} The target block lies in one of the rows, differing from the reference block in only one dimension. This results in 6 basic spatial relations: left, right, top, bottom, front, and behind (e.g., \exampleutt{to the right of ...}, \exampleutt{behind ...}).
        
        \item \textbf{2D:} The target block lies in one of the quadrants, differing in exactly two dimensions. This results in 12 possible combinations of the 1D relations, such as top+right, bottom+left, etc. (e.g., \exampleutt{to the top right of ...}).
        
        \item \textbf{3D:} The target block lies in one of the octants, differing in all three dimensions. This results in 8 possible combinations of the 1D relations, such as top+right+front (e.g., \exampleutt{one to the left, one above, and one block behind ...}).
    \end{enumerate}
    These 1D/2D/3D relations apply when one needs to specify spatial relations between blocks or substructures/shapes.
    \item \textbf{Using \B as a Spatial Anchor:} Spatial relations can also be described using \B as a spatial anchor  directly (e.g., \exampleutt{going up and to the left of you}). These only apply when one needs to specify the direction in which to build a certain substructure/shape, e.g., a diagonal.
\end{enumerate}
We will see concrete examples as we describe each dataset simulator in turn. 

\subsection{Dataset \dr: Dialogues for random target structures}
\label{sec:random_data}

In this dataset, target structures are generated dynamically as a direct outcome of the dialogue simulation, corresponding to the final configuration of blocks at the end of the game. 
Dialogues are generated by running the simulation framework (Section~\ref{sec:syn_data_gen}) for a bounded, random number of turns. The key characteristics of this simulator emerge from its specific implementation of the Architect's planning and dialogue generation steps.
The \textbf{Architect Planning (\step{1})} process is intentionally minimalistic: at each turn, the Architect decides to place or remove  a single block at a random valid location adjacent to the existing structure. This lack of sophisticated, high-level planning results in the dynamically generated, disordered target structures mentioned above, and directly influences the \textbf{Generating the Main Dialogue (\step{3})}. Because the Architect considers only one block at a time, the resulting instructions are inherently low-level. They typically use 1D or 2D spatial relations that are anchored to arbitrary reference blocks, which are identified by their color or location (e.g., \exampleutt{behind} or \exampleutt{to the right of the red} or \exampleutt{to the right of the bottom-most red block}, or, as in Figure~\ref{fig:synthBAPs} (top), \exampleutt{put a yellow block behind the orange one}.
Further details on this simulator are available in Appendix~\ref{sec:appendix_syn_data_gen}.

\subsection{Datasets \dbs and \dss: Dialogues for shape-based target structures} 
Target structures in the MDC are not random; they are far more ordered (e.g., a dining table) and often composed of meaningful shapes/concepts (e.g., rows, towers, diagonals, planes, etc.). Architect instructions describe these structures at various levels of abstraction, from high-level references to the whole structure (\exampleutt{flower} or \exampleutt{bell tower}), mid-level references to sub-shapes (\exampleutt{row} or \exampleutt{plane}), down to low-level instructions referencing individual blocks. This observation motivates the need for similarly structured synthetic data as well.
We first introduce a simulator that can generate such ordered and shape-based target structures (Section~\ref{sec:targets_shapes_based}), and then outline two types of distinct dialogue simulators based on these structures (Sections~\ref{sec:blocks_based_dialogues} and~\ref{sec:shape_based_dialogues}).

\subsubsection{Shape-based structures}
\label{sec:targets_shapes_based}
We define six unique elementary shapes: \textbf{rows, diagonals, T-shapes, L-shapes, U-shapes, and planes}. These shapes were chosen for their simplicity and occurrence in the MDC. (Formal definitions of the shapes are provided in the appendix (Section~\ref{app:shape_structs})).
We then implemented a flexible target structure simulator that programmatically combines these elementary shapes into complex, composite structures. The simulator operates by randomly sampling multiple shape instances, allowing for repetition. For each instance, it further randomizes key properties such as size, color, orientation, and 3D location within specified constraints (e.g., minimum and maximum size bounds). The simulator is flexible, allowing parameters such as shape types, number of shapes per structure, etc. to be customized.

\subsubsection{Dataset \dbs: Blocks-based dialogues for shape-based targets}
\label{sec:blocks_based_dialogues}
Each target structure consists of three shape instances sampled from the full set of six elementary shapes. Dialogues are generated by applying the simulation framework to the predefined shape-based structures until construction is complete. The simulation breaks down the construction of complex shapes into a sequence of simpler, block-level instructions. The \textbf{Architect Planning (\step{1})} is more sophisticated here. The Architect formulates a plan to build the target one shape instance at a time. Within each shape, it uses a heuristic to select one or more nearby blocks to place in each turn, ensuring a contiguous and logical build order. Consequently, the \textbf{Generating the Main Dialogue (\step{3})} produces low-level instructions that decompose each shape into multiple steps, referencing either a single block or a small group of blocks. A key linguistic feature of this dataset is that spatial relations are primarily anchored to the last block placed by \B (e.g., \exampleutt{now, place two red blocks behind the last block}).  Figure~\ref{fig:synthBAPs} (middle) provides examples. The first blue square has been built, and the second (yellow) shape is under construction. \A instructs \B to continue building the latter by adding blocks relative to the last one placed. \A uses a 1D spatial relation \exampleutt{on top of} and the last block as the reference block. (In the previous step, \A used a 2D spatial relation, \exampleutt{diagonally in front of and to the right of}.) Further details on this simulator are available in Appendix~\ref{sec:appendix_syn_data_gen}.

\subsubsection{Dataset \dss: Shape-based dialogues for shape-based targets}
\label{sec:shape_based_dialogues}

This simulator generates dialogues at a higher level of abstraction than the block-based simulator described above, but, to ensure the dialogue simulation remains tractable and to avoid overly complex utterances,  each target structure here is limited to two shape instances, sampled from only three elementary types: rows, diagonals, and planes.
In \textbf{Architect Planning (\step{1})}, the Architect plans to construct an entire shape instance within a single turn. The plan involves simple heuristics to select the order of the two shape instances and to identify a starting block for the second shape relative to the first.
This high-level plan leads to more abstract, mid-level instructions during \textbf{Generating the Main Dialogue (\step{3})}. Instead of referring to individual blocks, utterances describe whole shapes (e.g., \exampleutt{add a five block long red line...}). Spatial relations are used in two distinct ways: 1D, 2D, or 3D relations are used to specify the starting point of the second shape relative to the first, while relations that use \B as a spatial anchor (Section~\ref{sec:spatial_rels}) are used to define a shape's orientation (e.g., \exampleutt{...going to the left of you}). This approach also simulates more complex \textbf{Generating the Builder Actions (\step{4})}. Given a high-level instruction to build a shape, the Builder must use its own heuristics to determine the actual sequence of block placements, mimicking how a human might build. E.g., 2D shapes like planes can be built either column-by-column or row-by-row. Additionally, \B may also adopt a zigzag pattern, alternating the starting side for each row or column.   Figure~\ref{fig:synthBAPs} (bottom) shows an example. The first shape (a yellow column) has been built, and \A instructs \B to build the second shape (an entire red line) in one go, using both types of spatial relations. \A uses the spatial relation \exampleutt{going to the left of you} (using \B as a spatial anchor) to indicate the direction for the row, along with the 3D spatial relation \exampleutt{one to the left, two underneath, and one block in front of} relative to the last block (the topmost yellow one), to indicate the starting position for the row.
Further details on this simulator are available in Appendix~\ref{sec:appendix_syn_data_gen}.

\subsection{A Qualitative Comparison of the Datasets and MDC}

The three synthetic datasets—\dr, \dbs, and \dss—are designed to augment the original MDC (\dmc) by providing targeted examples rich in spatial language. While they emulate many core aspects of the MCBT, they also systematically simplify certain phenomena to make data generation tractable and to focus on specific learning objectives. To provide a clear overview of the trade-offs and characteristics of each dataset, Table~\ref{tab:dataset_comparison} presents a detailed side-by-side qualitative comparison. The table highlights key differences in areas such as target structure complexity, architect planning, instruction abstraction, and dialogue structure, offering a comprehensive summary of their respective features. Note that for all rows, MDC features are a superset of the ones for the synthetic datasets.

\input{rw_syn_data}

%% file: rw_syn_data.tex
\subsection{Related Work: Synthetic Data}
\label{sec:rw_syn_data}

Synthetic data has proven beneficial for various tasks, including those mentioned in Section~\ref{sec:rw_task_and_data}. VLN agents, constrained by limited human instruction data and  diversity in training environments, often struggle with complex language grounding and spatial language understanding. \citet{Kamath_2023_CVPR} address this by using synthetic data, achieving SOTA performance on the RxR dataset \citep{ku-etal-2020-room}, while \citet{Wang_2023_ICCV} are similar in spirit and achieve SOTA on the aforementioned CVDN and other benchmarks. Similarly, \citet{Kang_2023_CVPR} demonstrate strong performance gains in low-data regimes for Visual Dialogue \cite{visdial} by leveraging synthetic data.
Synthetic data has also been effective for dialogue systems, particularly those focused on text-only dialogues \citep{kim-etal-2022-generating, bao-etal-2023-synthetic, zhan-etal-2023-turning}. \citet{zhan-etal-2024-going} is an example of extending this to multi-modal dialogue agents, demonstrating the effectiveness of synthetic visual descriptions in enhancing agents' grounding capabilities. We have also seen synthetic data used to improve VLM spatial reasoning. \citet{sparkel} do so by constructing simple images and fine-tuning over simple tasks such as 2D object localization, distance estimation, navigations, etc to improve spatial reasoning capabilities in 3D images.

\subsubsection*{Synthesizing Task-Oriented Embodied Dialogues}
\label{sec:syn_tod_embodied}
However, fewer efforts have addressed synthesizing complete task-oriented embodied dialogues (involving both utterances and environment actions). \citet{padmakumar-etal-2023-multimodal} were the first to design a framework to do so. They extend agenda-based dialogue simulation \citep{4806280} to a multimodal embodied agent framework, and demonstrate the impact of the synthetic dialogues on the TEACh task (see Section~\ref{sec:rw_task_and_data} for a comparison of TEACh to MDC). Our work serves to provide another useful example in this relatively underexplored general domain, further advancing investigation of synthetic data generation for such complex tasks. 
\citet{dan-etal-2021-compositional-data} is another closely related work that demonstrates the potential of synthetic data in the Blocks World domain. They use simpler synthetic data for the benchmarks of \citet{bisk-etal-2016-natural}, which require understanding single-shot instructions that transform one world state to another using simulated 3D blocks. Blocks are viewed from a fixed bird’s eye perspective, initialized randomly in the initial world state, and uniquely identifiable. The varying Builder perspective and lack of easily identifiable referents, along with the need to understand utterances in a dialogue context, make the BAP task a much more challenging problem.\footnote{As noted in Section~\ref{sec:features_mcbt}, unlike traditional Blocks World, Minecraft allows blocks to float (requiring non-monotonic action sequences where placement is followed by removal), or attach to any side of an existing block.} Our work on synthetic data thus serves as a demonstration of the potential for richer, embodied task-oriented dialogue tasks that extend beyond simpler Blocks World settings.

\subsubsection*{LLM-generated Synthetic Data}
\label{sec:llm_for_syn_data}
While other domains have adopted LLM-generated synthetic data, embodied task-oriented dialogue has generally not followed suit. LLMs are able to generate diverse dialogues \citep{samarinas2024simulatingtaskorienteddialoguesstate} or dialogue counterparts to an existing modality \citep{zhang2025proactiveassistantdialoguegeneration}, but they struggle to generate the real-world environment and track state-changes within it, even on the simpler un-embodied task-oriented dialogue settings \citep{li2024largelanguagemodelszeroshot, feng-etal-2023-towards, kulkarni-etal-2024-synthdst, hudecek-dusek-2023-large}. Generating dialogue from scratch (as \citet{du2025dflowdiversedialogueflow} do) requires complex systems of planning and synthesis that are extremely domain-specific and must be evaluated by humans to verify coherence, flow, and accuracy. Our work, therefore, utilizes a programmatic simulation framework to overcome such limitations. This method guarantees that all generated data is accurately grounded in the game's world state via \textit{highly nuanced geometric computations} (Section~\ref{sec:grounding_gen}), providing a reliable and systematic way to generate high-fidelity training data for complex, embodied tasks like BAP.

%% file: incorporate_syn_data.tex

\section{Revisiting BAP Modeling: Training with Synthetic Data}
\label{sec:syn_data_training}

Our next objective is to demonstrate that the synthetic data can help train much better models for the BAP task, even with straightforward training methods, thereby addressing the challenge of insufficient training data, and thus validating one of the motivations behind our synthetic data.
We demonstrate this for both non-LLM- and LLM-based models, highlighting the robustness of our approach.
The synthetic data together with the original BAP training data now constitutes the BAP v2 training data in the BAP v2 framework.

\paragraph{Key Terminology}
Recall that we refer to the
synthetic datasets as \dr, \dbs and \dss. The Minecraft-based original BAP data is denoted as \dmc. 
Each dataset includes train, test, and val splits.
The GRU-based baseline model (Section~\ref{sec:baseline_model_desc}), trained on \dmc alone, is denoted as \mmc.

\subsection{The GRU-based models}
\label{sec:joint_training}
The most straightforward approach to use the synthetic datasets (\syndata) is to just combine them all together along with the original \dmc data and retrain the GRU-based baseline model on this aggregated dataset.

\subsubsection{Experimental Setup}
\label{sec:exp_setup_batch_training}
Our setup closely follows the one used for the baseline model \mmc when it was trained on \dmc alone (Section~\ref{sec:baseline_model_desc}).
Training employs teacher forcing with cross-entropy loss, and decoding is performed via greedy decoding, with a maximum sequence length of 20 actions.
Further details are provided in the appendix (Section~\ref{app:exp_setup_gru}).
We use the fairer F1 metric to report micro F1 scores on the BAP v2 test set for \dmc (Section~\ref{sec:v2_benchmark}) and the test sets of \syndata. \footnote{For fairer F1 on \syndata, \( \mathcal{M} \) consists of all EB items, unlike \dmc, where only a subset is considered (Section~\ref{sec:fairer_f1}). This is because, by design, all EB items in \syndata have multiple correct interpretations.} 

\subsubsection*{How much of each synthetic dataset to use during train time?}
\label{sec:syn_data_props}
An important consideration is determining the optimal amount of each synthetic dataset to use alongside \dmc in the training dataset. We adopt an empirical yet systematic approach to address this, yielding a data mix that optimizes \dmc performance while also being highly performant on the synthetic data as much as possible. More details are provided in the appendix (Section~\ref{app:data_mix}).
Table~\ref{tab:results7} presents the final data statistics. 
Similar to the methodology used for data splits for the MDC ~\citep{narayan-chen-etal-2019-collaborative}, for each of the synthetic datasets, the game logs are split into disjoint test, training, and validation sets such that training target structures do not appear during test or validation.
We now have a total of 52419 items in the aggregated training dataset.

\begin{table}[htbp]
    \centering
    \begin{small}
    \begin{tabular}{lccc}
        \toprule
        \textbf{Dataset} & \textbf{Train} & \textbf{Val} & \textbf{Test} \\\midrule
        \dbs & 9890 & 1186 & 1181 \\
        \dss & 11868 & 1000 & 1000 \\
        \dr & 15825 & 1161 & 1089 \\
        \dmc & 14836 & 1331 & 1616 \\
        \bottomrule
    \end{tabular}
    \end{small}
    \caption{Data statistics (\#items) for training, validation, and test splits across datasets.}
    \label{tab:results7}
\end{table}

\subsubsection{Experiments and Results}
\label{sec:exps_and_results}
Table~\ref{tab:results2} report overall F1 performance  on the respective test sets of all four datasets (\alldata) for all variants of our GRU models. We denote the model trained on the aggregated dataset as \magg, and include comparisons to the baseline model (\mmc, trained on \dmc), and models trained on the three synthetic datasets separately (\mbs, \mss, \mr,  trained on \syndata respectively). 

\magg achieves an F1 score of 30.3 on \dmc, up from the baseline model \mmc’s 27.3, a notable 3- point improvement. Additionally, \magg is highly performant across the synthetic datasets (\syndata), surpassing \mmc by large margins on these (83.4 vs. 35.2 on \dbs, 68.8 vs. 19.1 on \dss, and 63.6 vs. 12.4 on \dr). It is also informative to compare both the \mmc and \magg numbers with the corresponding diagonal entries in the  \mbs, \mss and \mr rows, which report the performance of these models when trained only on these datasets:  83.9 for \dbs, 75.4 for \dss, and 64.4 for \dr with the baseline \mmc numbers. 
\mmc clearly underperforms significantly on the synthetic datasets, despite being trained on the far more complex \dmc data; it also performs lower on datasets \dss and \dr than on \dmc. In contrast, \magg is more robust, and retains much of the diagonal-entry performance on the synthetic datasets while also significantly improving on \dmc. 
Lastly, the models trained solely on synthetic datasets perform poorly on \dmc, as does \mmc on synthetic data, indicating that joint training on the aggregated dataset is essential for robust performance across both real and synthetic data.

\begin{table}[htbp]
    \centering
    \begin{small}
    \begin{tabular}{llcccc}
        \toprule
       \multirow{2}{*}{\textbf{Model}} & \multirow{2}{*}{\textbf{Train Data}} & \multicolumn{4}{c}{\textbf{Test Data}} \\\cmidrule(lr){3-6}
                       &                                      & \textbf{\dbs} & \textbf{\dss} & \textbf{\dr} & \textbf{\dmc} \\\midrule
        \textbf{\mmc}    & \textbf{\dmc}                         & 35.2       & 19.1       & 12.4       & 27.3       \\\midrule
        \textbf{\mbs}     & \textbf{\dbs}                          & 83.9       & 12.9       & 44.9       & 6.3        \\
        \textbf{\mss}     & \textbf{\dss}                          & 11.3       & 75.4       & 7.9        & 9.1        \\
        \textbf{\mr}     & \textbf{\dr}                          & 43.8       & 10.1       & 64.4       & 7.7        \\\midrule
        \textbf{\magg} & \textbf{\alldata}                      & 83.4       & 68.8       & 63.6       & 30.3       \\
        \bottomrule
    \end{tabular}
    \end{small}
    \caption{Models trained on datasets separately (\mmc, \mbs, \mss, \mr) and together (\magg)}
    \label{tab:results2}
\end{table}

\subsection{LLMs (Llama)}
\label{sec:llama_model_desc}
To demonstrate the robustness of our synthetic data and its applicability to contemporary Large Language Models (LLMs), we conduct experiments with Llama \citep{dubey2024llama}. Specifically, we consider the Nebula model recently proposed in \citet{chaturvedi-etal-2024-nebula} that established a new SOTA on the BAP task and aim to further improve upon it. We refer to it as our \textbf{LLM-based baseline model}. 
It is a Llama-3-8B model that takes as context all conversation and action sequences up to the action sequence \(A\) to predict \(A\) in a text-to-text fashion. It is finetuned on the Minecraft data \dmc using QLoRA \citep{dettmers2023qlora}.

\subsubsection{Experiments and Results}
\label{sec:llama_exp_setup}
We conduct experiments analogous to those performed with the GRU-based models, along with additional experiments on the input textual representation provided to the LLM. The results are detailed in Table~\ref{tab:llama_nebula_results}. All models involve finetuning Llama-3-8B using QLoRA. 

\subsubsection*{LLM-based baseline:} First, we reproduce the Nebula LLM-based baseline by fine-tuning Llama-3-8B on \dmc but this time evaluating it on \dmc using the v2 evaluation as well as on the synthetic data.
As shown in the first row of Table~\ref{tab:llama_nebula_results}, this model (\mmcp) achieves an F1 score of 47.0 on \dmc, significantly better than its GRU-based counterpart \mmc (27.3). However, similar to \mmc, it underperforms on the easier synthetic data (\syndata), despite being trained on the far more complex \dmc, albeit its performance is better than that of \mmc. Thus, this \textbf{counterintuitive behavior of the GRU-based model still persists even with the far more advanced LLM-based model.
}
\subsubsection*{Joint training on aggregated data} Next, we perform joint training on the aggregated dataset (\alldata) to produce model \maggp. This yields substantial improvements on \syndata, as expected, but only a marginal gain on \dmc (47.8 compared to 47.0. Furthermore, its performance on the synthetic datasets still lags behind its GRU-based counterpart, \magg.

\begin{table*}[ht]
\centering
\scriptsize
\begin{tabular}{l}
\toprule
\textbf{Input Representation: \repV} \\
\midrule
\texttt{Predict the action sequence (AS) for the Minecraft excerpt:} \\
\texttt{...} \\
\texttt{<Architect> place one in the center of one of the lines near the edge} \\
\texttt{place yellow 5 1 0} \\
\texttt{...} \\
\texttt{<Architect> good; repeat to make a V} \\
\texttt{place yellow 3 1 -1} \\
\texttt{place yellow 3 1 1} \\
\texttt{<Architect> still diagonal from the previous squares} \\
\texttt{<Builder> extended?} \\
\texttt{...} \\
\texttt{<Architect> not really; the first step is orange, but it is in line with the yellow blocks} \\
\texttt{pick 3 2 1} \\
\texttt{pick 3 1 1} \\
\texttt{pick 2 2 2} \\
\texttt{...} \\
\texttt{<Architect> now one orange on top of teh second orange} \\
\texttt{place orange 1 2 2} \\
\texttt{<Architect> now delete the second orange} \\
\texttt{pick 1 1 2} \\
\texttt{<Architect> good repeat to make a third step} \\
\texttt{\#\#\# AS:} \\
\midrule
\textbf{Input Representation: \repVPosB} \textit{(appends builder's position and orientation to the input above)} \\
\midrule
\texttt{...} \\
\texttt{Builder's current position is 0.2 1.0 -0.4 and orientation (yaw) is -6.5 degrees} \\
\texttt{\#\#\# AS:} \\
\midrule
\textbf{Input Representation: \repVPosBS} \textit{(appends current built structure to the input above)} \\
\midrule
\texttt{...} \\
\texttt{Current built structure is:} \\
\texttt{ yellow 5 1 0} \\
\texttt{ yellow 4 1 1} \\
\texttt{ yellow 4 1 -1} \\
\texttt{ yellow 3 1 2} \\
\texttt{ yellow 3 1 -2} \\
\texttt{ orange 2 1 2} \\
\texttt{ orange 1 2 2} \\
\texttt{\#\#\# AS:} \\
\midrule
\textbf{Output (Common to all inputs)} \\
\midrule
\texttt{place orange 1 3 2} \\
\texttt{place orange 0 3 2} \\
\texttt{pick 1 3 2} \\
\bottomrule
\end{tabular}
\caption{The three different input representations (\repV, \repVPosB, \repVPosBS) and the (common) ground truth output. Ellipses (...) denote truncation for readability. \repV is from \citet{chaturvedi-etal-2024-nebula}, and \repVPosB, \repVPosBS are ours.}
\label{tab:model_inputs_outputs}
\end{table*}

\subsubsection*{Richer input representations} 
We hypothesize that the lower-than-expected results of \maggp stems from the input representation, which \textbf{lacks a critical component of the game context} available to the GRU models -- the Builder's position and orientation, \(\textsc{Pos}_\B\), which is essential for correctly interpreting spatial relations (Section~\ref{sec:bpos}). The GRU models incorporate \(\textsc{Pos}_\B\) \citep{jayannavar-etal-2020-learning}, making the comparison with the LLM models uneven. We therefore enrich the input text by appending information about \(\textsc{Pos}_\B\), as illustrated in Table~\ref{tab:model_inputs_outputs} (representation \repVPosB). We denote the original input representation used in Nebula as \repV and show that at the very top of the table. (The ground truth output representation is shown at the very bottom.)

Training on the aggregated data with this richer input yields model \maggpbpos, which shows significant improvement on \dmc and even more on the synthetic data, in comparison to \maggp. It achieves F1 scores of 93.4 on \dbs, 84.9 on \dss, and 82.3 on \dr, while also boosting the \dmc score to 52.5, and now also overtakes the performance of the GRU models on all datasets, validating out hypothesis that suitable input representations are essential for this task. 

However, to verify that this gain is not from the representation alone, we trained a model (\mmcpbpos) on just \dmc but with the new representation. This results in an F1 score of 47.6 on \dmc, a marginal improvement over the 47.0 from its counterpart \mmcp that used the original Nebula representation. This confirms that \textbf{both the enriched input representation and the synthetic data are necessary to achieve significant performance gains on both synthetic and real Minecraft data; each on their own are not enough}. The explicit spatial context in the input allows the model to better leverage the patterns in the synthetic data and transfer that knowledge to the \dmc domain.

We also note that while our textual representation of \(\textsc{Pos}_\B\) is fairly straightforward, its effect on performance is very substantial. This input only explicitly provides the Builder's frame of reference, but for the model to correctly ground spatial instructions, it must "infer" a more complex, situated understanding of the 3D world state from this frame. Our results demonstrate that the LLM seems to become, to a significant extent, capable of this complex inference, just through the textual signal in the data and standard training (teacher forcing and cross entropy loss). Crucially, however, this ability due to \(\textsc{Pos}_\B\) only emerges when the model is trained on enough signal from the rich spatial examples in our synthetic data in conjunction with the original \dmc.

\subsubsection*{A further boost} 
Finally, we investigate whether an \textbf{explicit representation of the current world state/built structure \(S\)} (like  in the GRU-based models) helps further. While the action history contained in the input text already "implicitly" defines \(S\), we append an explicit version of it to the input (representation \repVPosBS in Table~\ref{tab:model_inputs_outputs}). Joint training on the aggregated data with this input produces model \maggpbposws which further boosts performance on \dmc by a bit to 53.0 (up from the 52.5 of \maggpbpos), while performance on \syndata continues to remain strong and comparable to \maggpbpos. 

This result suggests that for the longer and more complex dialogues in \dmc, where tracking and inferring the world state/built structure \(S\) from just the textual action history is more challenging, an explicit representation of \(S\) provides a beneficial signal. On the shorter and simpler dialogues in the synthetic datasets, however, this inference is not that challenging, and therefore explicit state information may not be as necessary. This likely explains why adding in this representation only improves performance on \dmc. Moreover, given the already strong performance on the synthetic data after including \(\textsc{Pos}_\B\), further gains from additional inputs such as \(S\) may anyway be subject to diminishing returns on the synthetic data and the transfer of knowledge to the \dmc domain (unlike for \maggpbpos as discussed above).

We designate \maggpbposws as our \textbf{best model}. As it makes use of richer game context and synthetic data (with each one enabling the other), we name it \textbf{\llmbest} (\textbf{C}ontext \textbf{R}ich \textbf{A}nd \textbf{F}ine-\textbf{T}uned On \textbf{S}ynthetic Data). This model establishes a \textbf{new state of the art (SOTA) on the BAP task (\dmc)}, achieving an F1 score of 53.0. This represents a \textbf{6 point improvement} over the 47.0 F1 score of the previous SOTA, the Nebula (the LLM baseline).

\subsection{Broader Takeaways}

\subsubsection*{Robustness of our approach}
By providing the LLM-based models with an input representation that is more comparable to the one used by the GRU-based models, we show that our synthetic data significantly improves performance of both non-LLM- and more modern LLM-based models on both the synthetic and original Minecraft data. This highlights the \textbf{robustness of our data and simple training methodology} across different model architectures.

\subsubsection*{Motivating future work}
With the fairer v2 evaluation now possible, we can more accurately gauge current performance on the BAP task and the remaining gap for improvement. While \llmbest's 53.0 F1 score marks improved performance, it also underscores that \textbf{even with SOTA LLMs, significant room for improvement remains}, thus reinforcing the task's challenging nature and its importance for future research.

\subsubsection*{Importance of evaluation on the synthetic data}
\label{sec:syn_data_eval}
Going forward, evaluation on the synthetic data will also be important for making more meaningful progress on the BAP task. 
First, the more structured and simpler nature of the synthetic data allows us \textbf{to assess more basic model competencies} than is possible on the more complex \dmc. Second, as noted above, both GRU and LLM baselines \mmc and \mmcp underperform significantly on the synthetic datasets, despite being trained on the far more complex \dmc; they also perform lower on datasets \dss and \dr than on \dmc. 
This counterintuitive behavior shows that \textbf{models (including SOTA LLMs) trained solely on \dmc may not be as robust}.
Evaluating on the synthetic data \textbf{will help build models that do not show such behavior} (like we have shown by using the data effectively for training), and \textbf{complements evaluation on \dmc}. In fact, future models should aim to surpass previous performance on \dmc while also achieving strong results on synthetic data, ideally improving on both.

\begin{table}[htbp]
    \centering
    \begin{small}
    \setlength{\tabcolsep}{5pt}
    \begin{tabular}{l l l cccc}
        \toprule
  %
        \multirow{2}{*}{\textbf{Model}} & 
        \multirow{2}{*}{\textbf{Input}}
        & \multirow{2}{*}{\textbf{Train Data}} & \multicolumn{4}{c}{\textbf{Test Data}} \\\cmidrule(lr){4-7}
    & & &     \textbf{\dbs} & \textbf{\dss} & \textbf{\dr} & \textbf{\dmc} \\\midrule
        \textbf{\mmcp} (\llmbaseline) & \repV & \textbf{\dmc} & 49.5 & 31.9 & 33.3 & 47.0 \\\midrule
        \textbf{\maggp} & \repV & \textbf{\alldata} & 75.5 & 54.9 & 43.8 & 47.8 \\
        \textbf{\mmcpbpos} & \repVPosB & \textbf{\dmc} & 49.7 & 32.2 & 32.0 & 47.6 \\
        \textbf{\maggpbpos} & \repVPosB & \textbf{\alldata} & 93.4 & 84.9 & 82.3 & 52.5 \\\midrule
        \textbf{\maggpbposws} (\llmbest) & \repVPosBS & \textbf{\alldata} & 92.3 & 83.7 & 82.2 & \textbf{53.0} \\
        \bottomrule
    \end{tabular}
    \end{small}
    \caption{Llama performance with different datasets and input representations.}
    \label{tab:llama_nebula_results}
\end{table}

\input{rw_modeling}

%% file: rw_modeling.tex
\subsection{Related Work: Role of specialized datasets and smaller finetuned models}
\label{sec:rw_modeling_specialized}
These results highlight a key contemporary question in applied NLP: the relative merits of fine-tuning smaller, specialized models versus leveraging large, general-purpose models via ICL. The performance parity between Nebula and GPT-4 on this task (Section~\ref{sec:rw_llms_bap}) suggests that a smaller, fine-tuned specialist model can be as effective as a much larger generalist one. This finding coupled with the fact that we are able to improve the former by a substantial 12.8\% via leveraging synthetic data, motivates a deeper exploration of the fine-tuning paradigm and the role of specialized datasets, especially concerning the most effective pathways to further improve performance.

While generalist models offer impressive few-shot capabilities, avenues for their improvement often center on prompt engineering, increased test-time computation, or scaling to even larger (and often proprietary) models, which may not always be practical. In contrast, the fine-tuning approach offers more vectors for enhancement, e.g., synthetic data (like we show), better training regimes, etc. in addition to the ones for ICL.  
Furthermore, smaller specialized and open models present practical advantages in terms of computational efficiency and accessibility for both academic research and deployment in resource-constrained settings. For instance, in real-world counterparts of BAP such as instruction-following robots, stringent low-latency requirements often render the extensive test-time computation demanded by larger models infeasible. This pragmatic consideration is supported by emerging theoretical work suggesting that fine-tuning is not merely a matter of convenience. Studies indicate that fine-tuned models can achieve generalization capabilities comparable to ICL \citep{mosbach-etal-2023-shot, zhao2025is, ponce2025incontextlearningvsinstruction} and may learn representations that are more efficiently embedded in lower-dimensional manifolds, potentially better capturing real-world data/features \citep{janapati-ji-2025-comparative}.

Finally, the development of specialized datasets need not be an endeavor exclusive to fine-tuning. The synthetic data we have generated, rich in specific linguistic and spatial phenomena, could also potentially enhance ICL performance. For instance, it could be used to select more effective few-shot exemplars, drawing inspiration from works like \citet{kulkarni-etal-2024-synthdst}, or be integrated into curriculum learning frameworks for ICL, as explored by \citet{liu2024letslearnstepstep}, given that the synthetic data is simpler than the original BAP data. This suggests that creating high-quality, task-specific data is a valuable pursuit that can advance multiple modeling paradigms.

%% file: analysis.tex
\section{Model Analysis}
\label{sec:analysis}


We now analyze the performance of the best LLM-based model \llmbest (\maggpbposws) relative to the LLM-based baseline \llmbaseline (\mmcp).
In Section~\ref{sec:quant_eval_analysis}, we do so using the BAP v2 evaluation benchmark introduced in Section~\ref{sec:v2_benchmark}. Section~\ref{sec:qual_eval_analysis} complements this with a qualitative evaluation of outputs from both models. 

\subsection{Quantitative evaluation}
\label{sec:quant_eval_analysis}

\begin{table}[htbp]
\centering
\begin{small}
\begin{tabular}{llccccc}
\toprule
\textbf{Dataset} & \textbf{Model} & \textbf{Type} & \textbf{Color} & \textbf{Location} & \textbf{Shape} & \textbf{Overall} \\
\midrule
\multirow{2}{*}{EB} & \llmbaseline  & 92.6 & 92.0 & 74.7 & 80.9 & 74.2 \\
 & \llmbest  & 94.7 & 94.7 & 80.3 & 85.8 & 80.3 \\
\midrule
\multirow{2}{*}{NEB} & \llmbaseline  & 78.8 & 77.6 & 44.3 & 57.1 & 43.5 \\
 & \llmbest  & 80.8 & 79.5 & 50.4 & 59.4 & 49.4 \\
\midrule
\multirow{2}{*}{Overall} & \llmbaseline  & 80.4 & 79.2 & 47.7 & 59.8 & 47.0 \\
 & \llmbest  & 82.4 & 81.2 & 53.8 & 62.4 & 53.0 \\
\bottomrule
\end{tabular}
\end{small}
\caption{F1 scores for the LLM-based baseline \llmbaseline and  \llmbest}
\label{tab:micro_best_v_baseline_new}
\end{table}

As can be seen from Table~\ref{tab:micro_best_v_baseline_new}, \llmbest outperforms \llmbaseline across all metrics on the v2 benchmark, reaching an overall F1 of 53.0, a notable 6.0 points improvement over \llmbaseline’s 47.0. This increase is primarily driven by an improvement in Location F1, which similarly rises from 47.7 to 53.8. However, the BAP task remains far from being solved. Although the empty board (EB) can be seen as the most basic variant of the BAP task (Section~\ref{sec:v2_benchmark}), the best model's 80.3 (albeit a substantial increase from the baseline’s 74.2) is a clear indication that there remains much room for improvement.

\subsection{Qualitative evaluation}
\label{sec:qual_eval_analysis}
\newcommand{\exa}{Example A}
\newcommand{\exb}{Example B}
\newcommand{\exc}{Example C}
\newcommand{\exd}{Example D}
\newcommand{\exe}{Example E}
\newcommand{\exf}{Example F}
\newcommand{\exg}{Example G}

\begin{figure*}[!htbp]
\centering
\begin{tabular}{l}
\toprule
\textbf{Scenario 1: One model is correct, the other makes mistakes}
\\
\midrule\textbf{\exa}\\
\midrule
\\
\includegraphics[width=.8\textwidth]{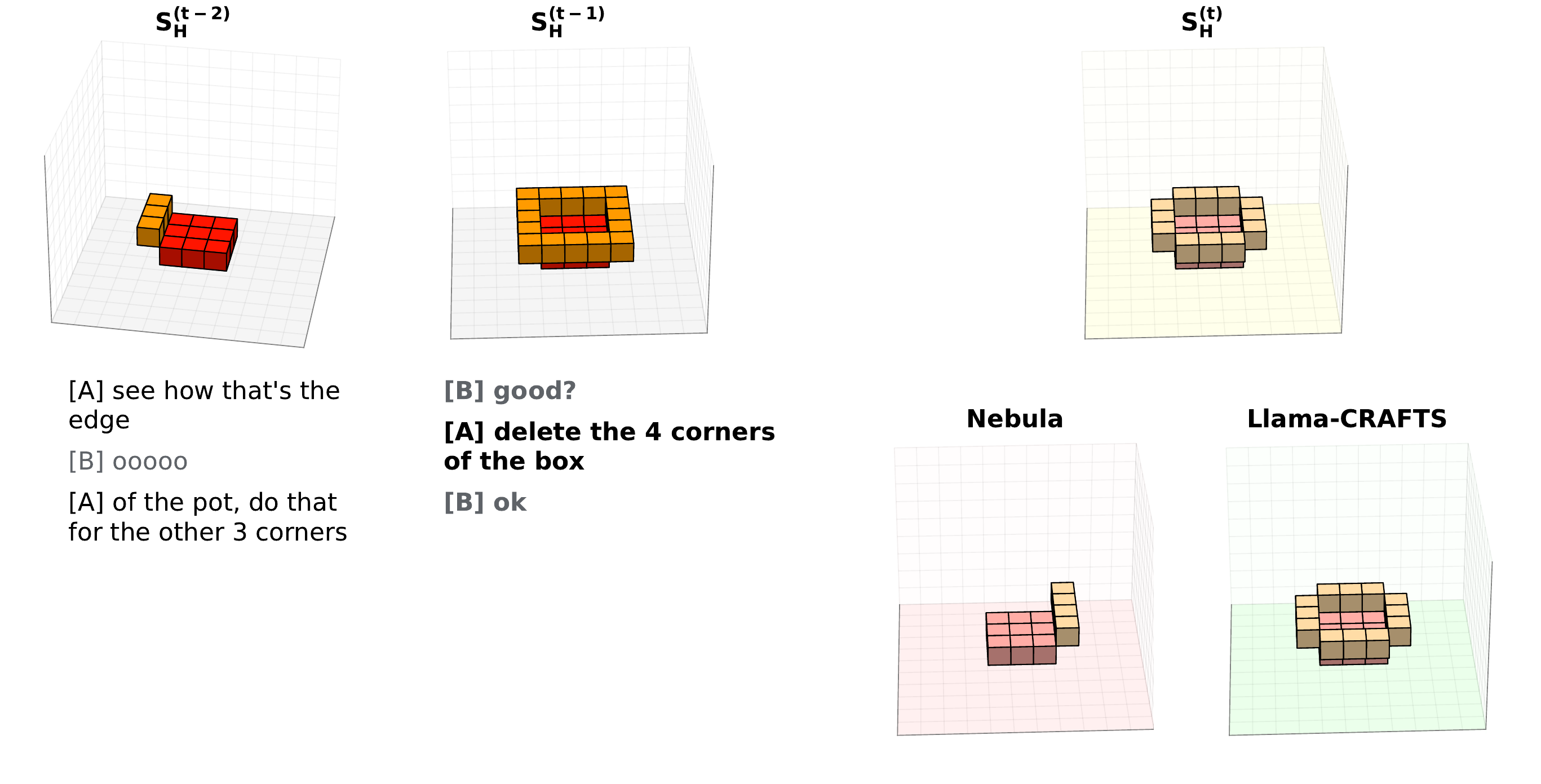}\\
\midrule
\textbf{\exb}\\
\midrule
\includegraphics[width=.8\textwidth]{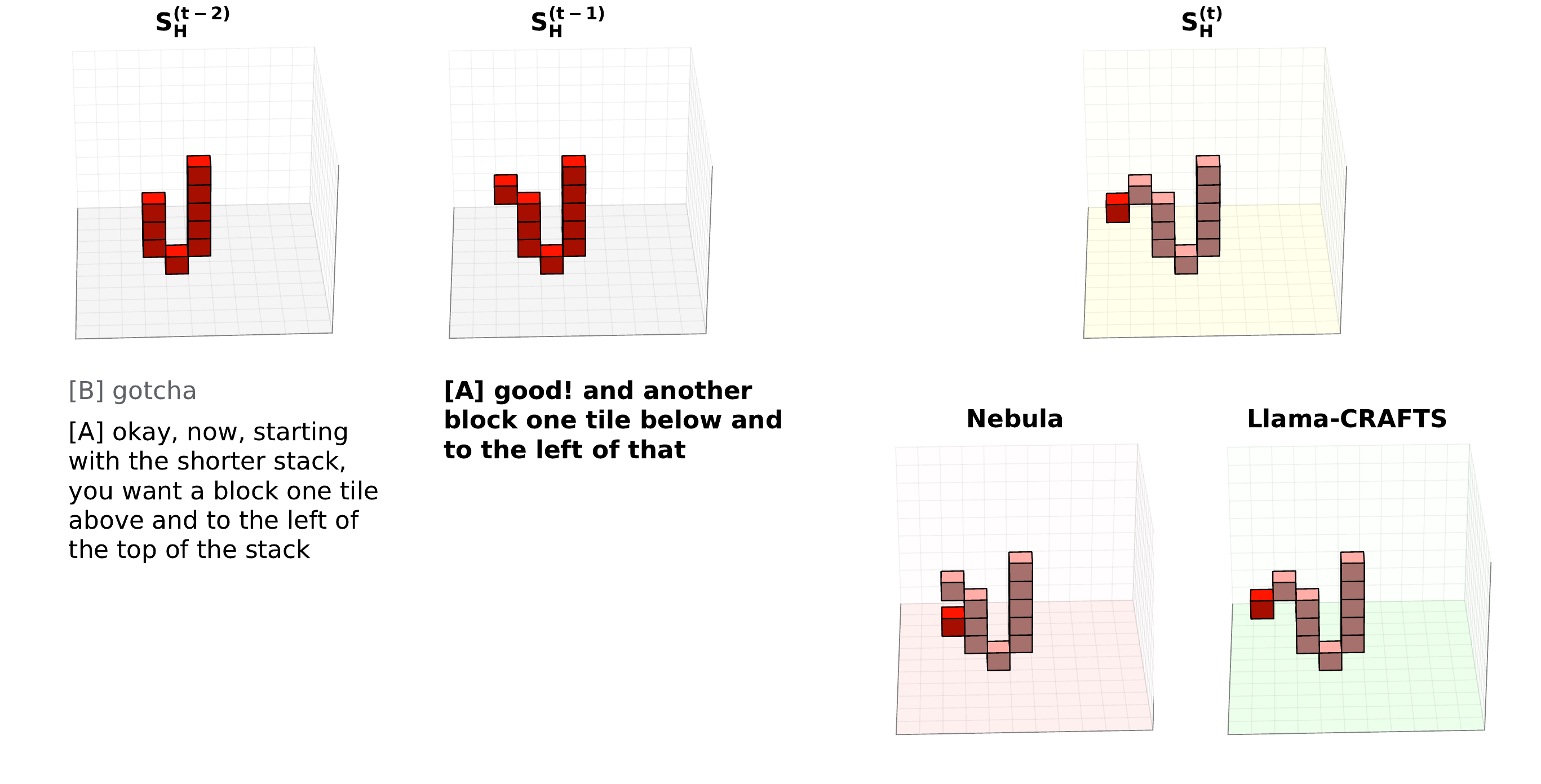}
\\
\midrule
\textbf{\exc}\\
\midrule
\includegraphics[width=.8\textwidth]{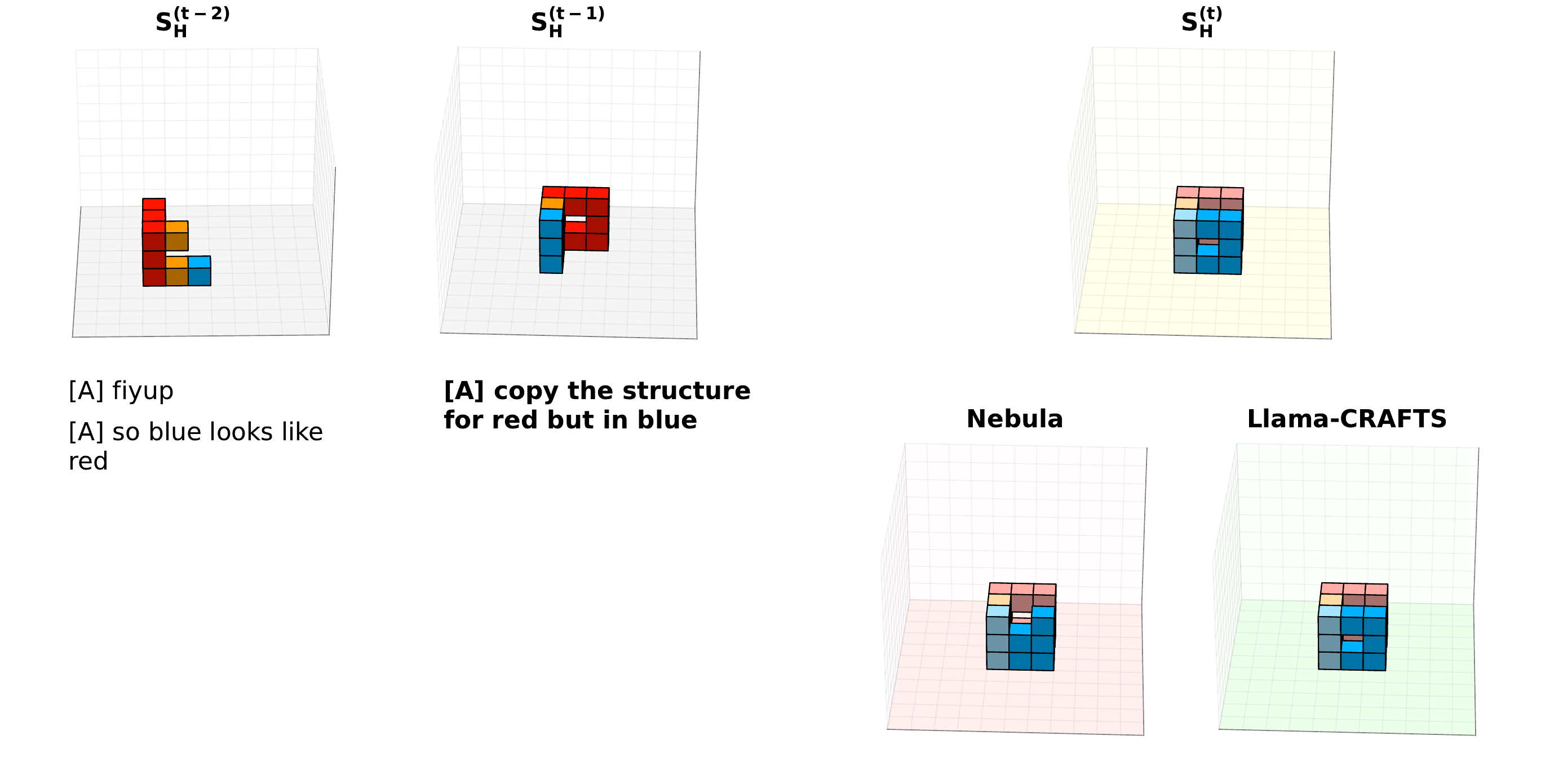}\\
\bottomrule
\end{tabular}
\caption{The overall F1 score is sufficient to distinguish between correct and incorrect model outputs (Scenario 1)}
\label{fig:qual_eval1}
\end{figure*}

\begin{figure*}[!htbp]
\centering
\begin{tabular}{l}
\toprule
\textbf{Scenario 2: The auxiliary metrics provide further insight}\\
\midrule
\textbf{\exd}\\
\midrule
\includegraphics[width=.8\textwidth]{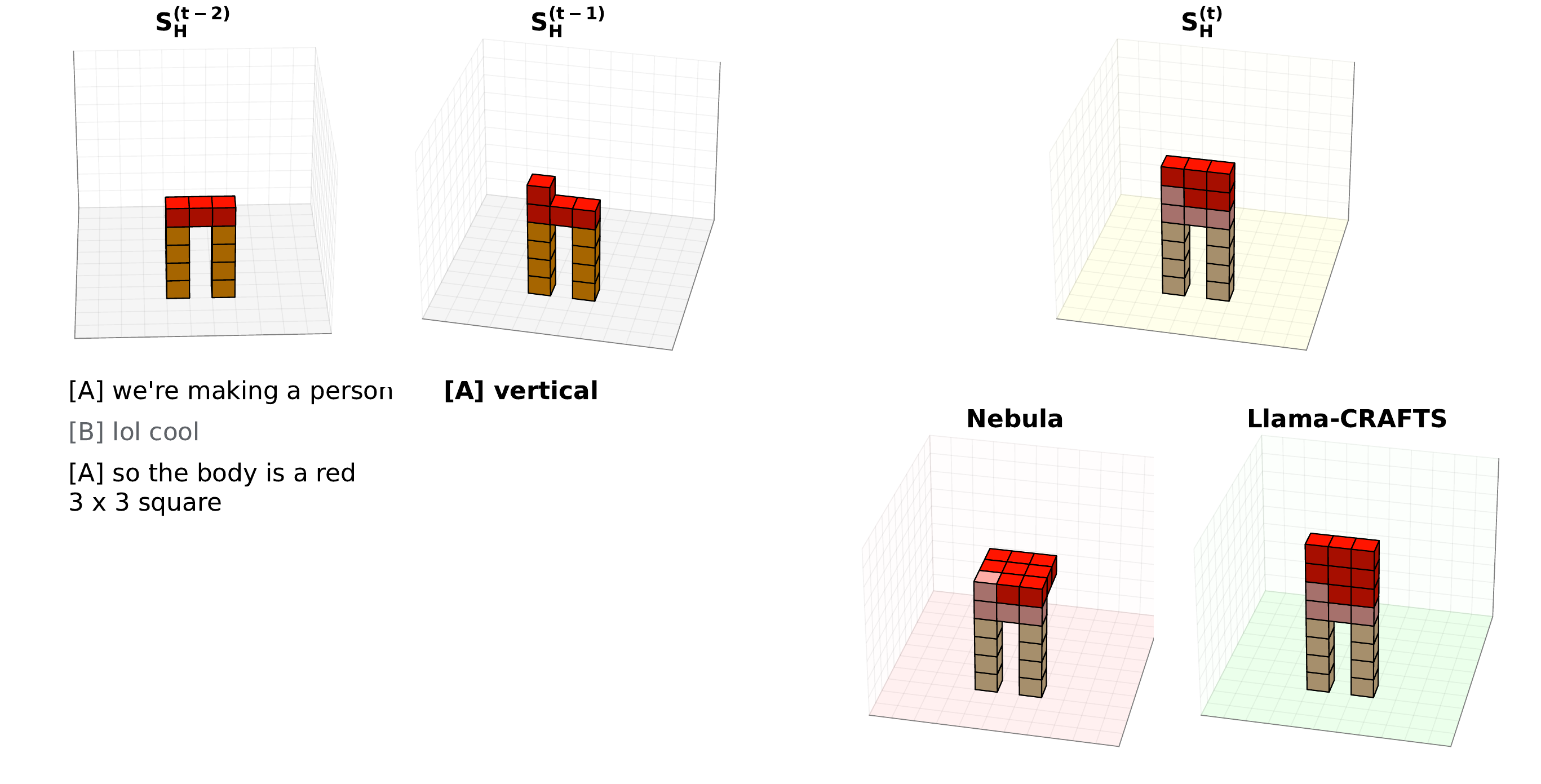}\\
\midrule 
\textbf{\exe}\\
\midrule
\includegraphics[width=.8\textwidth]{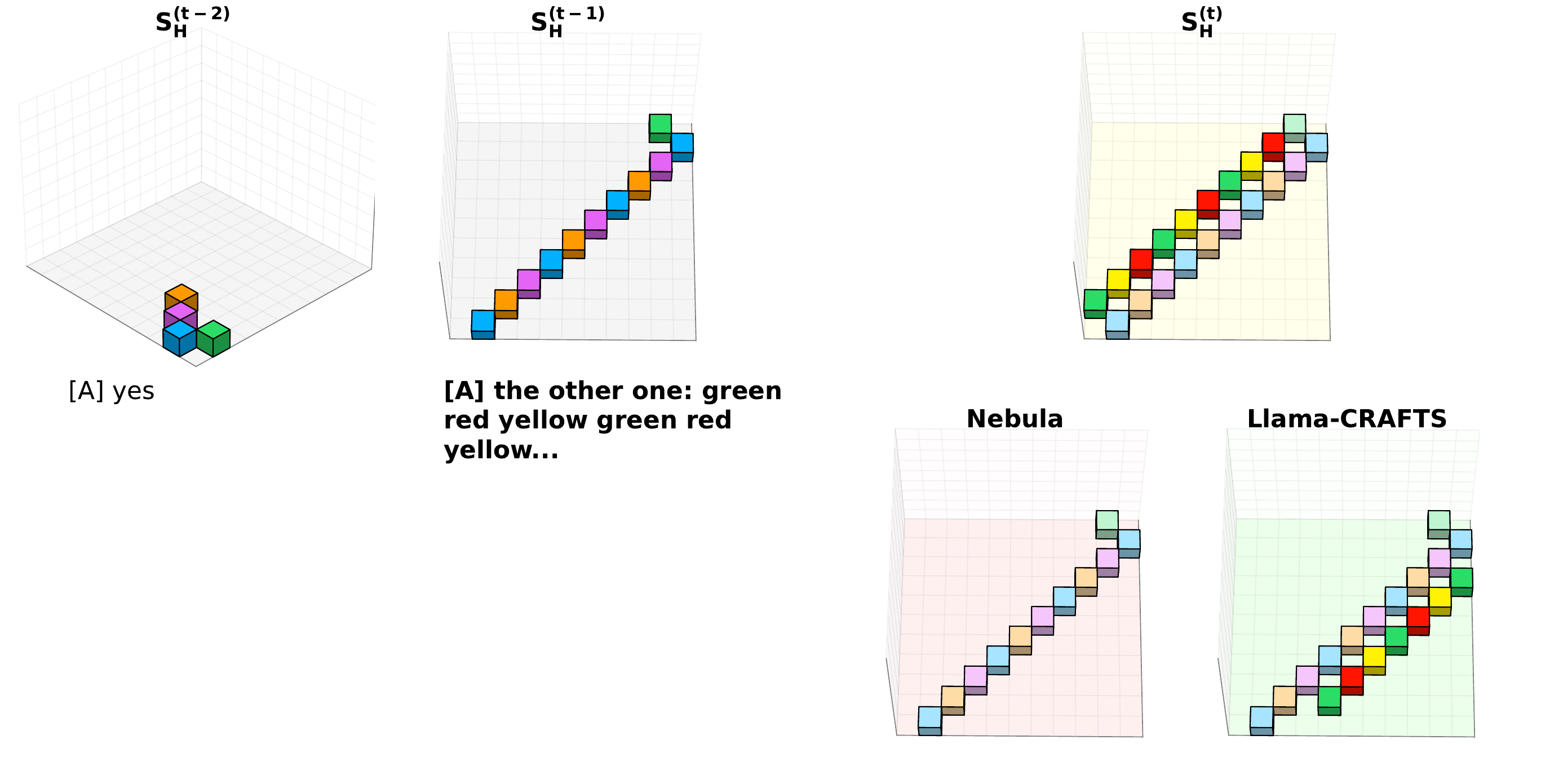}\\
\bottomrule
\end{tabular}
\caption{Auxiliary metrics provide further insight than the overall F1 score (Scenario 2)}
\label{fig:qual_eval2}
\end{figure*}

\begin{figure*}[!htbp]
\centering
\begin{tabular}{l}
\toprule
\textbf{Scenario 3: Mistakes that warrant a more semantic evaluation}\\
\midrule
\textbf{\exf}\\
 \midrule
 \includegraphics[width=.8\textwidth]{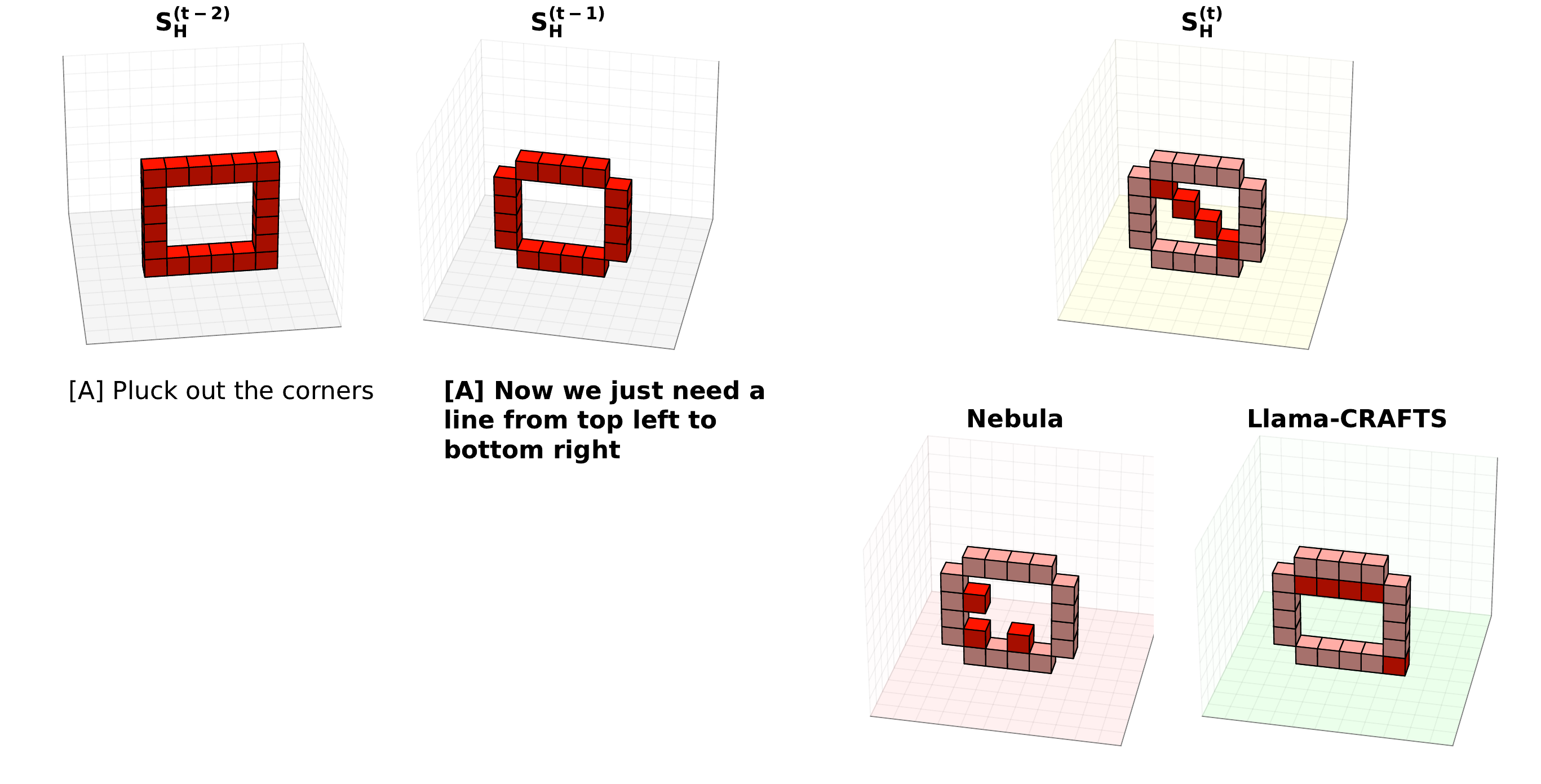}\\
\midrule
\textbf{\exg}\\
\midrule
\includegraphics[width=.8\textwidth]{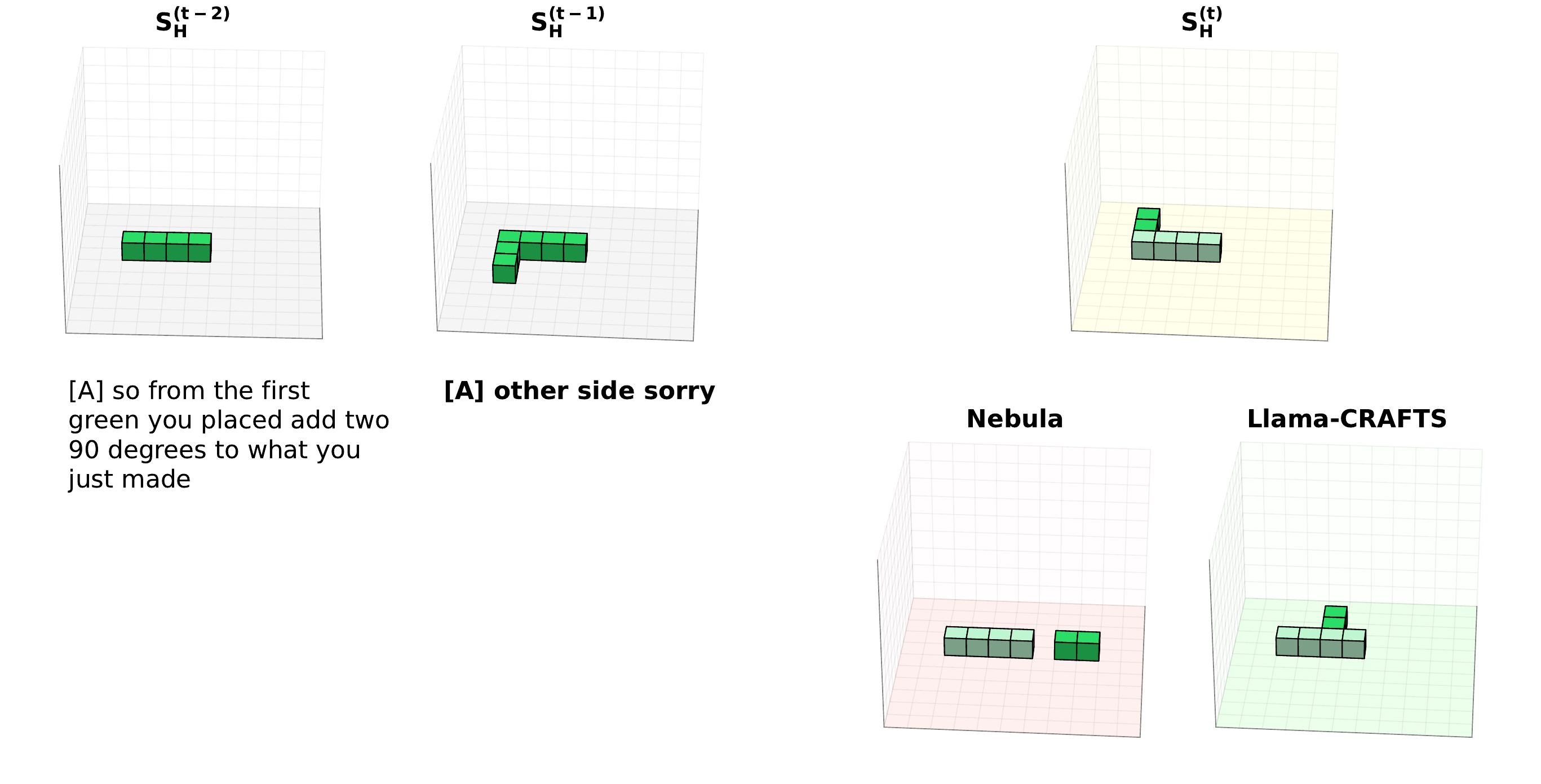}\\
\bottomrule
\end{tabular}
\caption{Examples that warrant a more semantic evaluation (Scenario 3)}
\label{fig:qual_eval3}
\end{figure*}

We now present three different scenarios, i.e. types of examples of output from \llmbest and \llmbaseline  that, on the one hand, illustrate different ways in which \llmbest arguably displays a better spatial understanding, and, on the other hand, allow us to discuss how this is reflected (or not) in our various metrics.

\paragraph{Scenario 1: The overall F1 score is sufficient when one model is correct and the other makes mistakes}
Figure~\ref{fig:qual_eval1} shows three examples that \llmbest gets completely right (receiving perfect scores across all metrics), and \llmbaseline does not.  In \exa, \A instructs \B to remove the four corners of the orange ring. \llmbaseline  removes many extraneous blocks but only three corners (achieving an overall F1 of 0.38). In \exb, \A asks \B to add another {red} block one tile below and to the left of the last block \B had placed. \llmbaseline 
only correctly identifies that a single red block should be placed below the last one, but goes too far down (2 tiles), and misses the second spatial relation (left) altogether, resulting in a perfect type and color F1 scores of 1.0. and an overall F1 of 0.0. In \exc, \A asks \B to \exampleutt{copy} the red ring in blue. Instructing Builders to copy substructures is a natural and efficient strategy that the human Architects in the MDC employ frequently. \llmbaseline achieves perfect type and color scores, but misplaces one of five blocks, resulting in location and shape scores of 0.8. In these clear cases, the strict F1 metric is obviously sufficient to capture the qualitative difference between the outputs.

\paragraph{Scenario 2: Both models make mistakes, but the auxiliary metrics provide further insight}
Figure~\ref{fig:qual_eval2} shows examples where both models make mistakes (although \llmbest's mistakes are arguably less egregious), and illustrates how our metrics capture different aspects of this task. 
In \exd, \B is asked to make \exampleutt{a red 3 x 3 square} that is \exampleutt{vertical} (and presumably includes the four red blocks already placed). Both models add eight red blocks to create squares that leave out the bottom three red blocks and only include the red block above. Since \llmbest's square is actually vertical, it achieves a score of 0.77 on all metrics, whereas \llmbaseline built a horizontal square, and hence achieves overall, shape and location scores of 0.31 (and type and color scores of 0.77). That is, although the overall score correctly identifies 
\llmbest as the better model, it is the auxiliary metrics that provide additional insight and assign credit to \llmbaseline. 
In \exe, \B has just finished the blue-purple-orange (from right to left) line, and \A references earlier instructions  when saying that  \exampleutt{the other one: [should be like the one \B already made but with] green red yellow [instead of blue purple orange] ...}. Based on the earlier dialogue (also not shown in the figure), this second line should include the green block already on the board.\footnote{This example illustrates a shortcoming of our figures: although BAP models have access to the entire preceding context, the reader does not.}  Here, both models receive a 0.0 overall F1 score, \llmbaseline because it produces no net actions (receiving 0.0 scores on all metrics), and \llmbest because none of its blocks appear in the intended places. However, because \llmbest  actually places a diagonal line of the correct color (albeit on the other side of the previous line and with the pattern starting from the left instead of the right), it receives type, color, and shape scores of 0.88. In this case, the auxiliary metrics are necessary to identify that \llmbest's output is actually quite similar to the desired structure.

\paragraph{Scenario 3: Shortcomings of our metrics}
Figure~\ref{fig:qual_eval3} illustrates shortcomings of our automatic evaluation procedures, and suggests the need for a more complex, semantic evaluation. 
In \exf, \B is asked to create a (diagonal) line of four blocks inside the ring-shaped structure, \exampleutt{from top left to bottom right}. \llmbaseline places three disconnected blocks inside the structure (achieving overall and location scores of 0.0, type and color scores of 0.86, and a shape score of 0.57), while \llmbest places a (horizontal) line of four blocks across the inside top edge, as well as an additional block in the (outside) bottom right corner (achieving overall, location and shape scores of 0.22, and type and color scores of 0.89). The fact that \llmbaseline's shape score is higher than \llmbest's points to a limitation of our shape metric (two of \llmbaseline's blocks can be mapped to the intended diagonal) This example also suggests a need for a more semantic evaluation: none of our metrics capture the fact that, arguably, \llmbest showed at least a partial understanding of the \exampleutt{line} concept (and, in a way, even the \exampleutt{bottom right}), but \llmbaseline did not.
In \exg,  \A asks \B to place two blocks on the \exampleutt{"other side"} of \exampleutt{"what you just made"} (i.e. the line of four blocks on the board). Such implicit references to spatial layouts and to existing structures are very common in the MDC dialogues. But we see another shortcoming of our metrics: because both models place a row of two green blocks on the ground (and because our shape metric allows for rotations on the horizontal plane), they achieve location, shape and overall F1 scores of 0.5, and type and color scores of 1.0, even though \llmbest arguably understood \exampleutt{other side} and \exampleutt{90 degrees} better than \llmbaseline.

\subsection{Takeaways}
\label{sec:qual_eval_concl}

Our qualitative analysis is consistent with  the quantitative results in Table \ref{tab:micro_best_v_baseline_new}: Both models are adept at identifying simpler aspects of the task, such as action type and color, reflected in their high Type and Color F1 scores. \llmbest also demonstrates a superior understanding of required block shapes (e.g., diagonals, corners), which aligns with its higher Shape F1 score. The primary differentiator, however, is spatial reasoning. \llmbest's  improvement in this area, demonstrated by its higher Location F1 score, validates the impact of our synthetic data and its richer input representations. 

\paragraph{Error Analysis} However, our error analysis illustrates that \llmbest still makes significant errors in precise spatial understanding, indicating that much work remains to be done to solve this task.

\paragraph{Evaluation Challenges and the Role of Auxiliary Metrics}
The qualitative analysis reveals a more nuanced challenge for fair evaluation: the overall F1 metric lacks the sensitivity to distinguish differences in error quality. As our qualitative analysis reveals, it is possible for two models to make different types of errors, with one being qualitatively "better" or closer to the correct solution, yet the F1 metric may not capture this distinction. This demonstrates that our proposed auxiliary metrics can play a crucial role, either by identifying the types of mistakes a model makes or by identifying cases where the model displayed at least a partial understanding of the instruction. 
At the same time, our analysis  underscores the inherent challenges in automated evaluation for the BAP task. 

The examples in Scenario 3 also illustrate the difficulty of creating a clear, unambiguous rubric for grading model output on an absolute scale, making human evaluation equally complex. This complicates not only absolute scoring but also comparative pairwise evaluations (i.e., "Is model A better than model B?"). While our qualitative analysis includes examples where a clear distinction can be made, many cases are far more ambiguous. In fact, we initially considered human evaluation but ultimately did not pursue it due to these challenges. Also, given that the primary goal of this work is to provide a holistic revision of the entire BAP framework—spanning evaluation, data, and modeling—we concluded that a deep investigation into human evaluation methodologies was beyond the current scope and needed thorough investigation owing to the aforementioned challenges.
This highlights the need for further research into BAP evaluation by building on our v2 benchmark.

\paragraph{Implications for LLMs}
Finally, the persistence of these subtle spatial errors, as observed above in Section~\ref{sec:qual_eval_analysis}, suggests limitations in how current text-only LLMs handle fine-grained 3D spatial grounding in embodied tasks, specifically in a 3D, embodied, collaborative dialogue setting (see Section~\ref{sec:rw_spatial_reasoning} below for more discussion in the broader context of related work), thus motivating future work.

\input{rw_spatial_reasoning}

%% file: rw_spatial_reasoning.tex
\subsection{Related Work: BAP evaluation} 
\label{sec:rw_bap_eval}

\citet{chaturvedi-etal-2024-nebula} also identify some of the issues with BAP evaluation on \dmc that we do, and hence propose evaluation on synthetic data. They design simpler, non-dialogue-based data with corresponding evaluation metrics. 
The data is single-turn and evaluates a model’s ability to follow an instruction in two scenarios: when the board is empty and the instruction is to build a single shape, or when a single shape is already present and the instruction involves placing or removing one block. 

While this data is intentionally simple to test elementary Builder capabilities, it overly simplifies the task and therefore deviates significantly from the target MCBT/BAP. It lacks a dialogue component, game history, and relative spatial references such as "left" and "right" as the Builder's position and orientation are not considered. The world state/built structure is also fairly minimal, either empty or containing a single shape.
While useful for elementary testing, it is a bit unclear how well insights from this data translate to the full BAP task.  
In contrast, our synthetic data remains simpler than \dmc while also closely emulating the MCBT, addressing these limitations and preserving many of the key BAP features and challenges, such as diverse instructions, rich spatial relations and floating blocks for example (more in Sections~\ref{sec:features_mcbt} and~\ref{sec:syn_data_gen}).
Although designed/motivated for training more than evaluation, its simplicity allows assessment of basic model competencies, while its emulation of the MCBT enhances its applicability to the overall BAP task, thus striking a better balance from an evaluation standpoint.
Section~\ref{sec:syn_data_eval} details additional benefits of evaluating on our synthetic data.  

Additionally, our synthetic data evaluation, combined with fairer F1, enables meaningful performance comparisons between synthetic data and \dmc. \citet{chaturvedi-etal-2024-nebula} are unable to conduct such comparisons (albeit understandably) due to the limitations of the legacy/original F1 metric.  

At a high level, \citet{chaturvedi-etal-2024-nebula} propose evaluation on their synthetic data as a solution to BAP evaluation issues, whereas we take the complementary approach by addressing them directly to propose the BAP v2 evaluation benchmark. Additionally, as described above, we also highlight the importance of evaluation on our synthetic data and contrast it with theirs.

\subsection{Related Work: Spatial Reasoning}
\label{sec:rw_spatial_reasoning}
Early efforts to understand spatial reasoning in language models focused on extracting spatial relations from text and observing what conclusions about a real world a model trained only on rich language data could learn \citep{mirzaee-kordjamshidi-2022-transfer}. Spatial reasoning was often explored in simple text-based games \citep{shi2022stepgamenewbenchmarkrobust} or with carefully constructed reasoning tasks \citep{rizvi-etal-2024-sparc}. 

Spatial reasoning has also been explored in conjunction with other modalities. In Vision-and-Language Navigation many explore it over scene-graphs \citep{gopinathan-etal-2024-spatially, li-etal-2024-topviewrs}. Others explore spatial reasoning with text and image inputs. \citet{10.5555/3504035.3504651} explore what vision and language activations models learn from block tasks represented in 2D images: mapping out how models interpreted spatial relations but also verbs like "mirror" or "twist". \citet{du-etal-2024-embspatial} also explore how VLMs can reason over space in photographs, but in reference to objects within those objects (i.e. an object being on the right of the photo but to the left of a person in the photo). However,  \citet{kamath2023whatsupvisionlanguagemodels} found that VLMs spatial reasoning can lag well behind human performance on VQA-style tasks and \citet{wang2024pictureworththousandwords} find that VLMs often do not make adequate use of the images that they are given, especially when provided with rich text representations. 

Unfortunately, there is not much parallel work exploring the spatial capabilities of LLMs on text-only inputs, particularly for embodied tasks. \citet{yamada2024evaluating} explore some basic grid navigation abilities of prompted LLMs, which struggle to traverse simple grids represented in text. \citet{aghzal2024can} consider the problem from the perspective of text-only grid path-planning, showing that models can effectively link coordinates and directions, but struggle to generalize to larger grids with more obstacles. Others have explored LLMs in playing chess \citep{feng2023chessgptbridgingpolicylearning, zhang-etal-2025-complete} or sliding puzzle games \citep{chollet2025arcprize2024technical} to some degree of success, but we found no works combining 3D text-only grids and embodied collaborative dialogue.
Together, these gaps in the literature motivate the MDC setup: by excluding the visual modality used in \hyperref[sec:rw_task_and_data]{other tasks}, it allows us to explore the baseline spatial capabilities of text-only LLMs on embodied tasks in a small controlled setting that is still generalizable to concrete end-goals.

%% file: discussion.tex
\section{Future Work}
\label{sec:discussion}
We now discuss the implications of our work and outline directions for future research. This discussion is organized into two parts: the path forward for BAP (Section~\ref{sec:fw_bap_only}), and the broader impact of our work beyond BAP (Section~\ref{sec:broader_impl}).

\subsection{The path forward for BAP}
\label{sec:fw_bap_only}

The introduction of the BAP v2 framework, while a significant step forward, also illuminates several key directions for future research. The work detailed in this paper suggests that the next advancements on the BAP task can be organized around three core pillars: evaluation, data, and modeling.

\paragraph{Advancing Evaluation}
Section~\ref{sec:qual_eval_concl} discusses the persisting problems around both automated and human evaluation for BAP, in spite of the progress made in BAP v2. This highlights the need for further research into evaluation by building on our v2 benchmark.

\paragraph{Enhancing Synthetic Data Generation}
We note the broader challenges in using LLMs to generate coherent, state-aware synthetic data for embodied dialogue tasks (Section~\ref{sec:llm_for_syn_data}). A promising research direction is to explore methods for overcoming these limitations, potentially leveraging LLMs to create even more diverse, complex, and naturalistic training dialogues that still maintain grounding in the 3D world state.

\paragraph{Improving Model Capabilities}
With a new SOTA F1 score of 53.0, there is still considerable room for improvement. Key areas for future modeling work include
\begin{itemize}
    \item \textbf{Spatial Reasoning:} Our analysis (Section~\ref{sec:analysis}) confirmed that highly accurate and fine-grained 3D spatial reasoning remains a primary bottleneck, even for top-performing models. Future models must develop more robust mechanisms for interpreting complex, perspective-dependent spatial language.
    \item \textbf{Advanced Training Regimes:} More sophisticated training strategies could unlock further gains. For instance, exploring curriculum learning, where models are first trained on simpler synthetic data before moving to the more complex human-generated dialogues, could prove effective.
    \item \textbf{Architectural Exploration:} The relative success of a smaller fine-tuned specialist model motivates deeper exploration of this paradigm. Future efforts can build on \llmbest with more advanced architectures or fine-tuning techniques. Concurrently, the synthetic data could also be leveraged to improve the ICL performance of larger, general-purpose models (Section~\ref{sec:rw_modeling_specialized}).
\end{itemize}

\subsection{Broader implications of our work beyond BAP}
\label{sec:broader_impl}
While our focus in this work is on the BAP subtask within the MCBT setting, our contributions have broader applications and takeaways as well. Below, we discuss their relevance for other subtasks within MCBT, for research building directly on the MCBT, and for similar domains and tasks in the field; thus highlighting the potential applications of our work and directions for future research.

\paragraph{Within MCBT}
\label{sec:within_mcbt}
Although our primary focus is on BAP, the synthetic data and training methodologies introduced here are  applicable to other MCBT subtasks constrained by the limited size of the MDC, such as the Architect Utterance Generation (AUG) subtask \citep{narayan-chen-etal-2019-collaborative} summarized briefly in Section~\ref{sec:mcbt_subtasks}. AUG and BAP represent two key pillars for building fully interactive Architect and Builder agents. Additionally, other subtasks, such as deciding when the Builder should act versus speak or generating clarification questions during dialogue, could also benefit from our synthetic datasets. 
Our simulators and data were carefully designed to emulate the nuances of the MCBT itself (Section~\ref{sec:syn_data_gen}), making them directly usable or adaptable to these subtasks. The simulators' parameterized nature enables generating tailored data distributions to this end, supporting further advancements across the broader MCBT ecosystem.

\paragraph{Beyond MCBT and Related Works}
Potential applications of our work also go beyond the MCBT. Several works directly build upon or are inspired by the MCBT/MDC, including IGLU \citep{mohanty2024idat}, \citet{bonn-etal-2020-spatial}, \citet{thompson-etal-2024-discourse}, \citet{bonial-etal-2021-builder}, and \citet{madge2025mdcrminecraftdialoguecorpus}. Therefore, future work on them can also potentially benefit from the synthetic data.

\paragraph{For Other Similar Domains and Tasks}
Our findings also offer a key takeaway for research on embodied, multimodal, task-oriented dialogue in dynamic and spatially complex settings. We demonstrate a methodology for generating rich synthetic data within game-based or simulated environments by carefully emulating task- and game-specific nuances. This approach yields complete embodied task-oriented dialogues that integrate utterances, environmental actions, and other game state information, providing a robust strategy for such complex dialogue settings—an area that remains underexplored, as highlighted in Section~\ref{sec:syn_tod_embodied}. This enables future work in the domain.

%% file: related_work.tex


%% file: conclusion.tex
\section{Conclusion}
\label{sec:concl_fw}

In this work, we addressed the challenges of the Builder Action Prediction (BAP) task, a demanding testbed for grounded instruction following within the Minecraft Collaborative Building Task. To foster more efficient and meaningful progress, we introduced BAP v2, a holistically re-examined and upgraded task framework.
BAP v2 is built on three core contributions. First, we established an enhanced evaluation benchmark, featuring a cleaner test set and fairer, more insightful metrics. This new benchmark not only provides a more accurate measure of performance but also revealed that spatial reasoning is the primary performance bottleneck. Second, to address both data scarcity and the spatial reasoning challenge, we developed novel dialogue simulators to generate diverse synthetic training data rich in spatial language. Third, we leveraged this new data to train stronger models. Our new SOTA model, \llmbest, utilizes rich, context-aware input representations and the synthetic data to achieve an F1 score of 53.0 ---a significant 6 point improvement over previous work.

While this result marks a substantial advance, it also underscores the task's remaining difficulty, even for SOTA LLMs. By systematically improving the evaluation, data, and modeling paradigms, the BAP v2 framework provides a robust and reliable foundation for future research. It paves the way for developing more sophisticated models with advanced training regimes, creating richer synthetic data, and exploring new evaluation techniques to push the boundaries of what is possible for interactive agents in complex, collaborative, and embodied environments.

%% file: appendix_pdoc/appendix_full.tex
\input{appendix_pdoc/appendix_data_tmp}

\input{appendix_pdoc/appendix_exp_stuff}

\input{appendix_pdoc/appendix_ann_stuff}

%% file: appendix_pdoc/appendix_data_tmp.tex
\section{Appendix: Generating Synthetic Data}
\label{sec:appendix_syn_data_gen}
Sections~\ref{app:rand_data},~\ref{app:blocks_data}, and~\ref{app:shapes_data}  provide additional details (beyond those covered in Section~\ref{sec:syn_data_gen}) on how each of the three dialogue simulators implements the general simulation framework (Section~\ref{sec:syn_data_gen}) -- specifically, \step{1} (Architect Planning), \step{3} (Architect Instruction), and \step{4} (Optional Clarification Exchange).  Section~\ref{app:shape_structs} provides details on the elementary shapes used to define shape-based target structures (Section~\ref{sec:targets_shapes_based}).

\paragraph{Note}  
All distances are measured using Manhattan Distance.  
For \step{3} and \step{4}, we outline only the key elements of our template-based generation framework. Further details will be included in our eventual code and data release. Thus, we omit detailed lexical and syntactic variations of utterances, providing only a few representative examples.  

\subsection{Target structures}
Some global constraints on target structures are enforced during their generation:
\begin{enumerate} 
    \item The structure must have at least one block on the ground. 
    \item The entire structure must be "connected," i.e., each block shares at least one face or edge with at least one other block.
\end{enumerate}

\subsection{Random target structures and dialogues (\dr)}
\label{app:rand_data}

\begin{description}
\item[\step{1}: Architect Planning]\hfill
\begin{description}
\item[Next Action:]
\label{app:rand_data_next_action}
The first four actions in the dialogue are placements. Subsequent actions are sampled with a 90\% probability for placement and 10\% for removal.  
For placements, the algorithm selects a color uniformly randomly from the six colors and a location from a set of candidate positions. For removals, a block from the built structure is chosen at random. Candidate positions are defined as follows. If the board is empty (start of dialogue), candidate positions are limited to the ground. But if at least one block exists, a candidate position must be within the "connected neighborhood" of an existing block. The "connected neighborhood" includes all locations within a Manhattan distance of 2 that share a face or edge with an existing block (but not a sole corner). This forms a 3x3x3 cube around the block, excluding the 8 diagonal corners but including the 6 adjacent and 12 2D diagonal locations.

\end{description}

\item[\step{3}: Generating the Main Dialogue ]\hfill

\begin{description}
\item[Next Action]  
The instruction specifies the color of the block (e.g. \exampleutt{blue}), and when necessary, includes an optional adjective such as \exampleutt{floating} if temporary supporting blocks are required.  

\item[Reference Block]
A phrase is needed to refer to the reference block, determined by the following priority rules (from highest to lowest):  
\begin{enumerate}  
    \item If the reference block is the last placed/removed block, the phrase used is \exampleutt{the last block you added/removed} etc. If the last action was a placement, the algorithm may omit the reference block in the instruction (ellipsis) with a small probability.  
    \item If the reference block's color is unique in the built structure, it is referred to as \exampleutt{the [COLOR] block,} where \textit{[COLOR]} represents the block's color.  
    \item If the color is not unique, the block is disambiguated using its spatial position relative to other blocks of the same color, allowing references such as \exampleutt{the closest/leftmost [COLOR] block,} etc.  
\end{enumerate}  

\item[Optional Clarification Exchange]
The elements of the instruction that may be omitted include the location (i.e., the spatial relation + reference block) or the color of the next action.  
\B's clarification questions, such as \exampleutt{Where?} or \exampleutt{What color?} are then generated accordingly, followed by \A's response. 
\end{description}

\hfill\\
\item[Special Cases with No Reference Block]
\label{special_cases_random_data}  
There are two special cases where no reference block (and consequently no spatial relation involving it) is used: 
\begin{description}
\item[Removal of the Last Placed Block]
If the next action is a removal and the block to be removed is the last placed block, the following adjustments are made while keeping everything else unchanged:  
\begin{inparaenum}[1)]  
    \item No reference block is needed.  
    \item \A's instruction will be a simple command such as \exampleutt{remove that block}.
    \item No clarification exchanges are generated, as there is nothing to omit in the instruction regarding the location or color of the block to be removed.  
\end{inparaenum}  

\item[Start of Dialogue]
At the start of the dialogue (empty board), the following modifications are made: 
\begin{inparaenum}  
    \item The first action is the of a block on the ground (randomly sampled from target structure blocks on the ground). 
    \item Since there are no prior blocks, \A's instruction will be a simple directive, such as \exampleutt{First start by placing an orange block on the ground}.  
\end{inparaenum}  
\end{description}
\end{description}

\subsection{Blocks-based dialogues for shape-based structures (\dbs)}
\label{app:blocks_data}

\begin{description}
\item[\step{1}: Architect Planning]
Recall that a target structure is provided as input to the dialogue simulation. In this iteration of the simulation, blocks to be placed next are selected from those that remain unplaced, given the current built structure and the target.  
The algorithm first selects a single block to place. Based on this block and the reference block, it determines whether multiple blocks can be placed simultaneously while maintaining the same spatial relation. (We give an example of this later.) If so, multiple blocks are added; otherwise, only a single block is placed.

\begin{description}

\item[Next Block]  
When deciding the next block to place, the overall goal is to ensure a natural (human-like), ordered, and simple building process. 
More specifically, the following principles help guide this decision.
The structure should be constructed shape by shape, and it should be ensured that the next block is connected to the existing built structure. Out of the possible candidate next blocks, those with the same color and within the connected neighborhood (defined in Section~\ref{app:rand_data_next_action}) of the last action should be given higher priority.
The building process should also aim to minimize branching within the structure (i.e., building as linearly as possible) so as to reduce the number of instructions and spatial relations that \A needs to provide.
Lastly, temporary supporting blocks should be used sparingly and only when necessary.  
We implement these principles via a few heuristics.

\item[Reference Block]  
The approach is similar to the one in Section~\ref{app:rand_data}, with one special case where a reference block cannot be assigned (see Section~\ref{special_cases_blocks_data}).  

\item[Multiple Blocks (if any)]  
Next, we check whether multiple blocks (i.e., apart from the next block) can be placed "greedily" using the same reference block and spatial relation. For instance, if the next block is a yellow block placed to the left of the reference block, and another yellow block exists further to the left, \A may plan to place both simultaneously for efficiency.  
To this end, the algorithm computes the direction of the next block relative to the reference block and checks if additional unplaced blocks of the same color exist along that direction. It then selects the longest contiguous sequence of such blocks connected to the next block.  
\end{description}
\item[\step{3}: Generating the Main Dialogue]

\begin{description}
\item The approach is similar to the one in Section~\ref{app:rand_data} but additionally accounts for the number of blocks when multiple blocks are placed (e.g. to specify  \exampleutt{three yellow blocks}).

\item[Optional Clarification Exchange]
The approach is similar to the one in Section~\ref{app:rand_data}.
\end{description}

\hfill\\
\item[Special Cases with No Reference Block]  
\label{special_cases_blocks_data}  
There is only one special case, which is the same as the "Start of Dialogue" one described in Section~\ref{special_cases_random_data}.  
\end{description}

\subsection{Shape-based dialogues for shape-based structures (\dss)}
\label{app:shapes_data}

\begin{description}
\item[\step{1}: Architect Planning]
Recall that a target structure consisting of two shape instances is provided as input to the dialogue simulation.  

\begin{description}
\item[Next Shape and Its Starting Block]
Shapes are built sequentially. If both shape instances are on the ground, one is chosen randomly as the first to be built, otherwise the algorithm starts with the shape on the ground. 
Shapes are naturally built from the bottom up. The algorithm first selects a block within the shape to define spatial relations needed to specify where this shape should be built (e.g. wrt. the reference block). This block also serves as the starting point for \B when building the shape (the order of the remaining blocks is decided by \B's planning in Step 5 of the simulation algorithm of Section~\ref{sec:syn_data_gen}).
Potential starting blocks are always located in the bottom corners of the shape. For the first shape, the starting block is the one farthest from the second shape. 
The starting block for the second shape is the block with the shortest distance to the last placed block in the first shape. %

\item[Reference Block]  
For the second shape being placed, the reference block is the last placed block of the first shape. No reference block is used for the first shape.  
\end{description}
\item[\step{3}: Generating the Main Dialogue]\hfill
\begin{description}
\item[Next Shape]
A shape’s size can be described in multiple ways depending on its dimensions. For 1D shapes such as rows, the description includes a single number, such as \exampleutt{four blocks long} or \exampleutt{three blocks tall.} For 2D shapes like diagonals and planes, various formats can be used, such as \exampleutt{3x3}, \exampleutt{two blocks long and three tall}, etc.  

The instruction also includes the shape’s color and name. We provide examples here for three shapes -- rows, diagonals and planes. The primary shape names are \exampleutt{row,} \exampleutt{diagonal,} and \exampleutt{plane,} though synonyms can be used. A horizontal row may be called a \exampleutt{line,} while a vertical row may be referred to as a \exampleutt{column}, \exampleutt{pillar}, or \exampleutt{tower.} Similarly, a horizontal plane may be described as a \exampleutt{layer,} whereas a vertical plane may be called a \exampleutt{wall.} A horizontal diagonal is typically referred to as a \exampleutt{diagonal} or \exampleutt{diagonal line,} but a vertical diagonal can also be described as a \exampleutt{staircase} or \exampleutt{stairway.} 

\item[Reference Block] 
This applies only to the second shape being placed. Since the reference block is the last placed block, we use phrases such as \exampleutt{the last block you placed,} similar to the approach in Sections~\ref{app:rand_data} and~\ref{app:blocks_data}.

\item[Optional Clarification Exchange]
The approach is similar to the one in Sections~\ref{app:rand_data} and~\ref{app:blocks_data}, but now, the size/dimension and relative location/orientation of the shape to be placed (using expressions such as \exampleutt{going to the left of you} that use \B as a spatial anchor, c.f. Section~\ref{sec:spatial_rels}) can also be omitted. 
\end{description}

\end{description}

\subsection{Elementary shapes}
\label{app:shape_structs}
The six elementary shapes used to generate the structures for \dbs and \dss are defined  as follows: 
\begin{description}
    \item[Rows] consist of at least 3 blocks, and can point in three different directions (\texttt{X}, \texttt{Y} and \texttt{Z}), aligning to each of the $x$, $y$, and $z$ axes. Those that align to the $y$ axis are columns.
	\item[Diagonals] consist of at least 3 blocks, and are planar (along any of the $xy$, $yz$, or $xz$ planes). \textit{Horizontal} diagonals lie along the $xz$ plane,  \textit{vertical} along either $xy$ or $yz$.
	\item[T-shapes] ($\geq 3\times3$ blocks) are composed of two orthogonal rows, in which one end of one row intersects the midpoint of another. The orientations of these rows determine the orientation of the T. Horizontal T-shapes lie completely along the $xz$ plane, while vertical ones span either $xy$ or $yz$. Additionally, vertical T-shapes can be up (facing right-side up) or down (upside-down).
	\item[L-shapes] ($\geq 2\times2$ blocks) are similar to T-shapes in construction in that they consist of two orthogonal rows attached at their ends. Orientations  are similar to those of T-shapes, including horizontal L-shapes as well as vertical ones that can point either up or down.
	\item[U-shapes] are symmetrical structures  built out of rows: one row (with a length $\geq 3$) constitutes its base, while two orthogonal rows that are parallel to each other (with equal lengths $\geq 2$) constitute the sides.
	U-shape orientations are  similar to those of T- and L-shapes, including horizontal and vertical (up and down).
	\item[Planes]  are 2D shapes  ($\geq 3\times2$ blocks)  that are either horizontal along the $xz$ plane or vertical along the $xy$ or $yz$ planes.
\end{description}

%% file: appendix_pdoc/appendix_exp_stuff.tex
\section{Appendix: Experimental Details}

\subsection{How much of each synthetic dataset to use during train time?}
\label{app:data_mix}
We provide further details of the procedure in Section~\ref{sec:syn_data_props}.
We first train models on the three synthetic datasets (\syndata) separately to study how increasing the training data affects performance. Focusing on the saturation regions of the learning curves—where performance plateaus and additional data has diminishing returns—we identify various "candidate" data amounts (\#items) that yield near-peak performance for each dataset. 
Using these data amounts, we construct a grid of combinations across the three datasets, where each grid point corresponds to a different data mix and proportion of each dataset. 
For each grid point, we train a model on the combined \syndata and \dmc datasets. The combination that achieves the best performance on \dmc is selected. Thus, the approach yields a data mix that optimizes \dmc performance while also being highly performant on the synthetic data as much as possible.
Also, in the best data mixes, the \#items per dataset generally followed the order \dbs < \dss < \dr, which matches their difficulty ordering.

\subsection{Experimental Setup for GRU-based Models}
\label{app:exp_setup_gru}
We provide further details of the experimental setup for models \magg and \maggcl (apart from those in Section~\ref{sec:exp_setup_batch_training}), again closely following the setup used for the baseline model \mmc when it was trained on \dmc alone (Section~\ref{sec:baseline_model_desc}).

\paragraph{Training}
\label{app:gru_training_setup}
We train models using AdamW~\cite{loshchilov2017decoupled} and weight decay regularization with a weight decay factor of 0.1. The learning rate is set to 0.0001, and the batch size is 1. Early stopping is applied when the loss on the held-out development set increases monotonically for ten epochs.  
The development set for \magg and \maggcl, used for early stopping, is constructed from the validation splits of \alldata (Table~\ref{tab:results7}) to match the dataset distribution in the training set.

\paragraph{Model Hyperparameters}
\label{sec:gru_hparams}
We continue to use the same hyperparameter values as those of the best model in \citet{jayannavar-etal-2020-learning}.

%% file: appendix_pdoc/appendix_ann_stuff.tex
\section{Appendix: Annotation Setup} 
\label{sec:app_ann_details}

Our annotation team comprises one expert annotator (an author) and 10 non-expert annotators, who are undergraduate and graduate computer science students. The non-experts received elaborate training from the expert annotator before the final data collection.  
The NEB (non-empty board) subset is annotated by the 10 non-expert annotators, with each BAP item annotated by two annotators. Disagreements were resolved by the expert annotator, and the workload was evenly distributed among the annotators (in terms of the number of dialogues per annotator). The smaller and easier-to-annotate EB (empty board) subset was exclusively annotated by the expert annotator.  
Annotators were instructed to annotate one dialogue at a time, processing items within a dialogue in chronological order. This ensures they accurately understand the evolving game context and remain unbiased by future "unseen" events in a dialogue.